\documentclass[10pt,twocolumn,letterpaper]{article}

\usepackage{iccv}
\usepackage{times}
\usepackage{epsfig}
\usepackage{graphicx}
\usepackage{mathdefs}
\usepackage{caption}
\usepackage{color}
\usepackage{nicefrac}
\usepackage{units}
\usepackage{enumitem}
\usepackage{multirow}
\usepackage{subfigure}
\usepackage{xfrac}
\usepackage{bm}
\usepackage{colortbl}
\usepackage{xspace}
\usepackage[symbol]{footmisc}

\usepackage[accsupp]{axessibility}
\usepackage{appendix}

\definecolor{myyellow}{RGB}{230, 160, 0}
\definecolor{darkred}{RGB}{139, 0, 0}

\graphicspath{{figures/}}
\usepackage[pagebackref=true,breaklinks=true,letterpaper=true,colorlinks,bookmarks=false]{hyperref}

\iccvfinalcopy

\begin{document}

\title{Defocus Map Estimation and Deblurring from a Single Dual-Pixel Image}

\author{Shumian Xin$^1 \footnotemark$ \qquad
Neal Wadhwa$^2$\qquad
Tianfan Xue$^2$\qquad
Jonathan T. Barron$^2$\qquad
\\
Pratul P. Srinivasan$^2$\qquad
Jiawen Chen$^3$ \qquad
Ioannis Gkioulekas$^1$ \qquad
Rahul Garg$^2$
\\
$^1$Carnegie Mellon University \qquad
$^2$Google Research \qquad
$^3$Adobe Inc.
}

\maketitle
\footnotetext[1]{Work primarily done when Shumian Xin was an intern at Google.}

\vspace{-0.1in}
\begin{abstract}\vspace{-0.1in}
	We present a method that takes as input a single dual-pixel image, and simultaneously estimates the image's defocus map---the amount of defocus blur at each pixel---and recovers an all-in-focus image. Our method is inspired from recent works that leverage the dual-pixel sensors available in many consumer cameras to assist with autofocus, and use them for recovery of defocus maps or all-in-focus images. These prior works have solved the two recovery problems independently of each other, and often require large labeled datasets for supervised training. By contrast, we show that it is beneficial to treat these two closely-connected problems simultaneously. To this end, we set up an optimization problem that, by carefully modeling the optics of dual-pixel images, jointly solves both problems. We use data captured with a consumer smartphone camera to demonstrate that, after a one-time calibration step, our approach improves upon prior works for both defocus map estimation and blur removal, despite being entirely unsupervised.
\end{abstract}

\vspace{-0.21in}
\section{Introduction}

\begin{figure}[t]
	\begin{center}
		\includegraphics[width=\linewidth]{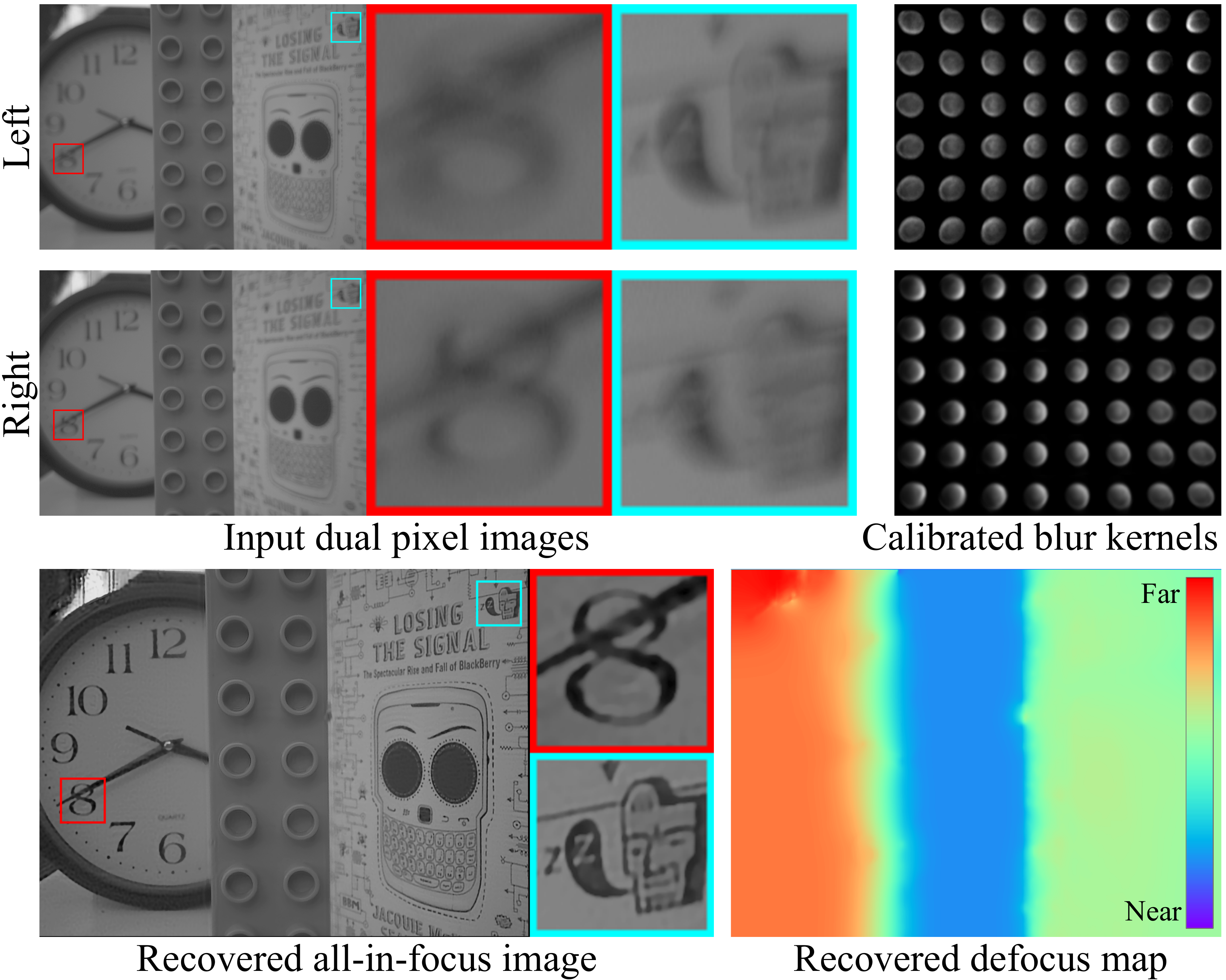}
	\end{center}
	\vspace{-0.25in}
	\caption{Given left and right dual-pixel (DP) images and  corresponding spatially-varying blur kernels,
		our method jointly estimates an all-in-focus image and defocus map.
	}
	\label{fig:teaser}
	\vspace{-0.2in}
\end{figure}

\begin{figure*}[t]
	\centering
	\vspace{-0.1in}
	\includegraphics[width=\textwidth]{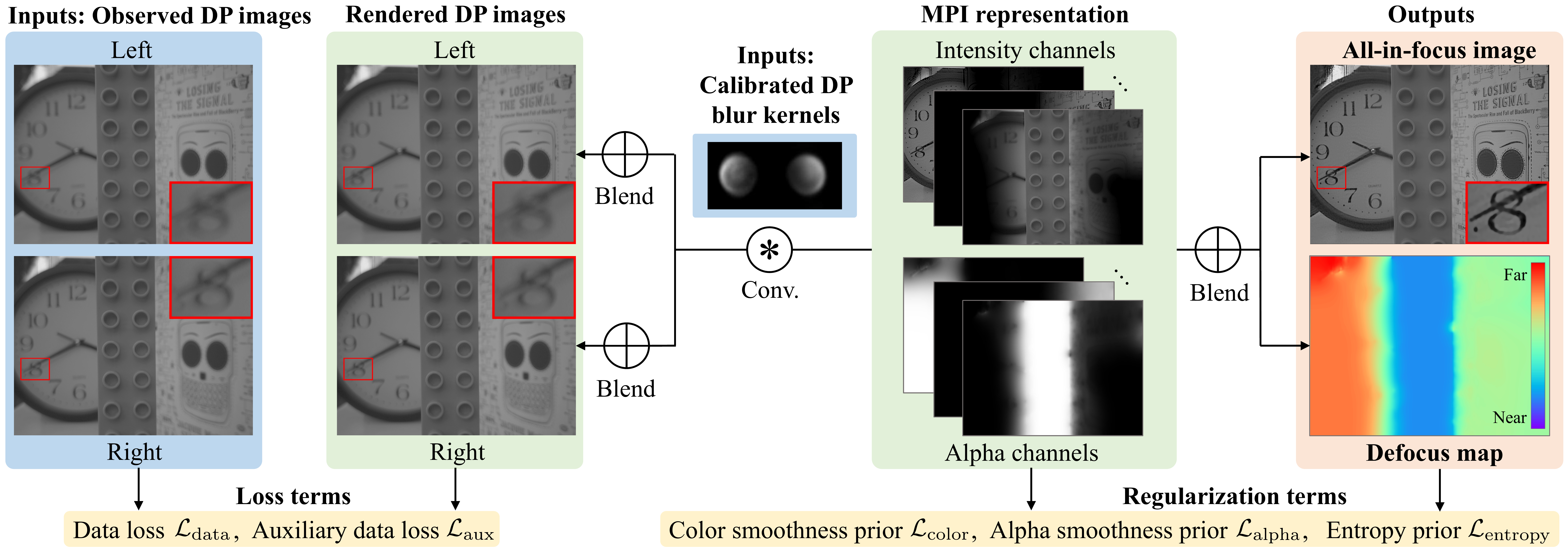}
	\vspace{-0.25in}
	\caption{Overview of our proposed method. We use input left and right DP images to fit a multiplane image (MPI) scene representation, consisting of a set of fronto-parallel layers. Each layer is an {\RGBA} image containing the in-focus scene content at the corresponding depth. The MPI can output the all-in-focus image and the defocus map by blending all layers. It can also render out-of-focus images, by convolving each layer with pre-calibrated blur kernels for the left and right DP views, and then blending. We optimize the MPI by minimizing a regularized loss comparing rendered and input images.}
	\label{fig:teaser2}
	\vspace{-0.15in}
\end{figure*}

Modern DSLR and mirrorless cameras feature large-aperture lenses that allow collecting more light, but also introduce
{\em defocus} blur, meaning that objects in images appear blurred by an amount proportional to their distance from the focal plane. A simple way to reduce defocus blur is to \emph{stop down}, i.e., shrink the aperture.
However, this also reduces the amount of light reaching the sensor, making the image noisier. 
Moreover, stopping down is impossible on fixed-aperture cameras, such as those in most smartphones. More sophisticated techniques fall into two categories. First are techniques that add extra hardware (e.g., coded apertures~\cite{levin2007image}, specialized lenses~\cite{levin20094d,dowski1995extended}), and thus are impractical to deploy at large scale or across already available cameras. Second are \emph{focus stacking} techniques~\cite{Supasorn2015} that capture multiple images at different focus distances, and fuse them into an all-in-focus image. These techniques require long capture times, and thus are applicable only to static scenes.

Ideally, defocus blur removal should be done using data from a single capture. Unfortunately, in conventional cameras, this task is fundamentally ill-posed: a captured image may have no high-frequency content because either the latent all-in-focus image lacks such frequencies, or they are removed by defocus blur. Knowing the \emph{defocus map}, i.e., the spatially-varying amount of defocus blur, can help simplify blur removal. However, determining the defocus map from a single image is closely-related to monocular depth estimation, which is a challenging problem in its own right. Even if the defocus map were known, recovering an all-in-focus image is still an ill-posed problem, as it requires hallucinating the missing high frequency content.

\emph{Dual-pixel} (DP) sensors are a recent innovation that makes it easier to solve both the defocus map estimation and defocus blur removal problems, with data from a single capture. Camera manufacturers have introduced such sensors to many DSLR and smartphone cameras to improve autofocus~\cite{abuolaim2018focus,herrmann2020}. Each pixel on a DP sensor is split into two halves, each capturing light from half of the main lens' aperture, yielding two sub-images per exposure (Fig.~\ref{fig:teaser}). These can be thought of as a two-sample lightfield~\cite{ng2005lightcamera}, and their sum is equivalent to the image captured by a regular sensor. The two sub-images have different half-aperture-shaped defocus blur kernels; these are additionally spatially-varying due to optical imperfections such as vignetting or field curvature in lenses, especially for cheap smartphone lenses.

We propose a method to simultaneously recover the defocus map and all-in-focus image from a single DP capture. Specifically, we perform a one-time calibration to determine the spatially-varying blur kernels for the left and right DP images. Then, given a single DP image, we optimize a \emph{multiplane image} (MPI) representation~\cite{szeliski1999stereo,zhou2018stereo} to best explain the observed DP images using the calibrated blur kernels.
An MPI is a layered representation that accurately models occlusions, and can be used to render both defocused and all-in-focus images, as well as produce a defocus map.
As solving for the MPI from two DP images is under-constrained, we introduce additional priors and show their effectiveness via ablation studies. Further, we show that in the presence of image noise, standard optimization has a bias towards underestimating the amount of defocus blur, and we introduce a bias correction term. Our method does not require large amounts of training data, save for a one-time calibration, and outperforms prior art on both defocus map estimation and blur removal, when tested on images captured using a consumer smartphone camera. We make our implementation and data publicly available~\cite{ProjectWebsite}.

\vspace{-0.1in}
\section{Related Work}

\noindent{\bf Depth estimation.}
Multi-view depth estimation is a well-posed and extensively studied problem~\cite{Hartley2003,scharstein2002taxonomy}. By contrast, single-view, or \emph{monocular}, depth estimation is ill-posed. Early techniques attempting to recover depth from a single image typically relied on additional cues, such as silhouettes, shading, texture, vanishing points, or data-driven supervision~\cite{bajcsy1976texture,Barron2012A,Brady84,choi2015depth,hane2015direction,Hoiem2005,horn1970shape,konrad2013learning,Ladicky_2014_CVPR,li2014dept,ranftl2016dense,Saxena2005,shi2015break}. The use of deep neural networks trained on large RGBD datasets~\cite{eigen2014, DORN2018, li2017two, liu2015learning, roy2016monocular, SilbermanECCV12NYUdv2} significantly improved the performance of data-driven approaches, motivating approaches that use synthetic data~\cite{atapour2018real,guo2018learning,mayer2018makes,nath2018adadepth,zou2018df}, self-supervised training~\cite{garg2016unsupervised,godard2017,Godard2018,Jiang2018,MahjourianWA18,zhou2017unsupervised}, or multiple data sources~\cite{facil2019cam,Ranftl2020}. Despite these advances, producing high-quality depth from a single image remains difficult, due to the inherent ambiguities of monocular depth estimation. 

Recent works have shown that DP data can improve monocular depth quality, by resolving some of these ambiguities. Wadhwa \etal~\cite{wadhwa2018synthetic} applied classical stereo matching methods to DP views to compute depth. Punnappurath \etal~\cite{punnappurath2020modeling} showed that explicitly modeling defocus blur during stereo matching can improve depth quality. However, they assume that the defocus blur is spatially invariant and symmetric between the left and right DP images, which is not true in real smartphone cameras. Depth estimation with DP images has also been used as part of reflection removal algorithms~\cite{Punnappurath_2019_CVPR}. Garg \etal~\cite{garg2019learning} and Zhang \etal~\cite{zhang2020du2net} trained neural networks to output depth from DP images, using a captured dataset of thousands of DP images and ground truth depth maps~\cite{SoftwareSync2019}. The resulting performance improvements come at a significant data collection cost.

Focus or defocus has been used as a cue for monocular depth estimation prior to these DP works. Depth from defocus techniques~\cite{favaro2010recovering, Pentland87,Tang2017,watanabe1998rational} use two differently-focused images with the same viewpoint, whereas depth from focus techniques use a dense focal stack~\cite{Grossmann87,hazirbas18ddff,Supasorn2015}. Other monocular depth estimation techniques use defocus cues as supervision for training depth estimation networks~\cite{Srinivasan2018}, use a coded aperture to estimate depth from one~\cite{levin2007image,veeraraghavan2007dappled,zhou2009good} or two captures~\cite{Zhou2009ICCV}, or estimate a defocus map using synthetic data~\cite{Lee_2019_CVPR}. Lastly, some binocular stereo approaches also explicitly account for defocus blur~\cite{chen2015blur, li2010dual}; compared to depth estimation from DP images, these approaches assume different focus distances for the two views. 

\noindent{\bf Defocus deblurring.} 
Besides depth estimation, measuring and removing defocus blur is often desirable to produce sharp all-in-focus images.
Defocus deblurring techniques usually estimate either a depth map or an equivalent defocus map as a first processing stage~\cite{dandres2016non,karaali2017edge,park2017unified,shi2015just}. Some techniques modify the camera hardware to facilitate this stage. Examples include inserting patterned occluders in the camera aperture to make defocus scale selection easier~\cite{levin2007image,veeraraghavan2007dappled,zhou2009good,Zhou2009ICCV}; or sweeping through multiple focal settings within the exposure to make defocus blur spatially uniform~\cite{Nagahara08}.
Once a defocus map is available, a second deblurring stage employs non-blind deconvolution methods~\cite{levin2007image,fish1995blind,krishnan2009fast,Wang08TV,Tomer14,zhang2017learning} to remove the defocus blur.

Deep learning has been successfully used for defocus deblurring as well. Lee \etal~\cite{Lee_2019_CVPR} train neural networks to regress to defocus maps, that are then used to deblur. Abuolaim and Brown~\cite{abuolaim2020defocus} extend this approach to DP data, and train a neural network to directly regress from DP images to all-in-focus images. Their method relies on a dataset of pairs of wide and narrow aperture images captured with a DSLR, and may not generalize to images captured on smartphone cameras, which have very different optical characteristics. Such a dataset is impossible to collect on smartphone cameras with fixed aperture lenses. In contrast to these prior works, our method does not require difficult-to-capture large datasets. Instead, it uses an accurate  model of the defocus blur characteristics of DP data, and simultaneously solves for a defocus map and an all-in-focus image.

\vspace{-0.1in}
\section{Dual-Pixel Image Formation\label{sec:dual_pixel}}

\begin{figure}[t]
	\begin{center}
		\includegraphics[width=\linewidth]{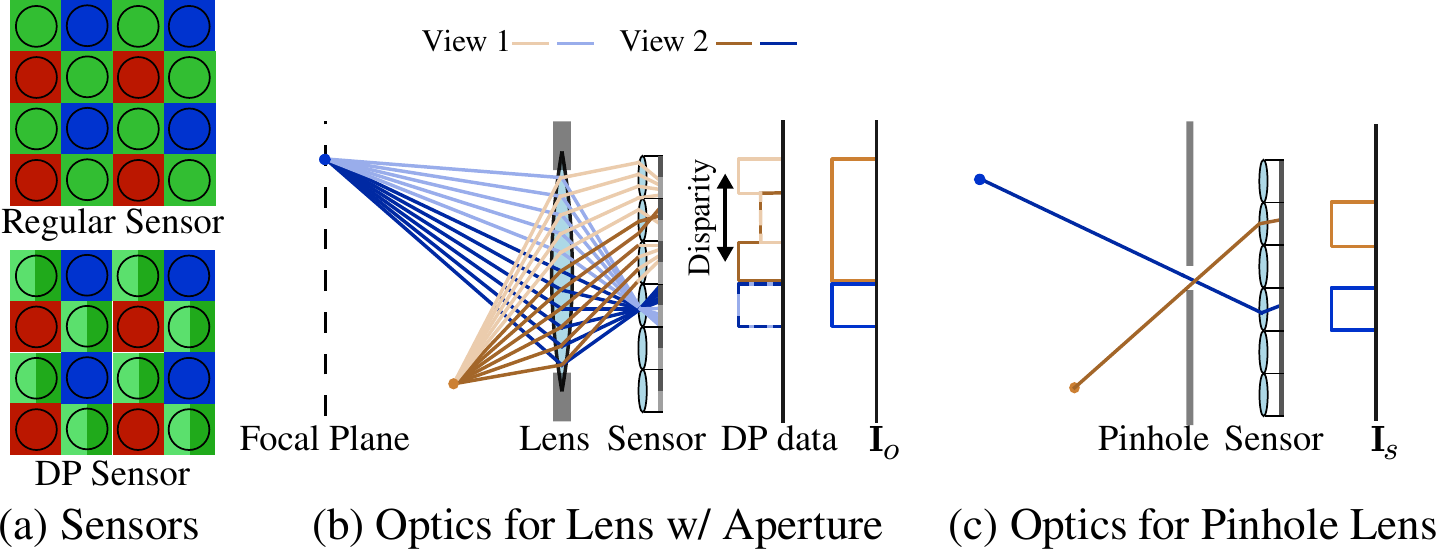}
	\end{center}
	\vspace{-0.2in}
	\caption{A regular sensor and a DP sensor (a) where each green pixel is split into two halves. For a finite aperture lens (b), an in-focus scene point produces overlapping DP images, whereas an out-of-focus point produces shifted DP images. Adding the two DP images yields the image that would have been captured by a regular sensor. (c) shows the corresponding pinhole camera where all scene content is in focus. Ignoring occlusions, images in (b) can be generated from the image in (c) by applying a depth-dependent blur.}
	\label{fig:dp_optics}
	\vspace{-0.2in}
\end{figure}

\begin{figure}[t]
	\centering
	\subfigure[Left DP image blur kernels]{\includegraphics[width=0.23\textwidth]{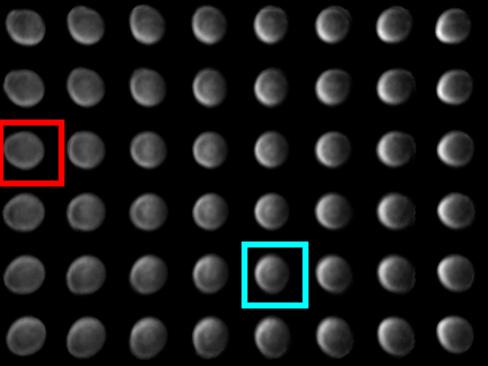}\label{subfig:calib_left}}
	\subfigure[Right DP image blur kernels]{\includegraphics[width=0.23\textwidth]{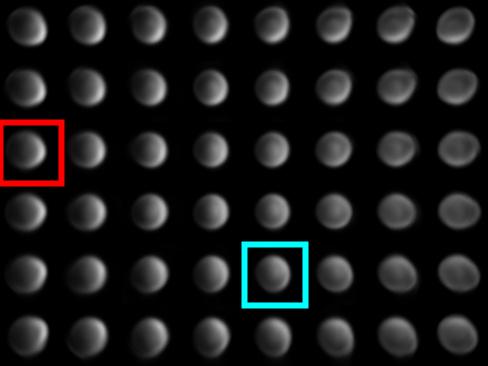}\label{subfig:calib_right}}\\
	\vspace{-0.1in}
	\subfigure[]{\includegraphics[width=0.15\textwidth]{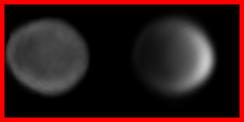}\label{subfig:calib_eg_1}} 
	\subfigure[]{\includegraphics[width=0.15\textwidth]{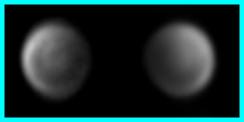}\label{subfig:calib_eg_2}} 
	\subfigure[Kernels from \cite{punnappurath2020modeling}]{\includegraphics[width=0.15\textwidth]{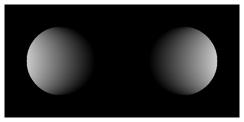}\label{subfig:calib_iccp}} 
	\vspace{-0.2in}
	\caption{Calibrated blur kernels \subref{subfig:calib_left} and \subref{subfig:calib_right} for the left and right DP images. \subref{subfig:calib_eg_1} and \subref{subfig:calib_eg_2} show example pairs of left and right kernels marked in red and cyan. Compared to the parametric kernels \subref{subfig:calib_iccp} from~\cite{punnappurath2020modeling}, calibrated kernels are spatially-varying, not circular, and not left-right symmetric.}
	\label{fig:dual_pixels}
	\vspace{-0.2in}
\end{figure}

We begin by describing image formation for a regular and a dual-pixel (DP) sensor, to relate the defocus map and the all-in-focus image to the captured image. For this, we consider a camera imaging a scene with two points, only one of which is in focus (Fig.~\ref{fig:dp_optics}(b)). Rays emanating from the in-focus point (blue) converge on a single pixel, creating a sharp image. By contrast, rays from the out-of-focus point (brown) fail to converge, creating a blurred image.

If we consider a lens with an infinitesimally-small aperture (i.e., a pinhole camera), only rays that pass through its center strike the sensor, and create a sharp all-in-focus image $\I_{s}$ (Fig.~\ref{fig:dp_optics}(c)).
Under the thin lens model, the blurred image $\I_{o}$ of the out-of-focus point equals blurring $\I_{s}$ with a depth-dependent kernel $\kernel{d}{}$, shaped as a $d$-scaled version of the aperture---typically a circular disc of radius $d = A + \nicefrac{B}{Z}$, where $Z$ is the point depth, and $A$ and $B$ are lens-dependent constants~\cite{garg2019learning}.
Therefore, the per-pixel signed kernel radius $d$, termed the \emph{defocus map} $\D$, is a linear function of inverse depth, thus a proxy for the depth map. Given the defocus map $\D$, and ignoring occlusions, the sharp image $\I_s$ can be recovered from the captured image $\I_o$ using non-blind deconvolution. In practice, recovering either the defocus map $\D$ or the sharp image $\I_s$ from a single image $\I_o$ is ill-posed, as multiple $(\I_s, \D)$ combinations produce the same image $\I_o$. Even when the defocus map $\D$ is known, determining the sharp image $\I_s$ is still ill-posed, as blurring irreversibly removes image frequencies.

DP sensors make it easier to estimate the defocus map.
In DP sensors (Fig.~\ref{fig:dp_optics}(a)), each pixel is split into two halves, each collecting light from the corresponding half of the lens aperture (Fig.~\ref{fig:dp_optics}(b)). Adding the two half-pixel, or DP, images $\I_{o}^{l}$ and $\I_{o}^{r}$ produces an image
equivalent to that captured by a regular sensor, i.e., $\I_o=\I_{o}^{l} + \I_{o}^{r}$.
Furthermore, DP images are identical for an in-focus scene point, and shifted versions of each other for an out-of-focus point.
The amount of shift, termed \emph{DP disparity}, is proportional to the blur size, and thus provides an alternative for defocus map estimation. In addition to facilitating the estimation of the defocus map $\D$, having two DP images instead of a single image provides additional constraints for recovering the underlying sharp image $\I_s$. Utilizing these constraints requires knowing the blur kernel shapes for the two DP images.

\noindent{\bf Blur kernel calibration.} As real lenses have spatially-varying kernels, we calibrate an $8\times6$ grid of kernels. To do this, we fix the focus distance, capture a regular grid of circular discs on a monitor screen, and solve for blur kernels for left and right images independently using a method similar to Mannan and Langer~\cite{mannan2016}. When solving for kernels, we assume that they are normalized to sum to one, and calibrate separately for vignetting: we average left and right images from six captures of a white diffuser, using the same focus distance as above, to produce left and right vignetting patterns $W_l$ and $W_r$. We refer to the supplement for details.

We show the calibrated blur kernels in Fig.~\ref{fig:dual_pixels}. We note that these kernels deviate significantly from parametric models derived by extending the thin lens model to DP image formation~\cite{punnappurath2020modeling}. In particular, the calibrated kernels are spatially-varying, not circular, and not symmetric.

\vspace{-0.1in}
\section{Proposed Method}

The inputs to our method are two single-channel DP images, and calibrated left and right blur kernels. We correct for vignetting using $W_l$ and $W_r$, and denote the two vignetting-corrected DP images as $\I_{o}^{l}$ and $\I_{o}^{r}$, and their corresponding blur kernels at a certain defocus size $d$ as $\kernel{d}{l}$ and $\kernel{d}{r}$, respectively. We assume that blur kernels at a defocus size $d'$ can be obtained by scaling by a factor $\nicefrac{d'}{d}$~\cite{punnappurath2020modeling, Zhou2009ICCV}. Our goal is to optimize for the multiplane image (MPI) representation that best explains the observed data, and use it to recover the latent all-in-focus image $\hat{\I}_s$ and defocus map $\hat{\D}$. We first introduce the MPI representation, and show how to render defocused images from it. We then formulate an MPI optimization problem, and detail its loss function.

\subsection{Multiplane Image (MPI) Representation}

\begin{figure}[t]
	\begin{center}
		\includegraphics[width=\linewidth]{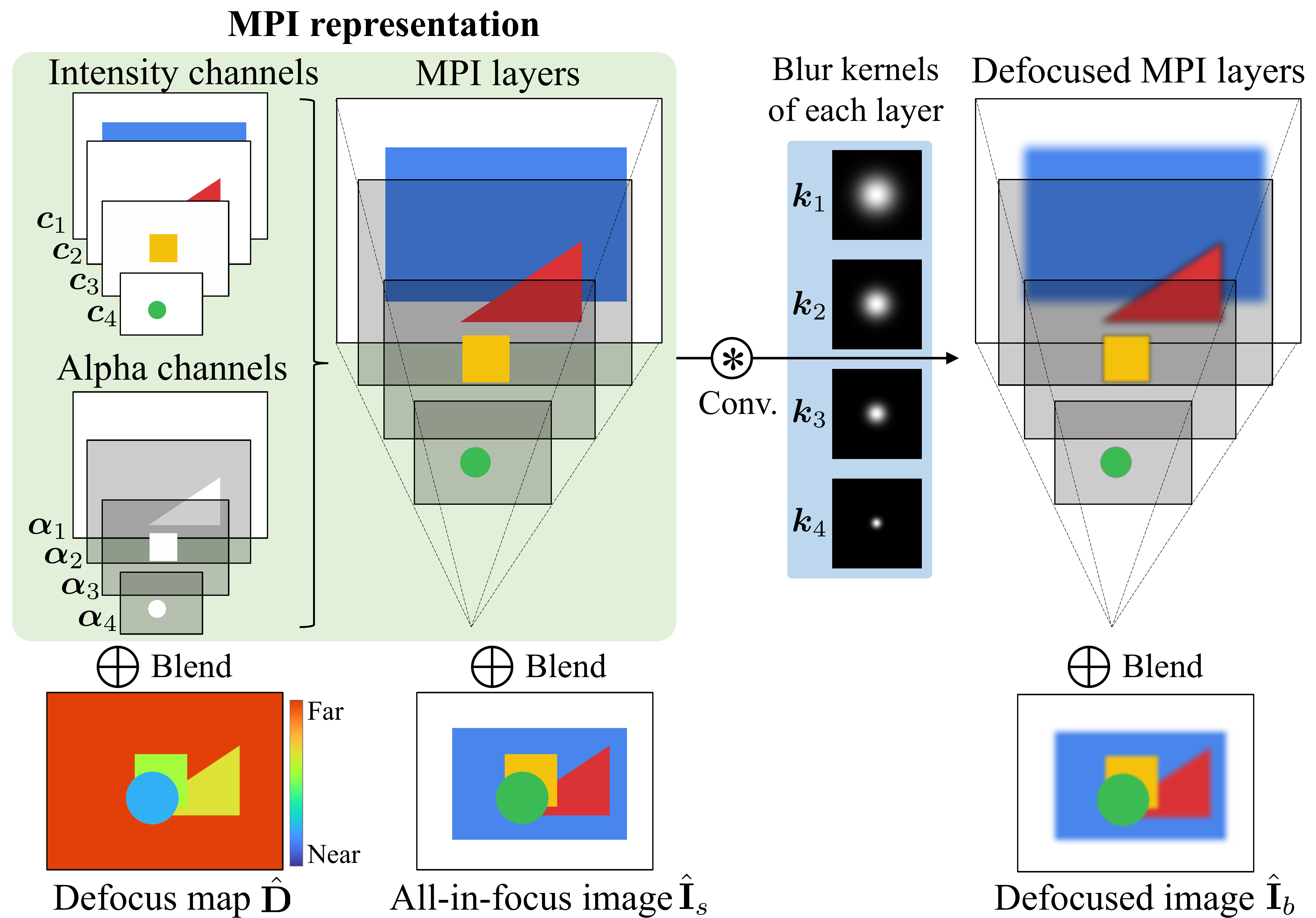}
	\end{center}
	\vspace{-0.25in}
	\caption{The multiplane image (MPI) representation consists of discrete fronto-parallel planes where each plane contains intensity data and an alpha channel. We use it to recover the defocus map, the all-in-focus image, and render a defocused image according to a given blur kernel.}
	\label{fig:mpi}
	\vspace{-0.2in}
\end{figure}

We model the scene using the MPI representation, previously used primarily for view synthesis~\cite{single_view_mpi, zhou2018stereo}. MPIs discretize the 3D space into $N$ fronto-parallel planes at fixed depths (Fig.~\ref{fig:mpi}).
We select depths corresponding to linearly-changing defocus blur sizes $\bracket{d_1, \dots, d_N}$.
Each MPI plane is an {\RGBA} image of the in-focus scene that consists of an intensity channel $\cc_i$ and an alpha channel $\ac_i$.

\noindent{\bf All-in-focus image compositing.} Given an MPI, we composite the sharp image using the \emph{over} operator~\cite{ljung2016state}:
we sum all layers weighted by the transmittance of each layer $\tc_i$,
\small
\begin{equation}
	\hat{\I}_s = \sum_{i=1}^{N} \tc_i \cc_i
	= \sum_{i=1}^{N} \bracket{ \cc_i \ac_i \prod_{j=i+1}^{N} \paren{1-\ac_j}}\,.
	\label{eq:sharp_im_mpi}
\end{equation}
\normalsize

\noindent{\bf Defocused image rendering.} 
Given the left and right blur kernels $\kernel{d_i}{\curlylr}$ for each layer,
we render defocused images by convolving each layer with its corresponding kernel, then compositing the blurred layers as in Eq.~\eqref{eq:sharp_im_mpi}:
\begin{equation}
	\resizebox{2.95in}{!}{
		$\displaystyle
		\hat{\I}_b^{\curlylr} = \sum_{i=1}^{N} \bracket{ {\paren{\kernel{d_i}{\curlylr} * \paren{\cc_i \ac_i}} \odot \prod_{j=i+1}^{N} \paren{\one-\kernel{d_j}{\curlylr} * \ac_j}}}\,,
		$}
	\label{eq:defocused_im_mpi}
\end{equation}
where $*$ denotes convolution. In practice, 
we scale the calibrated spatially-varying left and right kernels by the defocus size $d_i$, and apply the scaled spatially-varying blur to each {\RGBA} image $\cc_i \ac_i$. We note that we render left and right views from a single MPI, but with different kernels.

\subsection{Effect of Gaussian Noise on Defocus Estimation \label{subsec:bias_correction}}
\vspace{-0.2in}

Using Eq.~\eqref{eq:defocused_im_mpi}, we can optimize for the MPI that minimizes the $L_2$-error $\norm{\hat{\I}_b^{\curlylr} - \I_o^{\curlylr}}_2^2$ between rendered images $\hat{\I}_b^{\curlylr}$ and observed DP images $\I_o^{\curlylr}$. Here we show that, in the presence of noise, this optimization is biased toward smaller defocus sizes, and we correct for this bias.

Assuming additive white Gaussian noise $\noise^{\curlylr}$ distributed as $\Normal(0, \sigma^2)$, we can model DP images as:
\small
\begin{equation}
	\I_o^{\curlylr} = {\I}_b^{\curlylr} + \noise^{\curlylr} \,,
	\label{eq:formulation}
\end{equation}
\normalsize
where ${\I}_b^{\curlylr}$ are the latent noise-free images. For simplicity, we assume for now that all scene content lies on a single fronto-parallel plane with ground truth defocus size $d^{\star}$. Then, using frequency domain analysis similar to Zhou \etal~\cite{Zhou2009ICCV}, we prove in the supplement that for a defocus size hypothesis $d_i$, the expected negative log-energy function corresponding to the MAP estimate of the MPI is: 
\small
\begin{gather}
	\!\nonumber E\big(d_i| \kernelf{d^\star}{\curly{l,r}}, \sigma\big)\!=\!\sum_{f}\! C_1\big(\kernelf{d_i}{\curly{l,r}},\sigma,\boldsymbol{\Phi}\big)\!\abs{\!\kernelf{d^\star}{l} \kernelf{d_i}{r} - \kernelf{d^\star}{r} \kernelf{d_i}{l}\!}^2 \\ 
	+ \sigma^2\sum_{f} \bracket{ \frac{ |\kernelf{d^\star}{l}|^2 + |\kernelf{d^\star}{r}|^2 + \sigma^2 |\boldsymbol{\Phi}|^2 }{|\kernelf{d_i}{l}|^2 + |\kernelf{d_i}{r}|^2  + \sigma^2 |\boldsymbol{\Phi}|^2}  } + C_2(\sigma) \,,
	\label{eq:bias_derivation}
\end{gather}
\normalsize
where $\kernelf{d_i}{\curlylr}$ and $\kernelf{d^\star}{\curlylr}$ are the Fourier transforms of kernels $\kernel{d_i}{\curlylr}$ and $\kernel{d^\star}{\curlylr}$ respectively, $\boldsymbol{\Phi}$ is the inverse spectral power distribution of natural images, and the summation is over all frequencies. We would expect the loss to be minimized when $d_i = d^\star$. The first term measures the inconsistency between the hypothesized blur kernel $d_i$ and the true kernel $d^\star$, and is indeed minimized when $d_i = d^\star$. However, the second term depends on the noise variance and decreases as $\abs{d_i}$ decreases. This is because, for a normalized blur kernel ($\norm{\kernel{d_i}{\curlylr}}_1 = 1$), as the defocus kernel size $\abs{d_i}$ decreases, its power spectrum $\norm{\kernelf{d_i}{\curlylr}}_2$ increases. This suggests that white Gaussian noise in input images results in a bias towards smaller blur kernels. To account for this bias, we subtract an approximation of the second term, which we call the \emph{bias correction term}, from the optimization loss:
\begin{equation}
	\resizebox{2.95in}{!}{%
		$
		\bias\paren{d_i| \kernelf{d^\star}{\curly{l,r}}, \sigma} \approx \sigma^2\sum_{f}  \frac{ \sigma^2 \abs{\boldsymbol{\Phi}}^2 }{\abs{\kernelf{d_i}{l}}^2 + \abs{\kernelf{d_i}{r}}^2  + \sigma^2 \abs{ \boldsymbol{\Phi}}^2}  \,.
		$}
	\label{eq:bias_correction}
\end{equation}
We ignore the terms containing ground truth $d^{\star}$, as they are significant only when $d^\star$ is itself small, i.e., the bias favors the true kernels in that case. In an MPI with multiple layers associated with defocus sizes $\bracket{d_1, \dots, d_N}$, we subtract per-layer constants $\bias\paren{d_i}$ computed using Eq.~\eqref{eq:bias_correction}. 

We note that we use a Gaussian noise model to make analysis tractable, but captured images have mixed Poisson-Gaussian noise~\cite{hasinoff2010noise}. In practice, we found it beneficial to additionally denoise the input images using burst denoising~\cite{hasinoff2016burst}.
However, there is residual noise even after denoising, and we show in Sec.~\ref{sec:results} that our bias correction term still improves performance. An interesting future research direction is using a more accurate noise model to derive a better bias estimate and remove the need for any denoising.

\subsection{MPI Optimization \label{sec:opt_losses}}

We seek to recover an MPI $\curly{\cc_i, \ac_i}, i \in\bracket{1, \dots, N}$ such that defocused images rendered from it using the calibrated blur kernels are close to the input images. But minimizing only a reconstruction loss is insufficient: this task is ill-posed, as there exists an infinite family of MPIs that all exactly reproduce the input images. As is common in defocus deblurring~\cite{levin2007image}, we regularize our optimization:
\small
\begin{equation}
	\loss = \loss_{\mathrm{data}} + \loss_{\mathrm{aux}} + \loss_{\mathrm{intensity}} + \loss_{\mathrm{alpha}} + \loss_{\mathrm{entropy}}\,,
\end{equation} 
\normalsize
where $\loss_{\mathrm{data}}$ is a bias-corrected data term that encourages rendered images to resemble input images, $\loss_{\mathrm{aux}}$ is an auxiliary data term applied to each MPI layer, and the remaining are regularization terms. We discuss all terms below.

\noindent{\bf Bias-corrected data loss.} 
We consider the Charbonnier~\cite{charbonnier1994two} loss function $\huber\paren{x} = \sqrt{\nicefrac{x^2}{\gamma^2} + 1}$, and define a bias-corrected version as $\huber_{\bias}\paren{x, \bias} = \sqrt{\nicefrac{(x^2 - \bias)}{\gamma^2} + 1}$, where we choose the scale parameter $\gamma=0.1$~\cite{Barron19RobustLoss}.
We use this loss function to form a data loss penalizing the difference between left and right input and rendered images as:
\small
\begin{align}
	\loss_{\mathrm{data}} &= \sum_{\substack{x, y}} \huber_{\bias} \paren{\hat{\I}_b^{\curlylr}(x, y) - \I_o^{\curlylr}(x, y), \bias^{\curly{l,r}}_{\mathrm{all}}} \,, 
	\label{eq:dataloss}
	\\
	\bias^{\curly{l,r}}_{\mathrm{all}} &= \sum_{i=1}^{N} \bracket{ {\kernel{d_i}{\curlylr} *  \ac_i\!\!\prod_{j=i+1}^{N}\!\!\paren{1-\kernel{d_j}{\curlylr} * \ac_j}}\!}\!\bias(d_i) \,.
	\label{eq:dataloss_bias}
\end{align}
\normalsize
We compute the total bias correction $\bias^{\curly{l,r}}_{\mathrm{all}}$ as the sum of all bias correction terms of each layer, weighted by the corresponding defocused transmittance. Eq.~\eqref{eq:dataloss_bias} is equivalent to Eq.~\eqref{eq:defocused_im_mpi} where we replace each MPI layer's intensity channel $\cc_i$ with a constant bias correction value $\bias\paren{d_i}$. To compute $\bias\paren{d_i}$ from Eq.~\eqref{eq:bias_correction}, we empirically set the variance to $\sigma^2 = 5\cdot 10^{-5}$, and use a constant inverse spectral power distribution $\abs{\boldsymbol{\Phi}}^2=10^2$, following previous work~\cite{tang2012utilizing}.

\noindent{\bf Auxiliary data loss.} In most real-world scenes, a pixel's scene content should be on a single layer. However, because the compositing operator of Eq.~\eqref{eq:defocused_im_mpi} forms a weighted sum of all layers, $\loss_{\mathrm{data}}$ can be small even when scene content is smeared across multiple layers. To discourage this, we introduce a per-layer auxiliary data loss on each layer's intensity weighted by the layer's blurred transmittance:
\small
\begin{align}
	\nonumber \loss_{\mathrm{aux}} = \sum_{x, y, i} &\paren{\kernel{d_i}{\curlylr} * \tc_i(x, y)} \odot  \\
	&\huber_{\bias} \paren{\kernel{d_i}{\curlylr} * \cc_i(x, y) - \I_o^{\curlylr}(x, y), \bias\paren{d_i}} \,,
\end{align}
\normalsize
where $\odot$ denotes element-wise multiplication. This auxiliary loss resembles the data synthesis loss of Eq.~\eqref{eq:dataloss}, except that it is applied to each MPI layer separately.

\noindent{\bf Intensity smoothness.}
Our first regularization term encourages smoothness for the all-in-focus image and the MPI intensity channels.
For an image $\I$ with corresponding edge map $\edge$, we define an edge-aware smoothness based on total variation $\tv(\cdot)$, similar to Tucker and Snavely~\cite{single_view_mpi}:
\small
\begin{equation}
	\tv_{\edge} \paren{\I, \boldsymbol{E}} = \huber\paren{\tv\paren{\I}} + \paren{1 - \boldsymbol{E}} \odot \huber\paren{\tv\paren{\I}},
\end{equation}
\normalsize
where $\huber(\cdot)$ is the Charbonnier loss. We refer to the supplement for details on $\edge$ and $\tv(\cdot)$. Our smoothness prior on the all-in-focus image and MPI intensity channels is:
\begin{equation}
	\resizebox{2.9in}{!}{
		$\displaystyle
		\loss_{\mathrm{intensity}} =  \sum_{x,y} \tv_{\edge}\paren{\hat{\I}_s, \boldsymbol{E}\paren{\hat{\I}_s}} + \sum_{x, y, i}  \tv_{\edge}\paren{\tc_i \cc_i, \boldsymbol{E}\paren{\tc_i \cc_i}}.
		$}
\end{equation}

\noindent{\bf Alpha and transmittance smoothness.} We use an additional smoothness regularizer on all alpha channels and transmittances (sharpened by computing their square root), by encouraging edge-aware smoothness according to the total variation of the all-in-focus image:
\begin{equation}
	\resizebox{2.9in}{!}{
		$\displaystyle
		\loss_{\mathrm{alpha}} = \sum_{x, y, i} \bracket{\tv_{\edge}\paren{\sqrt{\ac_i}, \boldsymbol{E}\paren{\hat{\I}_s}} + \tv_{\edge}\paren{\sqrt{\tc_i}, \boldsymbol{E}\paren{\hat{\I}_s}} }.
		$}
\end{equation}

\noindent{\bf Alpha and transmittance entropy.} 
The last regularizer is a collision entropy penalty on alpha channels and transmittances. Collision entropy, defined for a vector $\x$ as $S\paren{\x} = -\log \nicefrac{\norm{\x}_2^2}{\norm{\x}_1^2}$, is a special case of Renyi entropy~\cite{renyi1961measures},
and we empirically found it to be better than Shannon entropy for our problem. Minimizing collision entropy encourages sparsity: $S\paren{\x}$ is minimum when all but one elements of $\x$ are $0$, which in our case encourages scene content to concentrate on a single MPI layer, rather than spread across multiple layers. Our entropy loss is:
\small
\begin{align}
	\nonumber \loss_{\mathrm{entropy}}\! &= \!\sum_{x,y}\! S \paren{\bracket{\sqrt{\ac_2}\paren{x,y}, \dots, \sqrt{\ac_N}\paren{x,y}}^{\mathrm{T}}}  \\ 
	&+\!\sum_{x,y}\! S \paren{\bracket{\sqrt{\tc_1}\paren{x,y}, \dots, \sqrt{\tc_N}\paren{x, y }}^{\mathrm{T}}}\,.
\end{align}
\normalsize
We extract the alpha channels and transmittances of each pixel $\paren{x,y}$ from all MPI layers, compute their square root for sharpening, compute a per-pixel entropy, and average these entropies across all pixels. When computing entropy on alpha channels, we skip the farthest MPI layer, because we assume that all scene content ends at the farthest layer, and thus force this layer to be opaque ($\ac_1 = \mathbf{1}$).

\vspace{-0.1in}
\section{Experiments}

We capture a new dataset, and use it to perform qualitative and quantitative comparisons with other state of the art defocus deblurring and defocus map estimation methods. The project website~\cite{ProjectWebsite} includes an interactive HTML viewer~\cite{bitterli2017reversible} to facilitate comparisons across our full dataset.

\noindent{\bf Data collection.}
Even though DP sensors are common, to the best of our knowledge, only two camera manufacturers provide an API to read DP images---Google and Canon.
However, Canon's proprietary software applies an unknown scene-dependent transform to DP data. Unlike supervised learning-based methods~\cite{abuolaim2020defocus} that can learn to account for this transform, our loss function requires raw sensor data. Hence, we collect data using the Google Pixel 4 smartphone, which allows access to the raw DP data \cite{dp_app}.

The Pixel 4 captures DP data only in the green channel. To compute ground truth, we capture a focus stack with $36$ slices sampled uniformly in diopter space, where the closest focus distance corresponds to the distance we calibrate for, $\unit[13.7]{cm}$, and the farthest to infinity. Following prior work~\cite{punnappurath2020modeling}, we use the commercial Helicon Focus software~\cite{helicon_focus} to process the stacks and generate ground truth sharp images and defocus maps, and we manually correct holes in the generated defocus maps. Still, there are image regions that are difficult to manually inpaint, e.g., near occlusion boundaries or curved surfaces. We ignore such regions when computing quantitative metrics. We capture a total of $17$ scenes, both indoors and outdoors.  Similar to Garg \etal~\cite{garg2019learning}, we centrally crop the DP images to $1008 \times 1344$. We refer to the supplement for more details. Our dataset is available at the project website~\cite{ProjectWebsite}.

\subsection{Results \label{sec:results}}

\begin{figure*}[ht]
	\centering
	\subfigure[\scriptsize Input image]{\includegraphics[width=0.157 \textwidth]{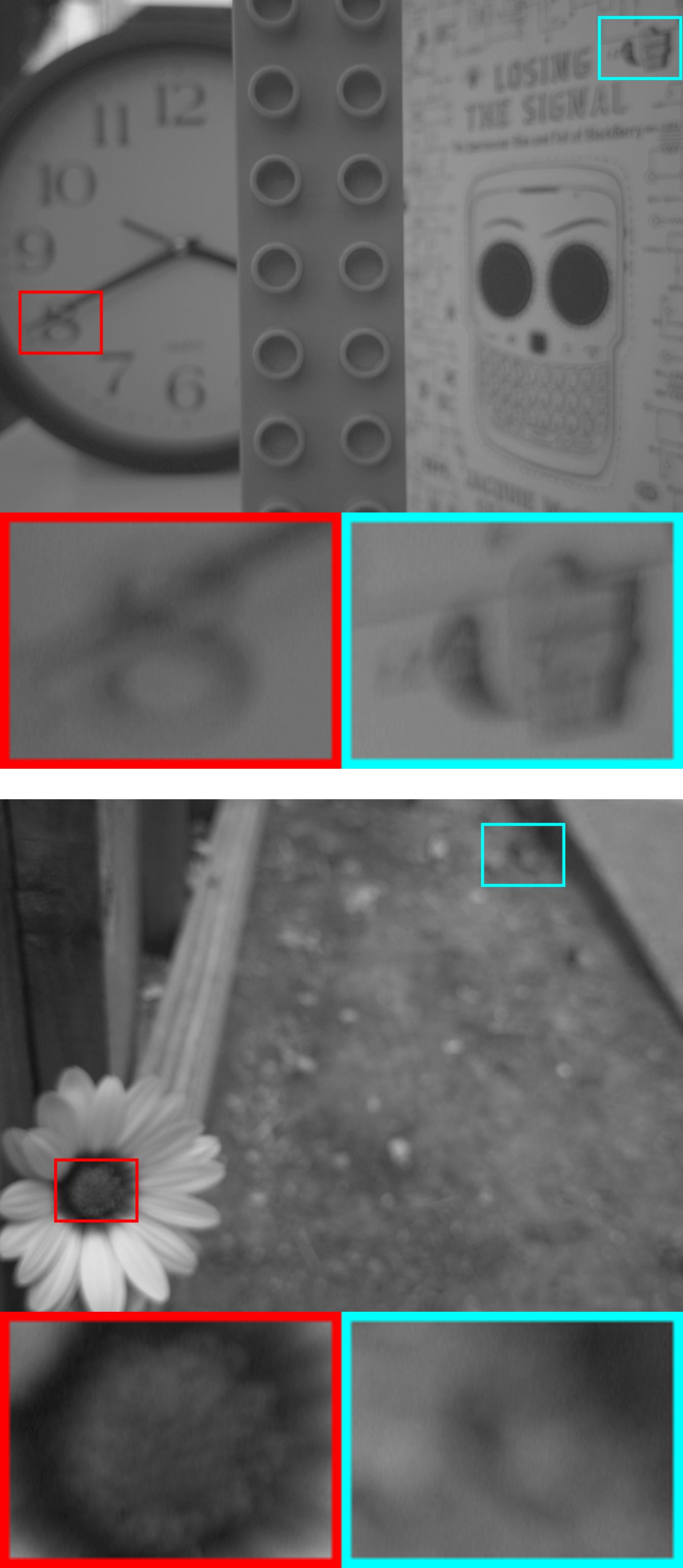} \label{subfig:inputs_dd_comparison}}
	\subfigure[\scriptsize GT all-in-focus image]{\includegraphics[width=0.157\textwidth]{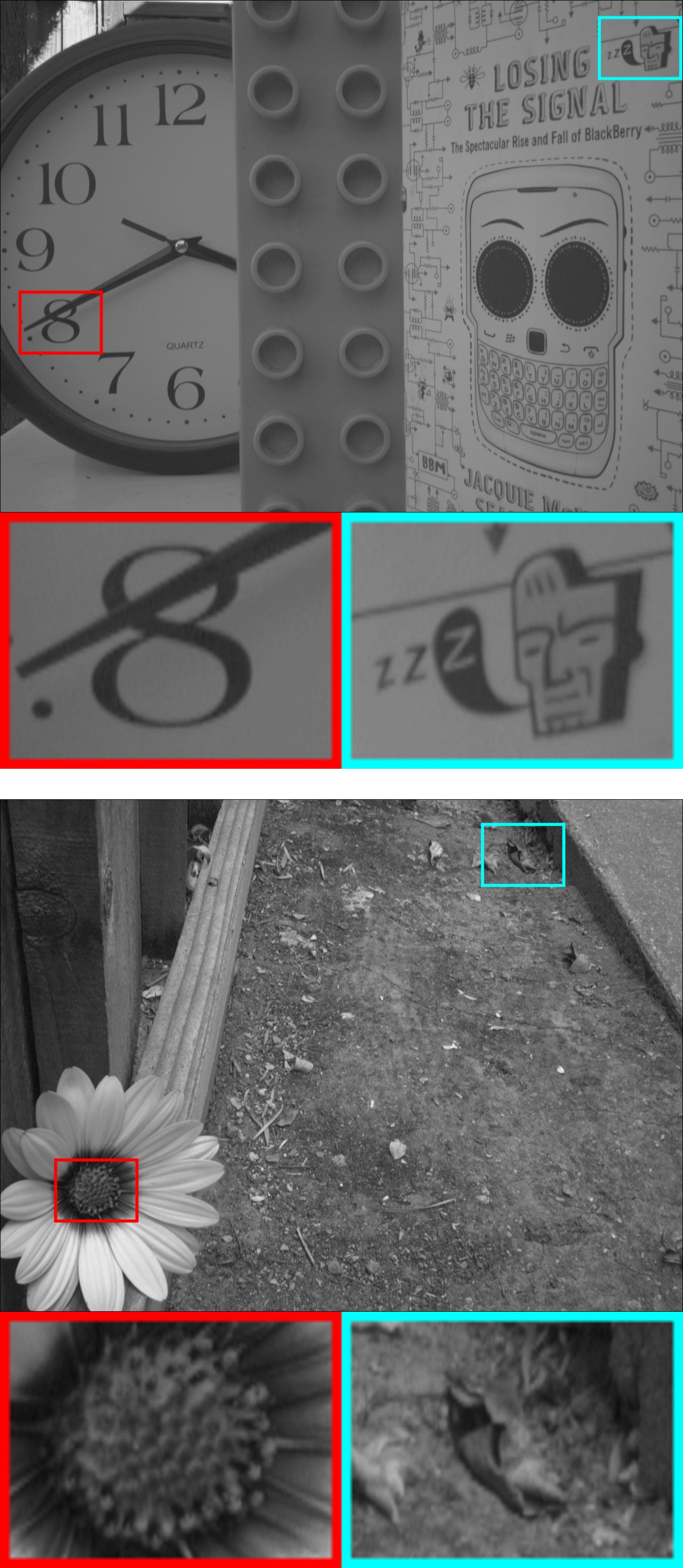} \label{subfig:sharp_im_gt_comparison}}
	\subfigure[\scriptsize Ours]{\includegraphics[width=0.157\textwidth]{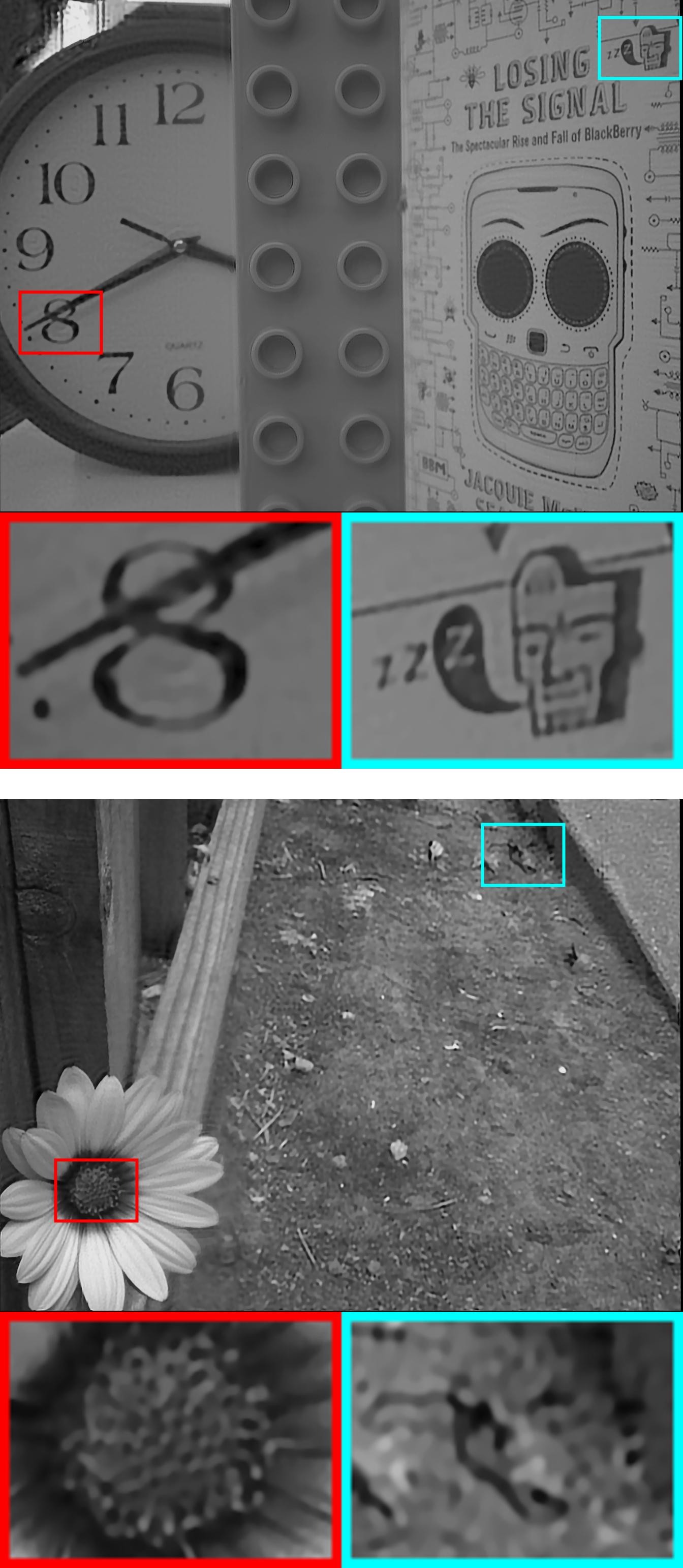} \label{subfig:sharp_im_ours_comparison}}
	\subfigure[\scriptsize Wiener deconv.~\cite{ Zhou2009ICCV}]{\includegraphics[width=0.157\textwidth]{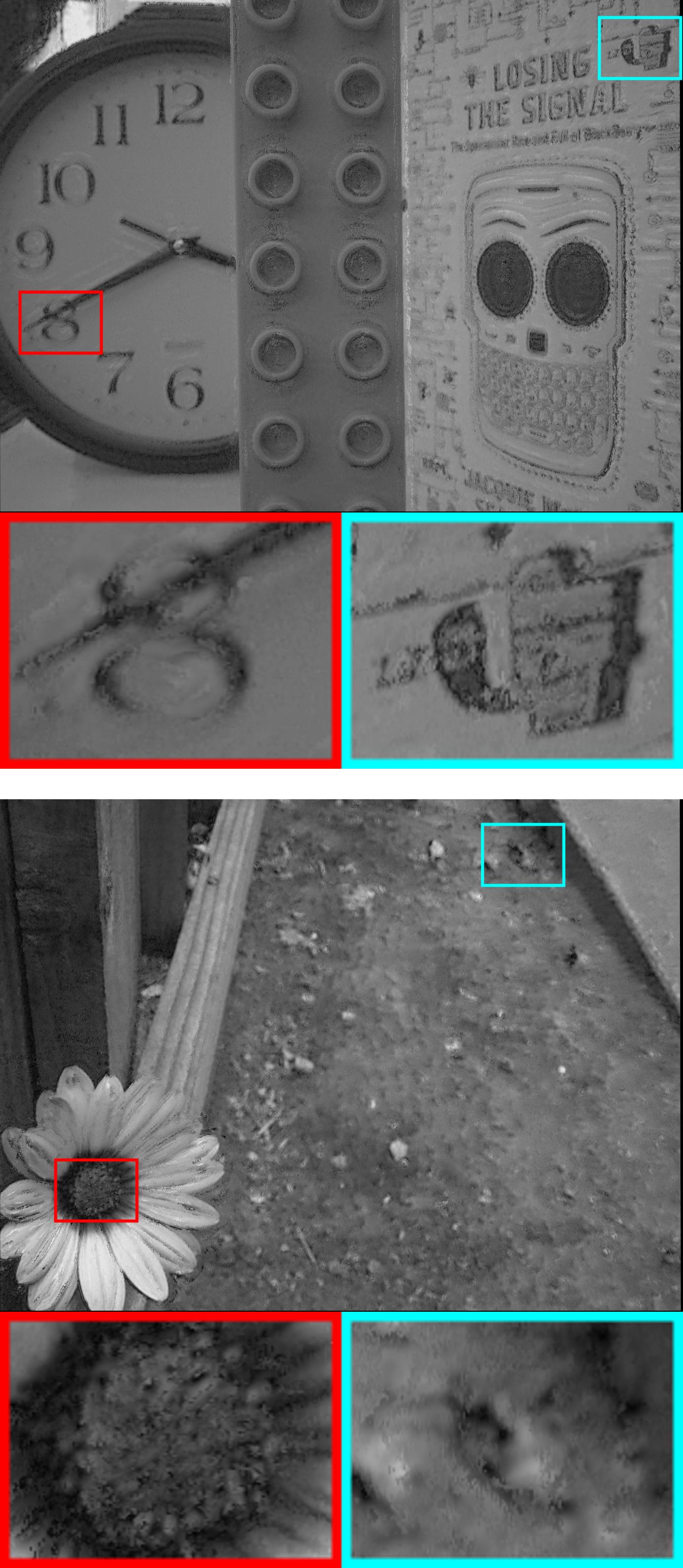} \label{subfig:sharp_im_wiener_comparison}}
	\subfigure[\scriptsize DPDNet~\cite{abuolaim2020defocus} \tiny (Orig. Input)]{\includegraphics[width=0.157\textwidth]{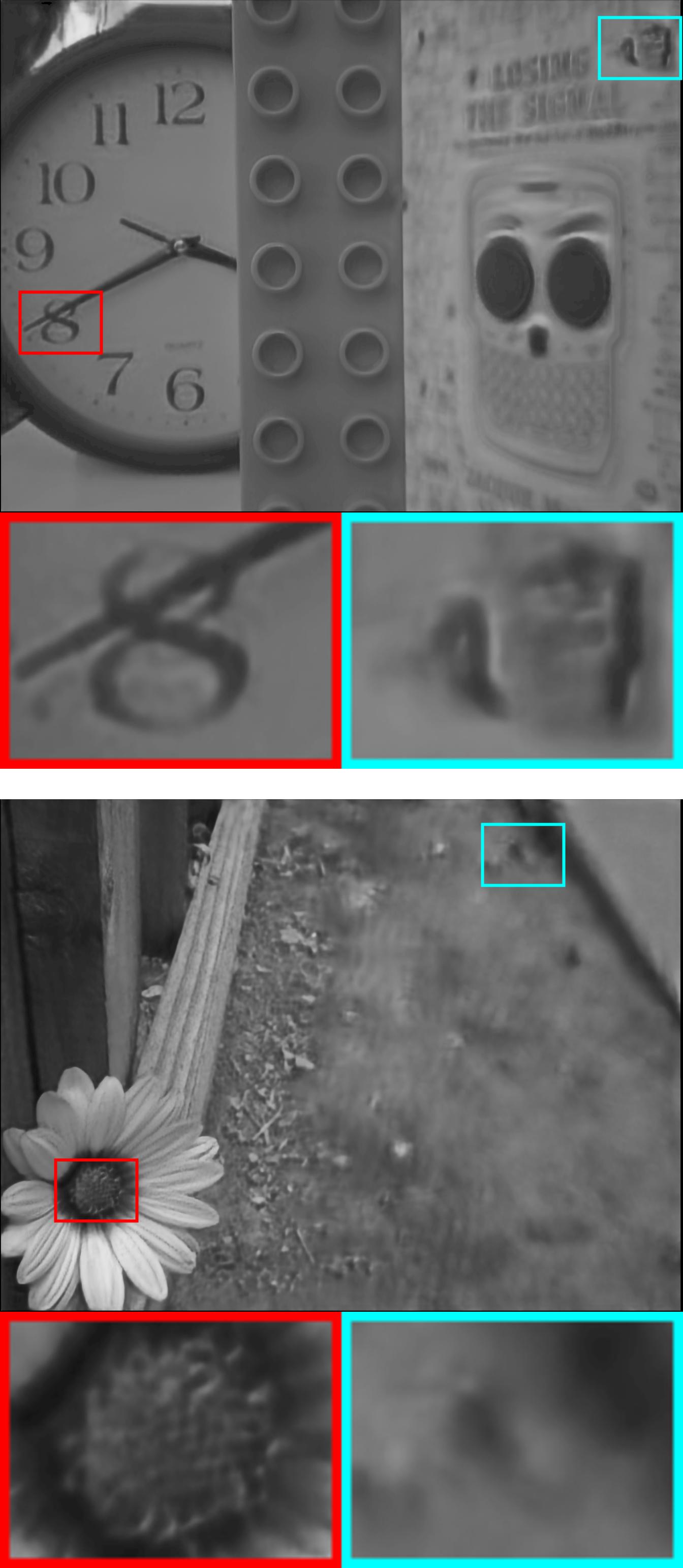} \label{subfig:sharp_im_ECCV_comparison}}
	\subfigure[\scriptsize DPDNet
	\cite{abuolaim2020defocus}]{\includegraphics[width=0.157\textwidth]{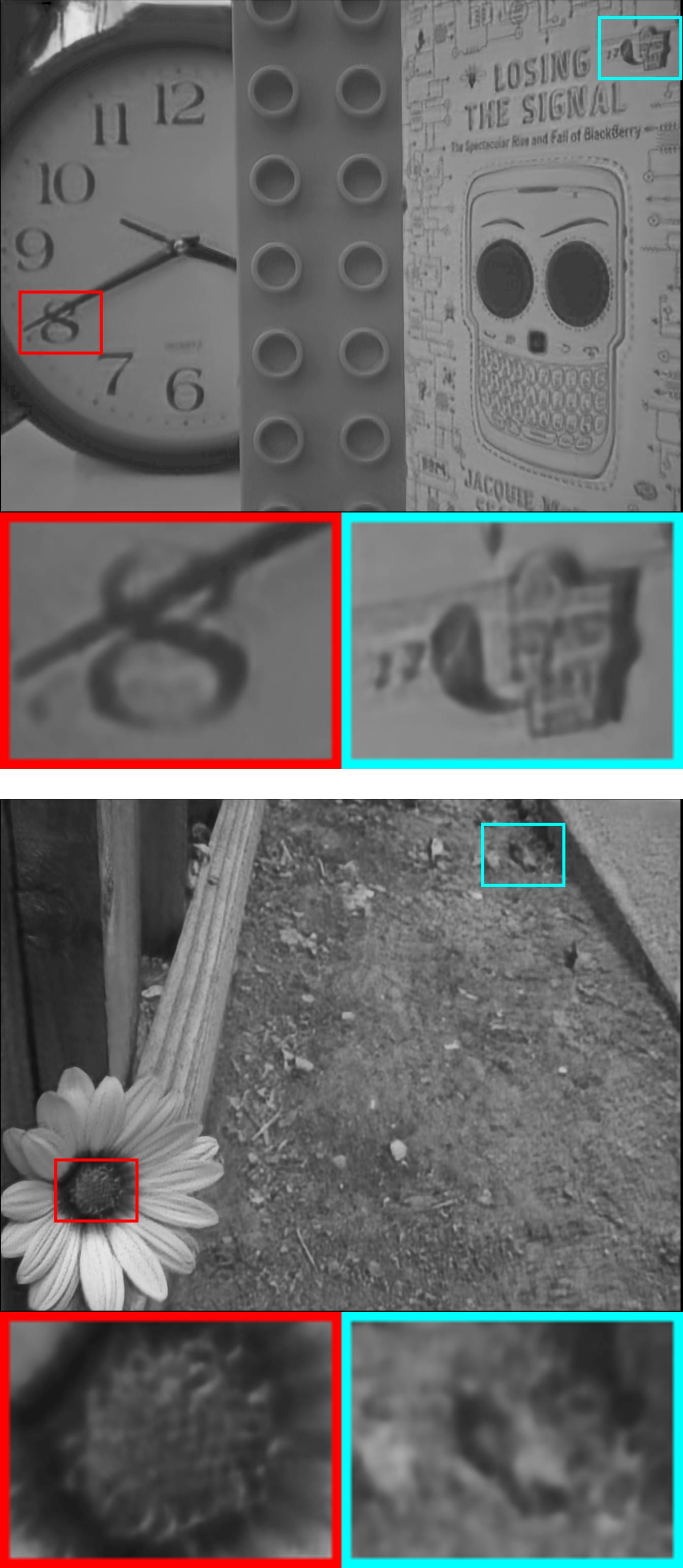} \label{subfig:sharp_im_ECCV_with_white_sheet_calibration_comparison}}
	\vspace{-0.2in}
	\caption{Qualitative comparisons of various defocus deblurring methods. Input images \subref{subfig:inputs_dd_comparison} shown as the average of two DP views, ground truth all-in-focus images  \subref{subfig:sharp_im_gt_comparison} computed from focus stacks, recovered all-in-focus images \subref{subfig:sharp_im_ours_comparison} from our method and other methods \subref{subfig:sharp_im_wiener_comparison}-\subref{subfig:sharp_im_ECCV_with_white_sheet_calibration_comparison}.
		We improve the accuracy of DPDNet~\subref{subfig:sharp_im_ECCV_comparison} trained on Canon data by providing vignetting corrected images~\subref{subfig:sharp_im_ECCV_with_white_sheet_calibration_comparison}.
		Our method performs the best in recovering high-frequency details and presents fewer artifacts.}
	\label{fig:comparison_deblurring}
	\vspace{-0.15in}
\end{figure*}

\begin{figure*}[ht]
	\centering
	\subfigure[\scriptsize Input image]{\includegraphics[width=0.105\textwidth]{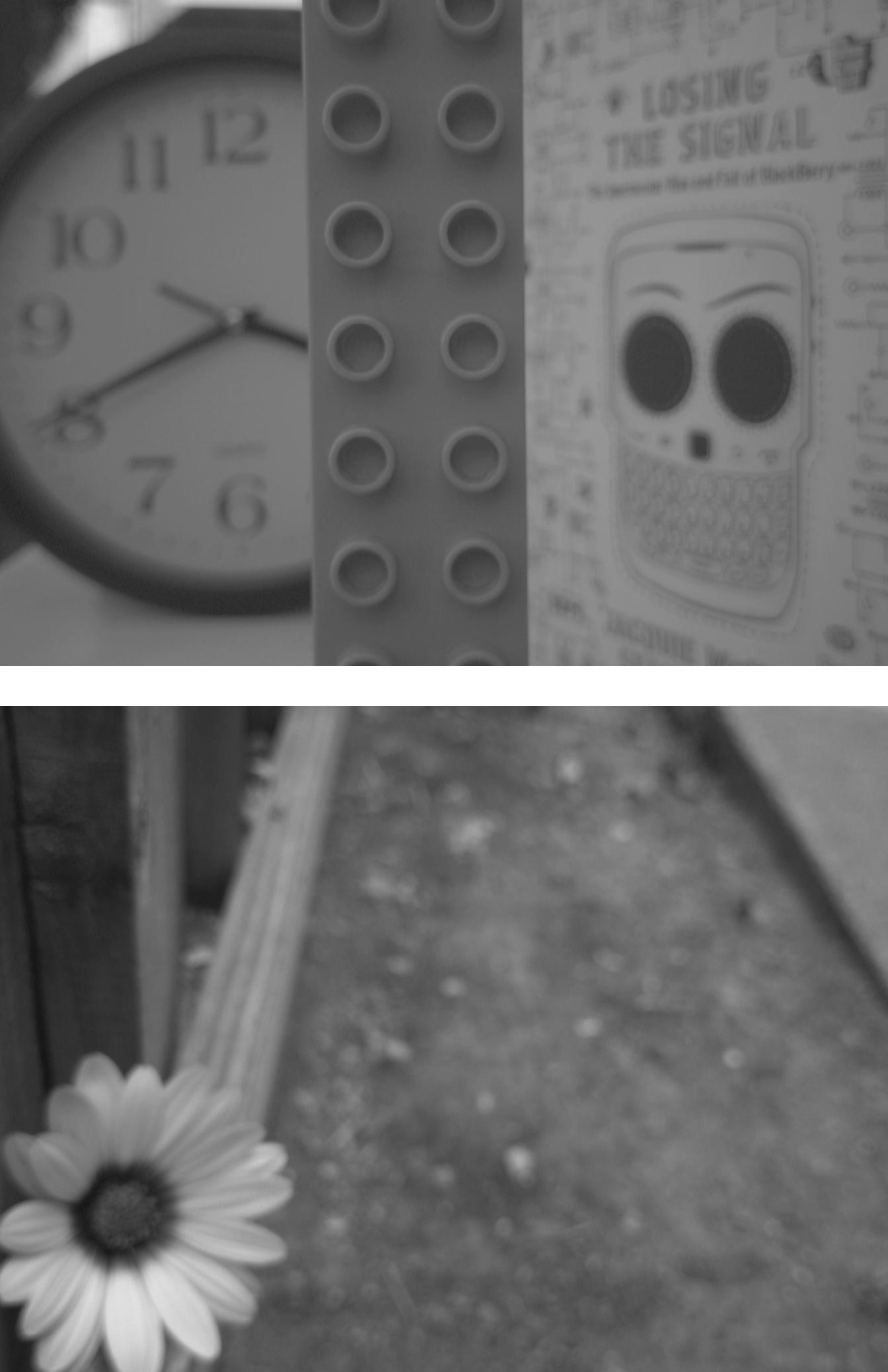}\label{subfig:inputs_defocus_comparison}}
	\subfigure[\scriptsize Ground truth]{\includegraphics[width=0.105\textwidth]{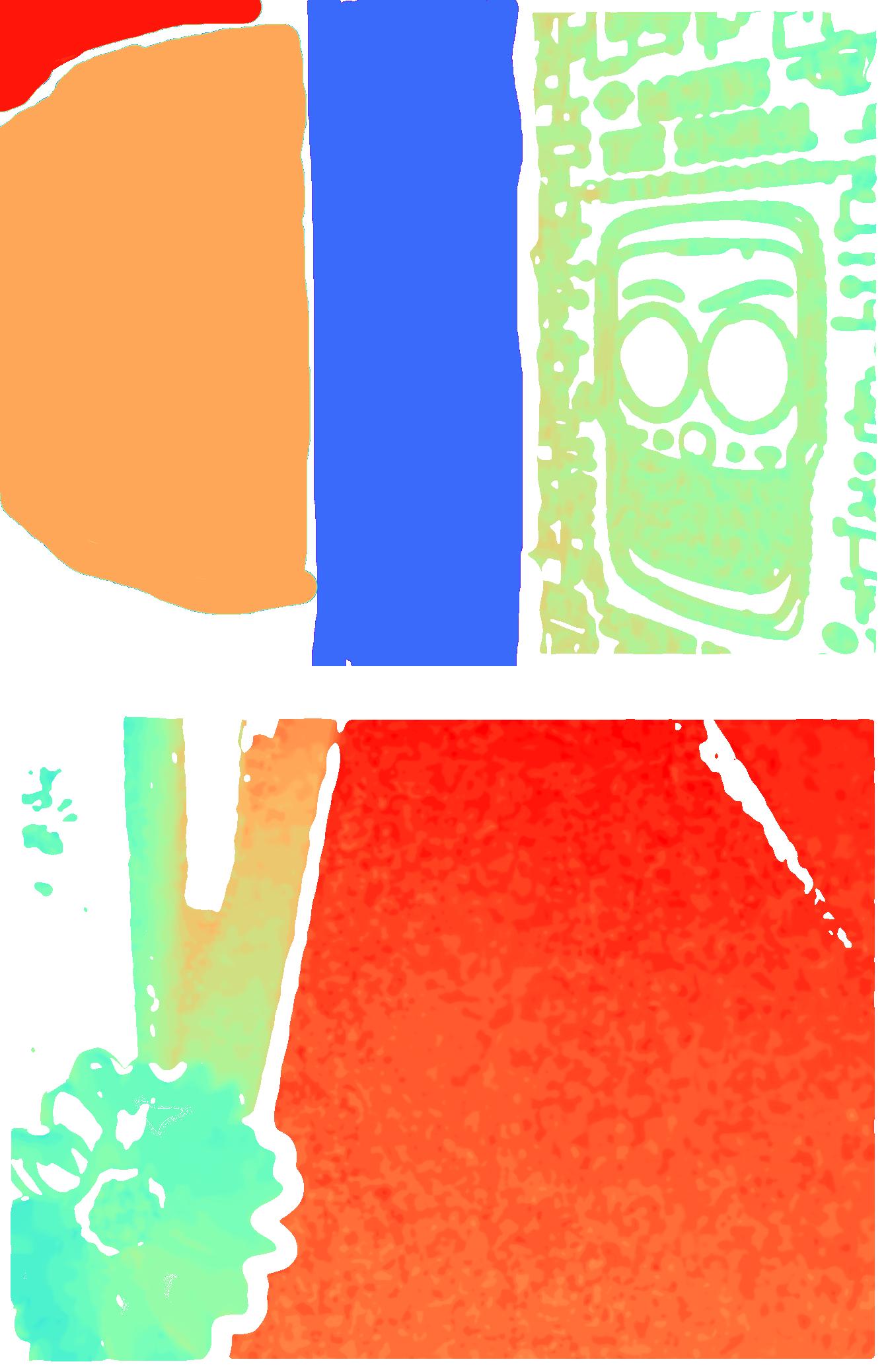}\label{subfig:defocus_gt_comparison}}
	\subfigure[\scriptsize Ours]{\includegraphics[width=0.105\textwidth]{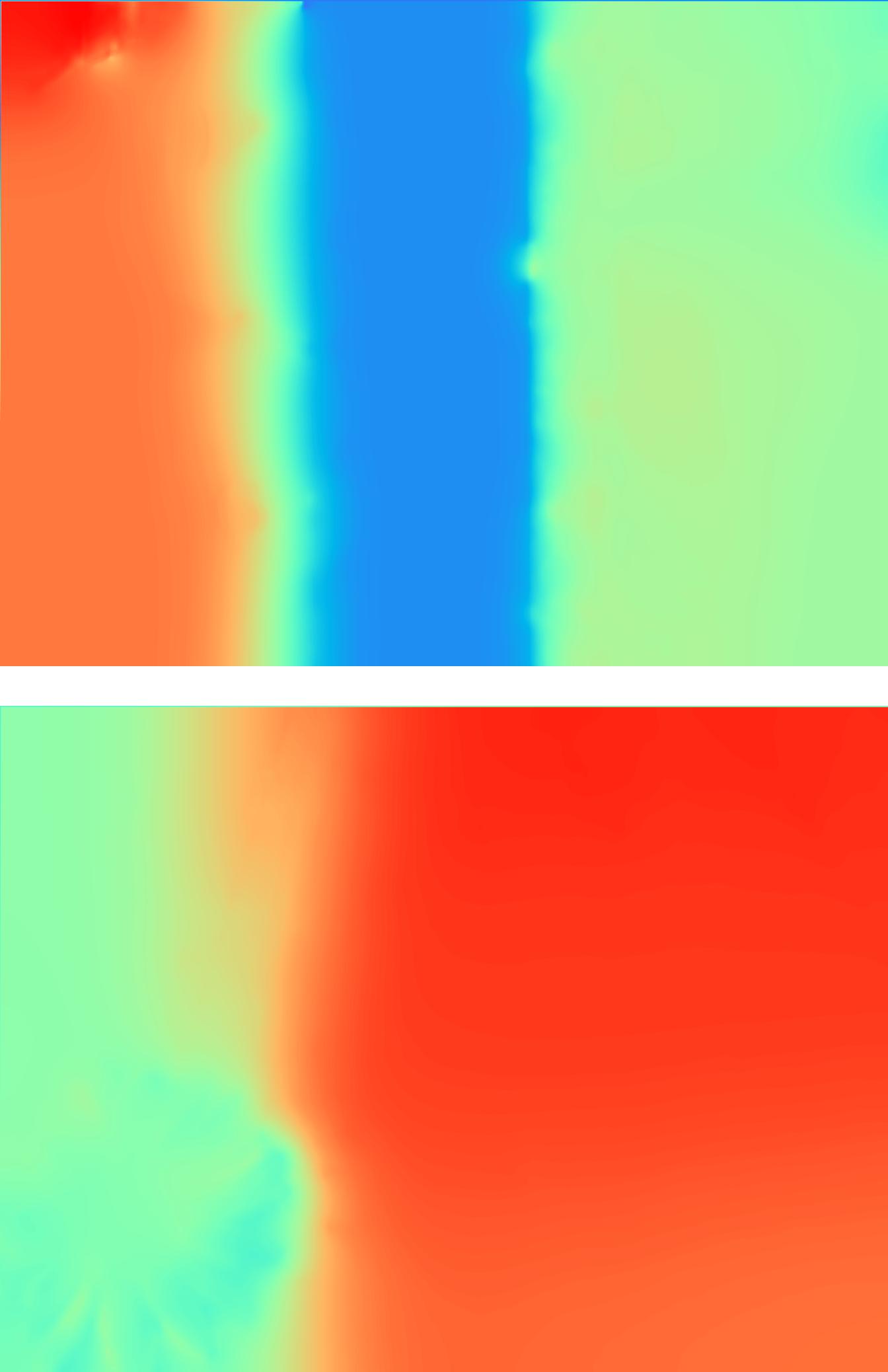}\label{subfig:defocus_ours_comparison}}
	\subfigure[\scriptsize Ours w/ GF]{\includegraphics[width=0.105\textwidth]{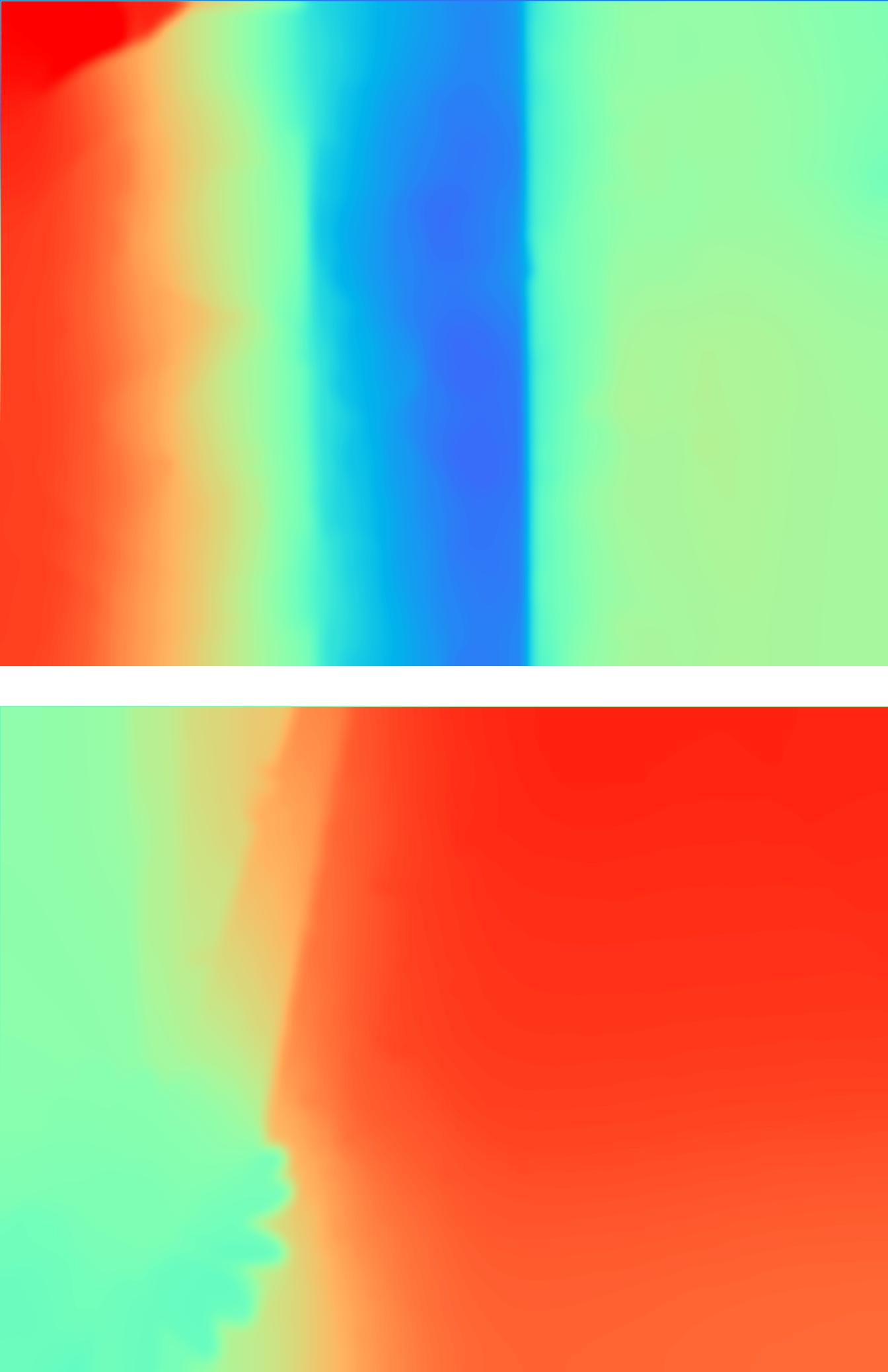}\label{subfig:defocus_ours_gf_comparison}}
	\subfigure[\scriptsize Wiener~\cite{ Zhou2009ICCV}]{\includegraphics[width=0.105\textwidth]{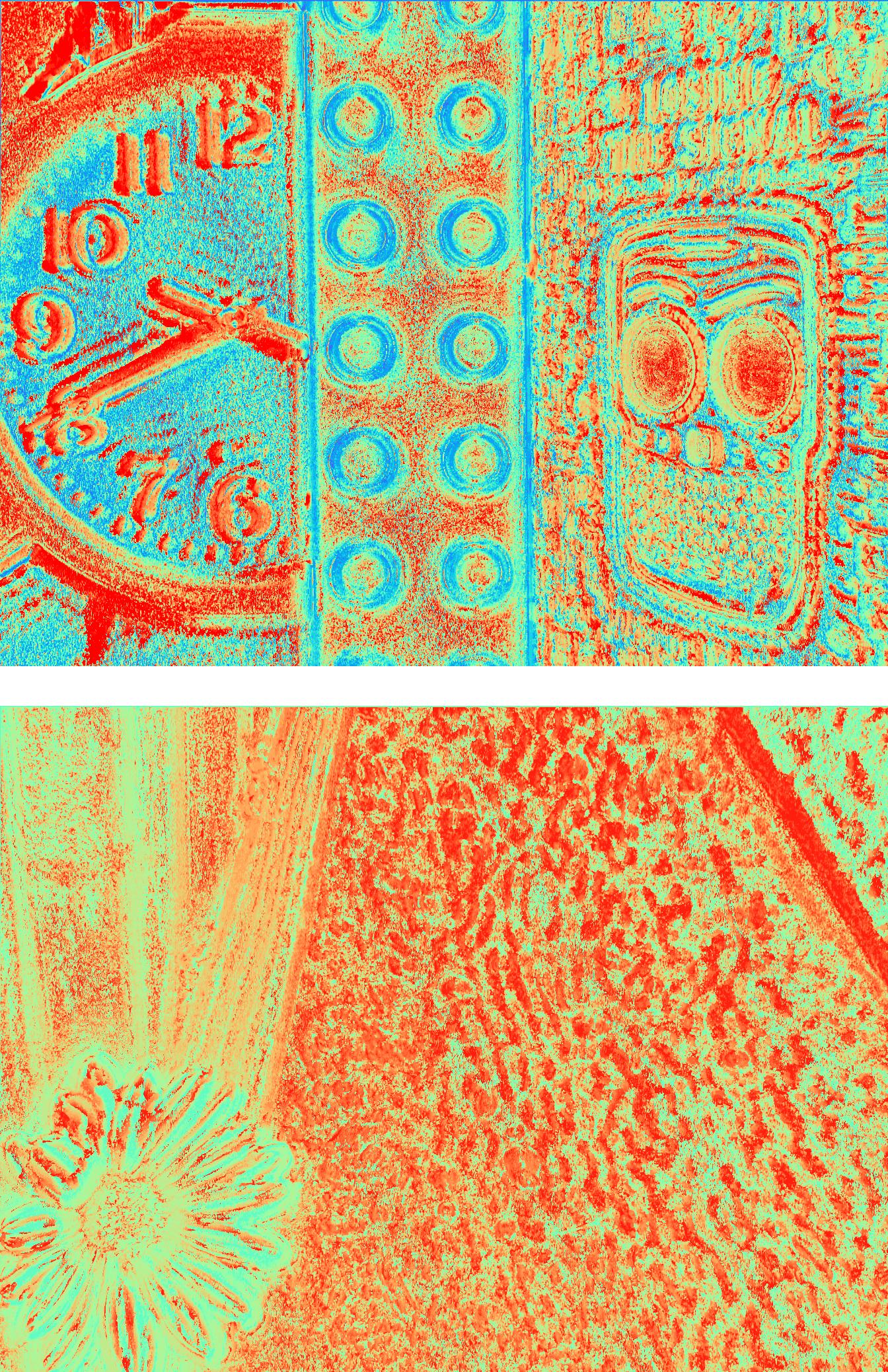}\label{subfig:defocus_wiener_comparison}}
	\subfigure[\scriptsize DMENet~\cite{Lee_2019_CVPR}]{\includegraphics[width=0.105\textwidth]{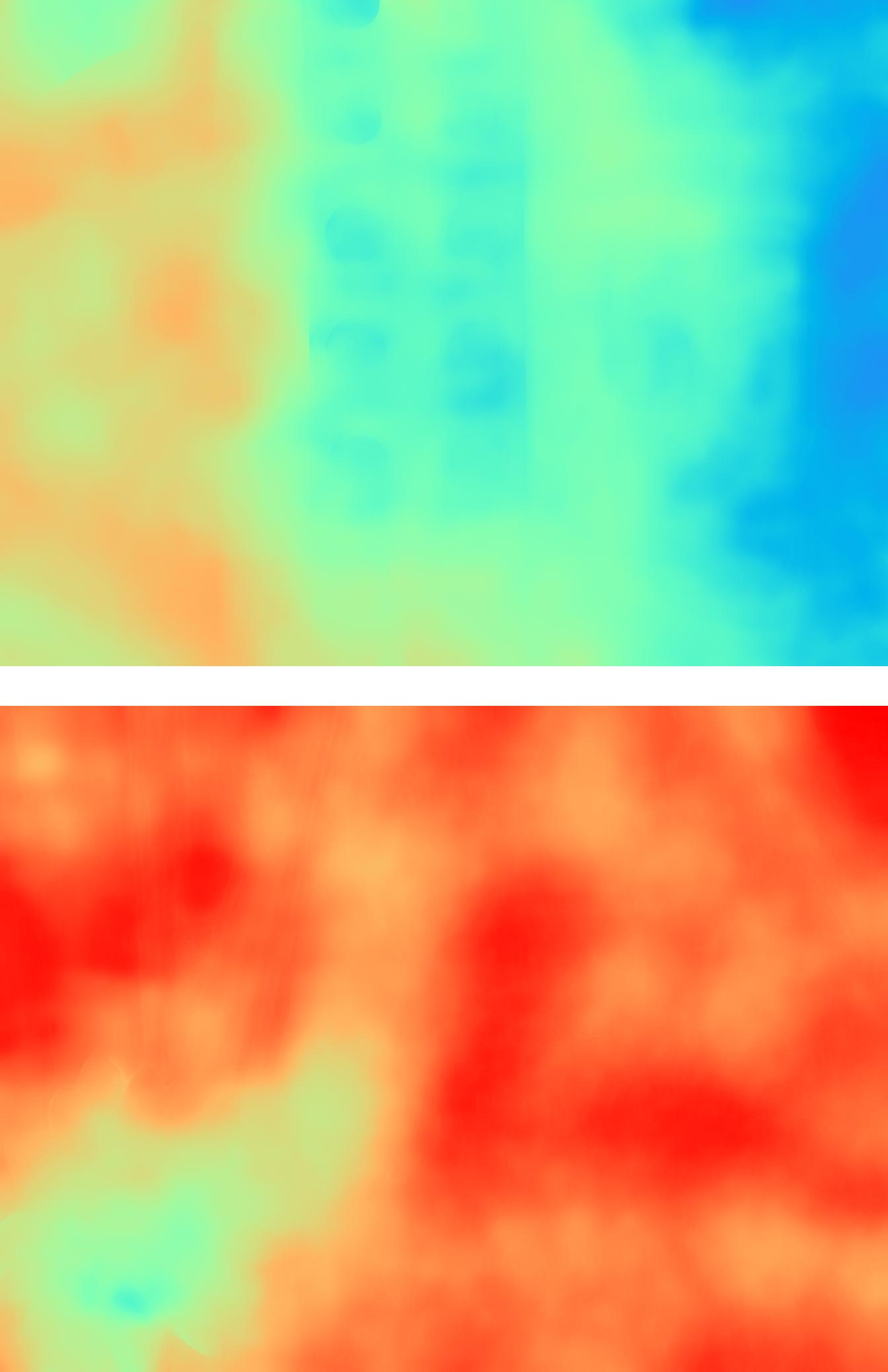}\label{subfig:depth_DMENet_comparison}}
	\subfigure[\scriptsize \cite{punnappurath2020modeling}]{\includegraphics[width=0.105\textwidth]{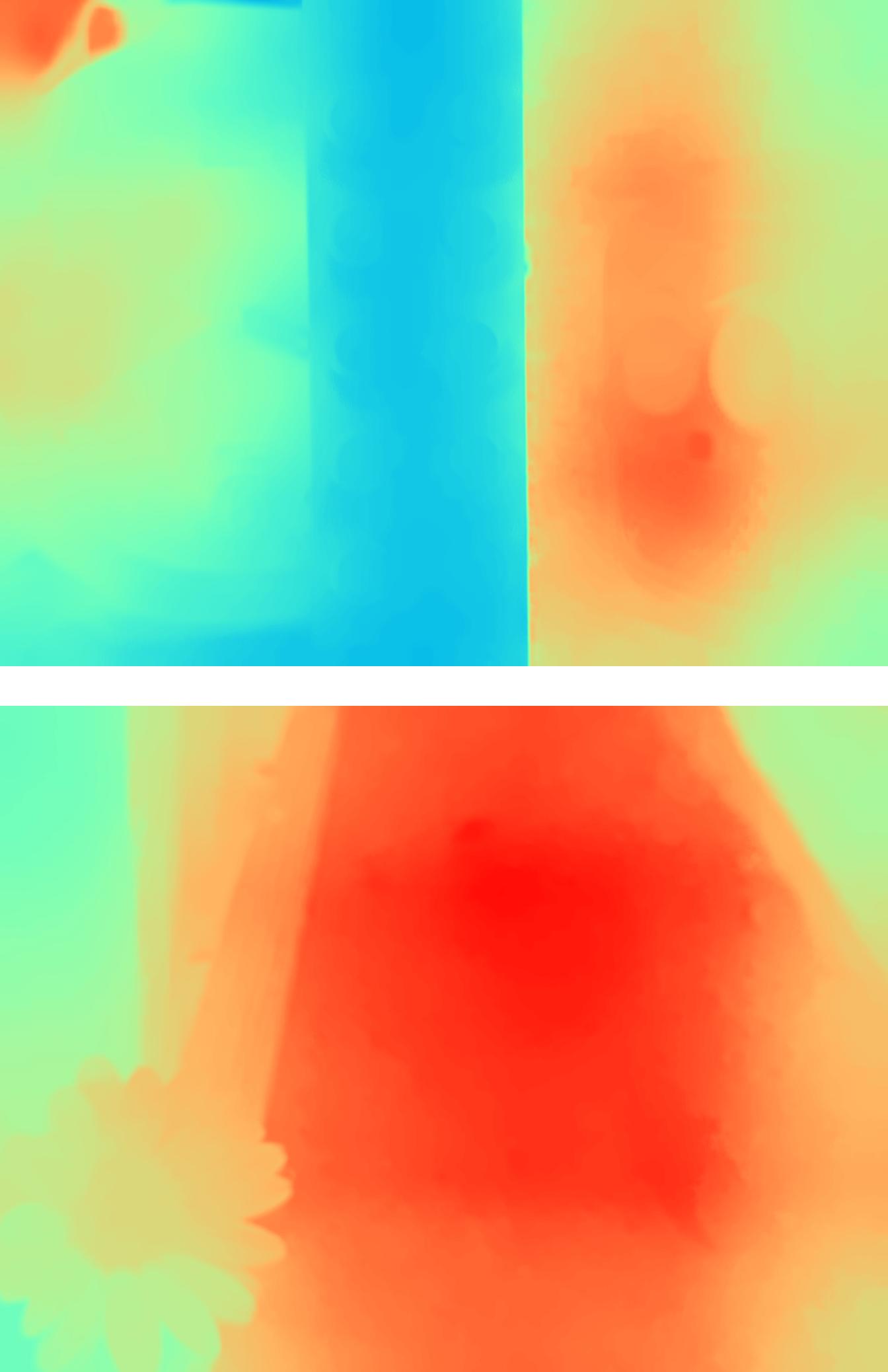}\label{subfig:defocus_ICCP_comparison}}
	\subfigure[\scriptsize Garg ~\cite{garg2019learning}]{\includegraphics[width=0.105\textwidth]{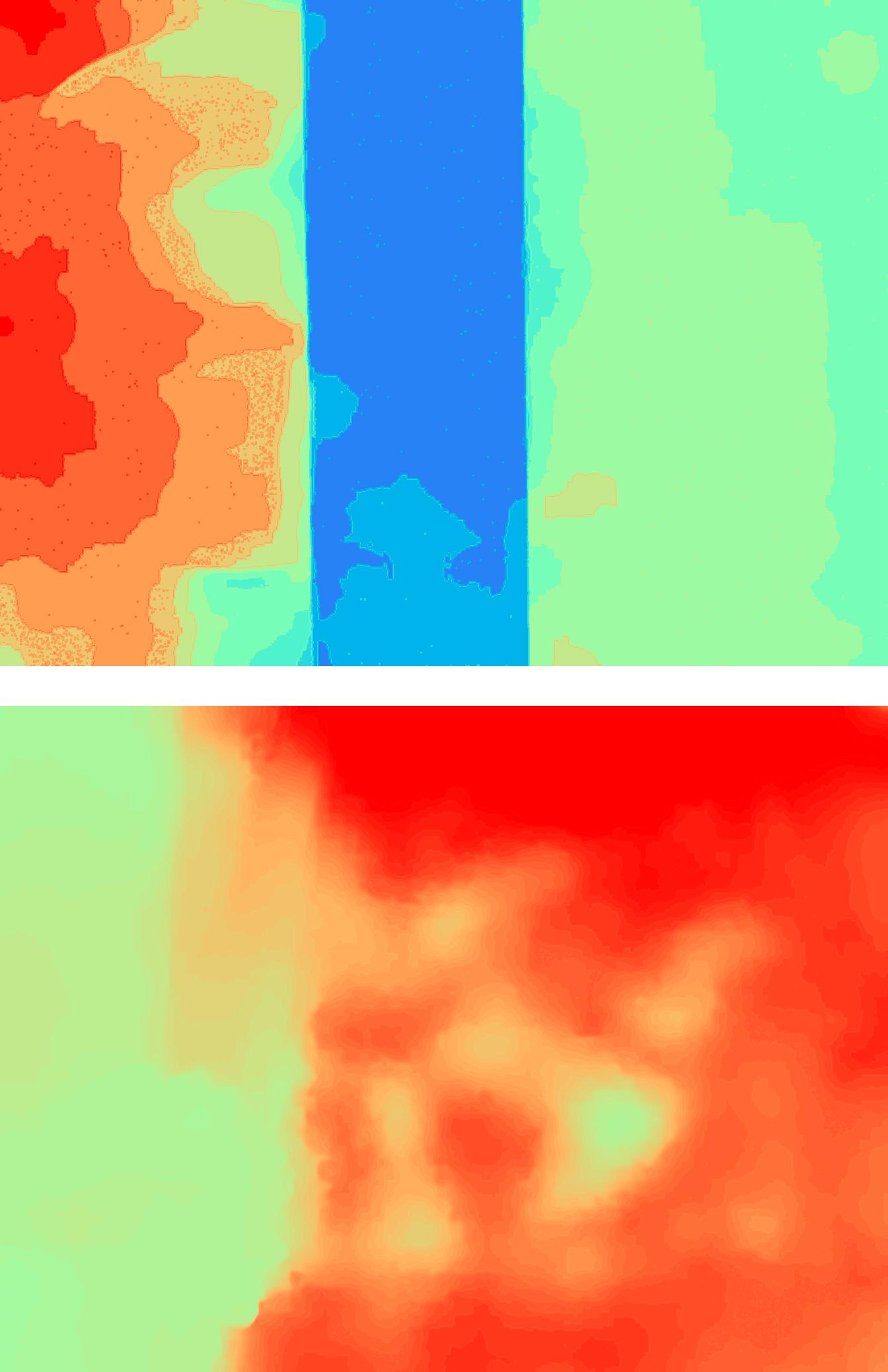}\label{subfig:defocus_ICCV_comparison}}
	\subfigure[\scriptsize Wadhwa ~\cite{wadhwa2018synthetic}]{\includegraphics[width=0.105\textwidth]{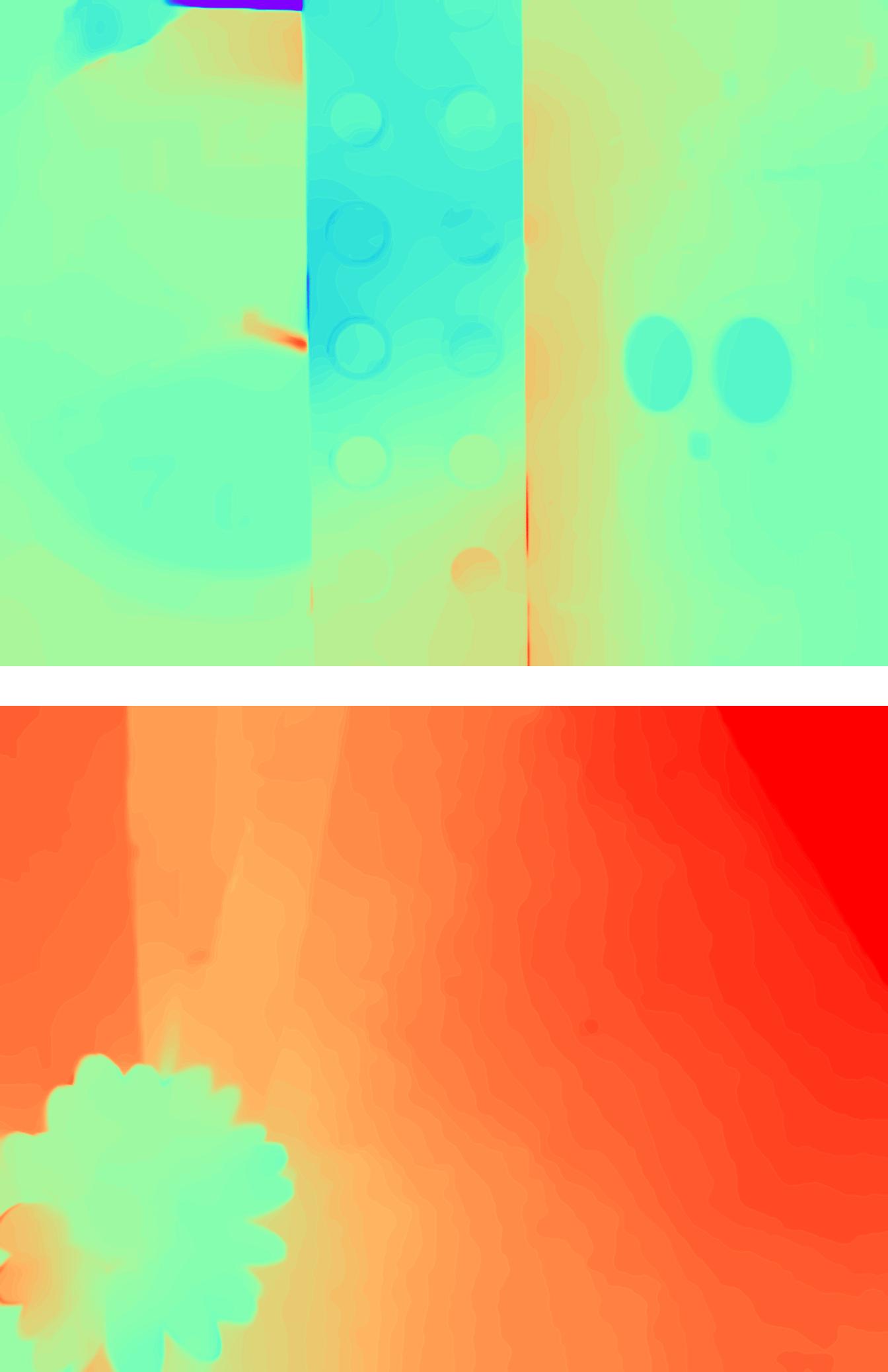}\label{subfig:defocus_Siggraph_comparison}}
	\vspace{-0.2in}
	\caption{Qualitative comparisons of defocus map estimation methods. Input images \subref{subfig:inputs_defocus_comparison} shown as the average of two DP views, ground truth defocus maps \subref{subfig:defocus_gt_comparison} from focus stacks with zero confidence pixels in white, our defocus maps \subref{subfig:defocus_ours_comparison}, and our defocus maps with guided filtering \subref{subfig:defocus_ours_gf_comparison}, and defocus maps from other methods \subref{subfig:depth_DMENet_comparison}-\subref{subfig:defocus_Siggraph_comparison}. 
		Overall, our method produces results that are closest to the ground truth, and correctly handles textureless regions as well.}
	\label{fig:comparison_defocus_map}
	\vspace{-0.15in}
\end{figure*}

We evaluate our method on both defocus deblurring and depth-from-defocus tasks. We use $N=12$ MPI layers for all scenes in our dataset. We manually determine the kernel sizes of the front and back layers, and evenly distribute layers in diopter space. 
Each optimization runs for 10,000 iterations with Adam~\cite{kingma2014adam}, and takes 2 hours on an Nvidia Titan RTX GPU. We gradually decrease the global learning rate from 0.3 to 0.1 with exponential decay. Our JAX~\cite{jax2018github} implementation is available at the project website~\cite{ProjectWebsite}.

We compare to state-of-the-art methods for defocus deblurring ( DPDNet~\cite{abuolaim2020defocus}, Wiener deconvolution~\cite{tang2012utilizing, Zhou2009ICCV}) and defocus map estimation (DP stereo matching~\cite{wadhwa2018synthetic}, supervised learning from DP views~\cite{garg2019learning}, DP defocus estimation based on kernel symmetry~\cite{punnappurath2020modeling}, Wiener deconvolution~\cite{tang2012utilizing, Zhou2009ICCV}, DMENet~\cite{Lee_2019_CVPR}).
For methods that take a single image as input, we use the average of the left and right DP images.
We also provide both the original and vignetting corrected DP images as inputs, and report the best result. 
We show quantitative results in Tab.~\ref{table:comparison} and qualitative results in Figs.~\ref{fig:comparison_deblurring} and \ref{fig:comparison_defocus_map}. 
For the defocus map, we use the affine-invariant metrics from Garg \etal~\cite{garg2019learning}. Our method achieves the best quantitative results on both tasks.

\noindent{\bf Defocus deblurring results.}
Despite the large amount of blur in the input DP images, our method produces deblurred results with high-frequency details that are close to the ground truth (Fig.~\ref{fig:comparison_deblurring}). 
DPDNet makes large errors as it is trained on Canon data and does not generalize. We improve the accuracy of DPDNet by providing vignetting corrected images as input, but its accuracy is still lower than ours. 

\noindent{\bf Defocus map estimation results.}
Our method produces defocus maps that are closest to the ground truth (Fig.~\ref{fig:comparison_defocus_map}), especially on textureless regions, such as the toy and clock in the first scene. Similar to \cite{punnappurath2020modeling}, depth accuracy near edges can be improved by guided filtering~\cite{He10guided} as shown in Fig.~\ref{subfig:defocus_ours_gf_comparison}.

\begin{figure*}[ht]
	\centering
	\subfigure[Input image]{\includegraphics[width=0.12\textwidth]{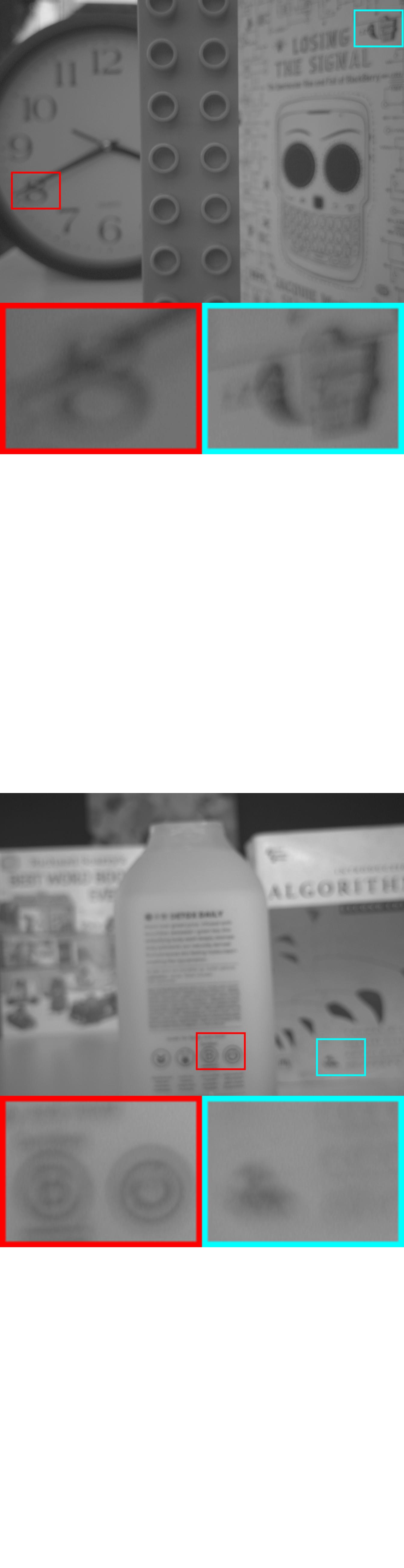}\label{subfig:inputs_combined_ablation}}
	\subfigure[Ground truth]{\includegraphics[width=0.12\textwidth]{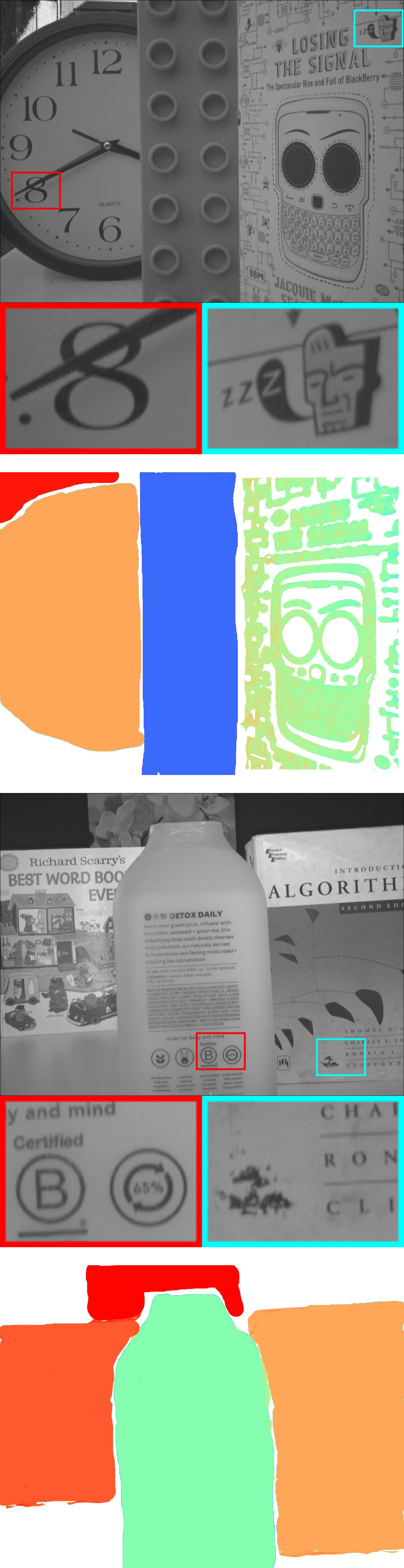}\label{subfig:combined_gt_ablation}}
	\subfigure[Ours full]{\includegraphics[width=0.12\textwidth]{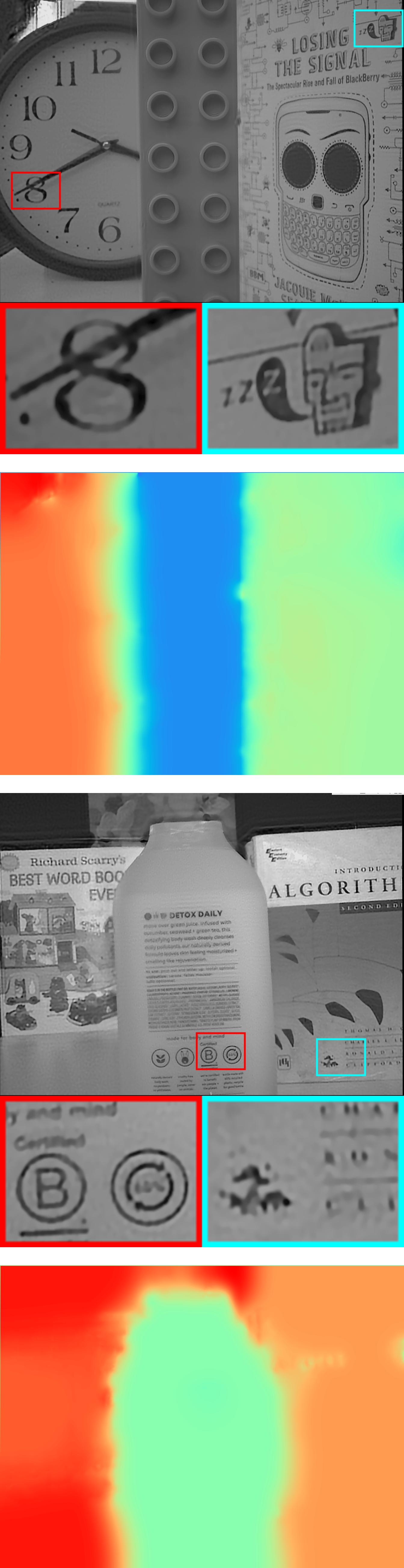}\label{subfig:combined_ours_ablation}}
	\subfigure[No $\loss_{\mathrm{intensity}}$]{\includegraphics[width=0.12\textwidth]{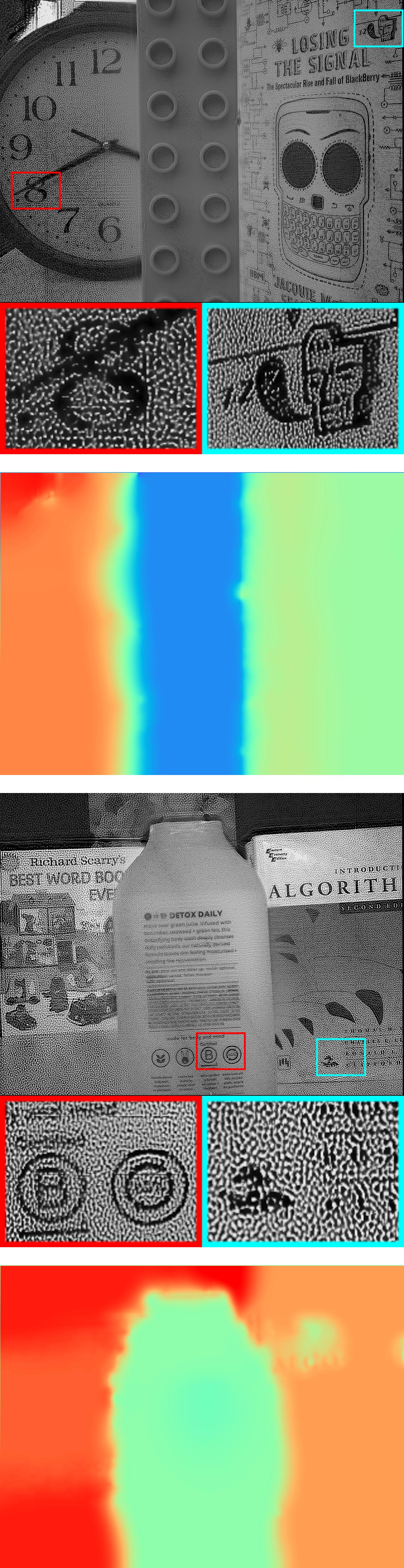}\label{subfig:combined_ours_no_im_tv_ablation}}
	\subfigure[No $\loss_{\mathrm{alpha}}$]{\includegraphics[width=0.12\textwidth]{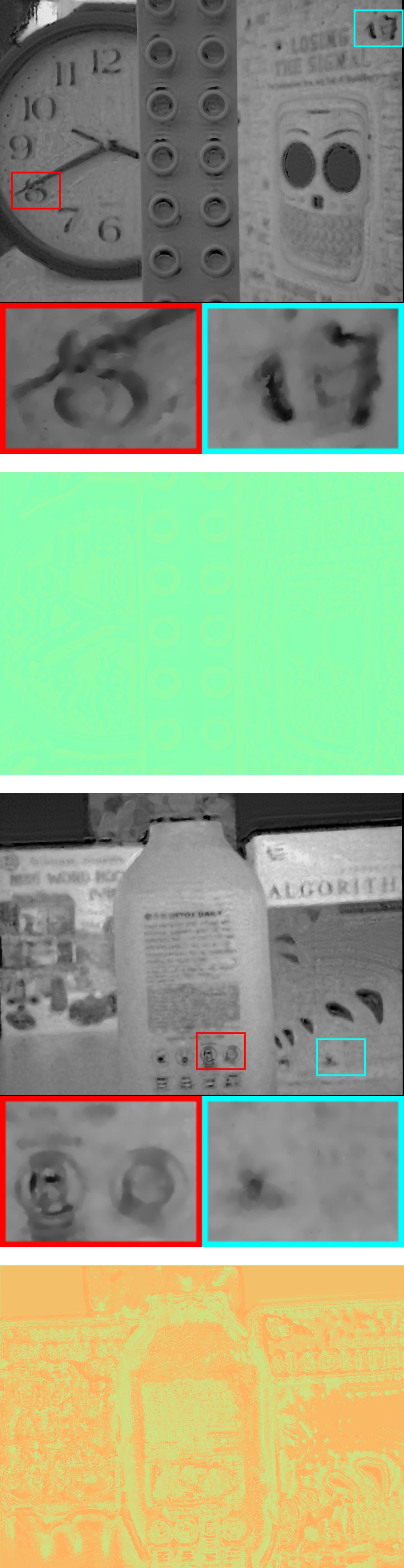}\label{subfig:combined_ours_no_alpha_tv_ablation}}
	\subfigure[No $\loss_{\mathrm{entropy}}$]{\includegraphics[width=0.12\textwidth]{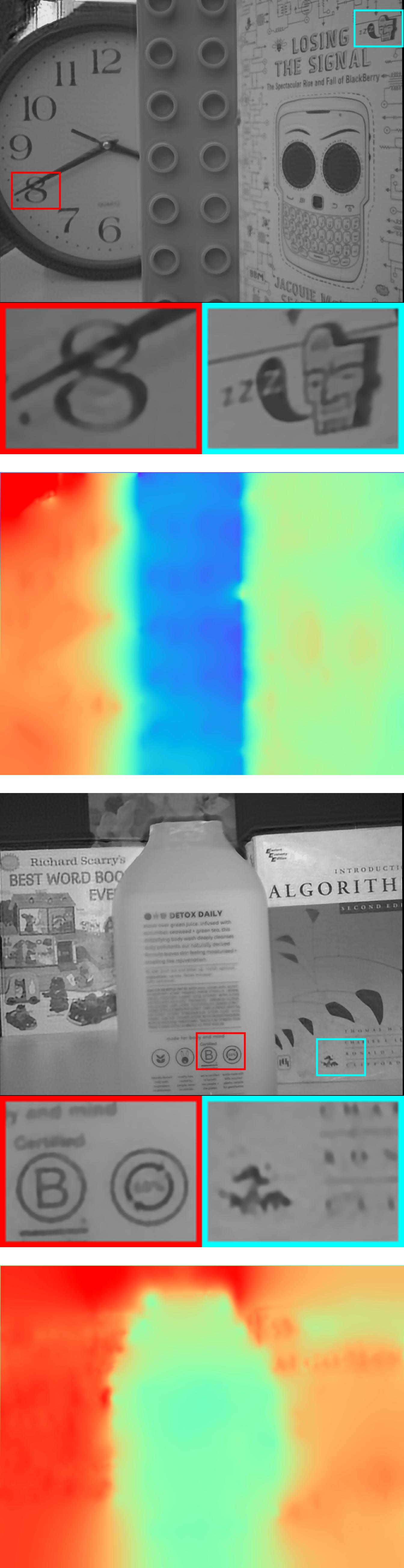}\label{subfig:combined_ours_no_entropy_ablation}}
	\subfigure[No $\loss_{\mathrm{aux}}$]{\includegraphics[width=0.12\textwidth]{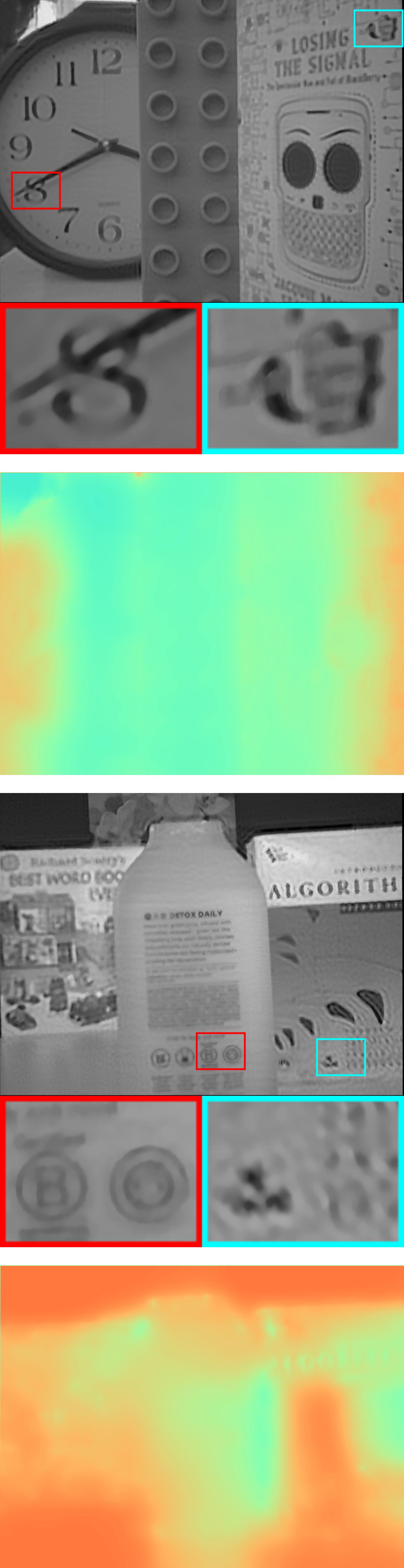}\label{subfig:combined_ours_no_auxiliary_ablation}}
	\subfigure[No $\bias$]{\includegraphics[width=0.12\textwidth]{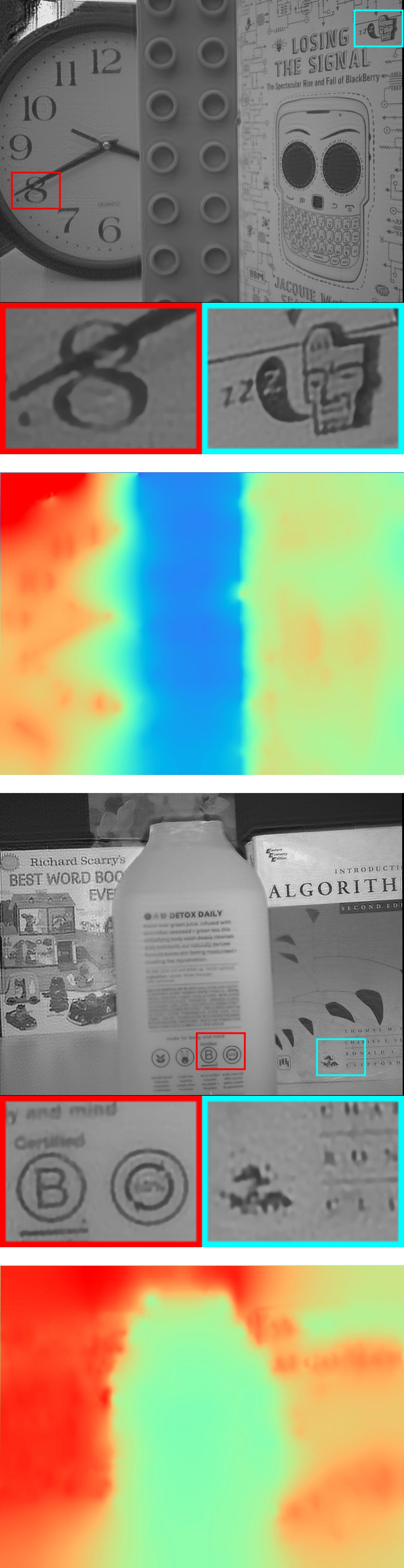}\label{subfig:combined_ours_no_bias_correction_ablation}}
	\vspace{-0.2in}
	\caption{Ablation study. Input images \subref{subfig:inputs_combined_ablation}, ground truth all-in-focus images, and defocus maps \subref{subfig:combined_gt_ablation} with zero confidence pixels in white, our results \subref{subfig:combined_ours_ablation}, and our results with different terms removed one at a time \subref{subfig:combined_ours_no_im_tv_ablation}-\subref{subfig:combined_ours_no_bias_correction_ablation}.
	Removing $\loss_{\mathrm{intensity}}$ and $\loss_{\mathrm{alpha}}$ strongly affects the smoothness of all-in-focus images and defocus maps respectively. Results without entropy regularization $\loss_{\mathrm{entropy}}$, $\loss_{\mathrm{aux}}$, or the bias correction $\bias$, exhibit more errors in defocus maps on textureless regions (clock).
	}
	\label{fig:ablation_combined}
	\vspace{-0.15in}
\end{figure*}

\noindent{\bf Ablation studies.}
We investigate the effect of each loss function term by removing them one at a time. Quantitative results are in Tab.~\ref{table:ablation}, and qualitative comparisons in Fig.~\ref{fig:ablation_combined}.

Our full pipeline has the best overall performance in recovering all-in-focus images and defocus maps. $\loss_{\mathrm{intensity}}$ and $\loss_{\mathrm{alpha}}$ strongly affect the smoothness of all-in-focus images and defocus maps, respectively. 
Without $\loss_{\mathrm{entropy}}$ or $\loss_{\mathrm{aux}}$, even though recovered all-in-focus images are reasonable, scene content is smeared across multiple MPI layers, leading to incorrect defocus maps.
Finally, without the bias correction term $\bias$, defocus maps are biased towards smaller blur radii, especially in textureless areas where noise is more apparent, e.g., the white clock area.

\begin{figure}[t]
	\captionsetup[subfigure]{font=scriptsize,labelfont=scriptsize}
	\centering
	\subfigure[Input from \cite{abuolaim2020defocus}]{\includegraphics[width=0.11\textwidth]{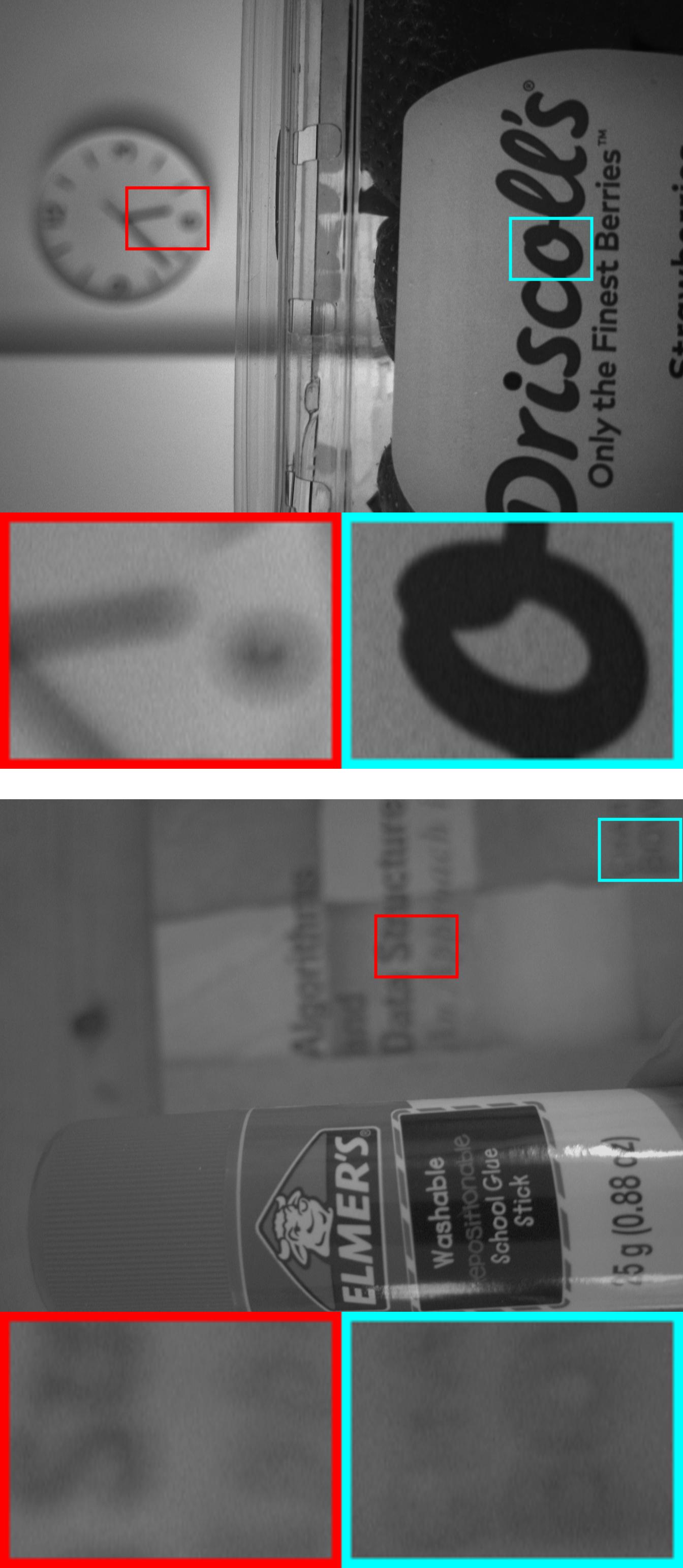} \label{subfig:ECCV_data_inputs_comparison}}
	\subfigure[DPDNet \cite{abuolaim2020defocus}]{\includegraphics[width=0.11\textwidth]{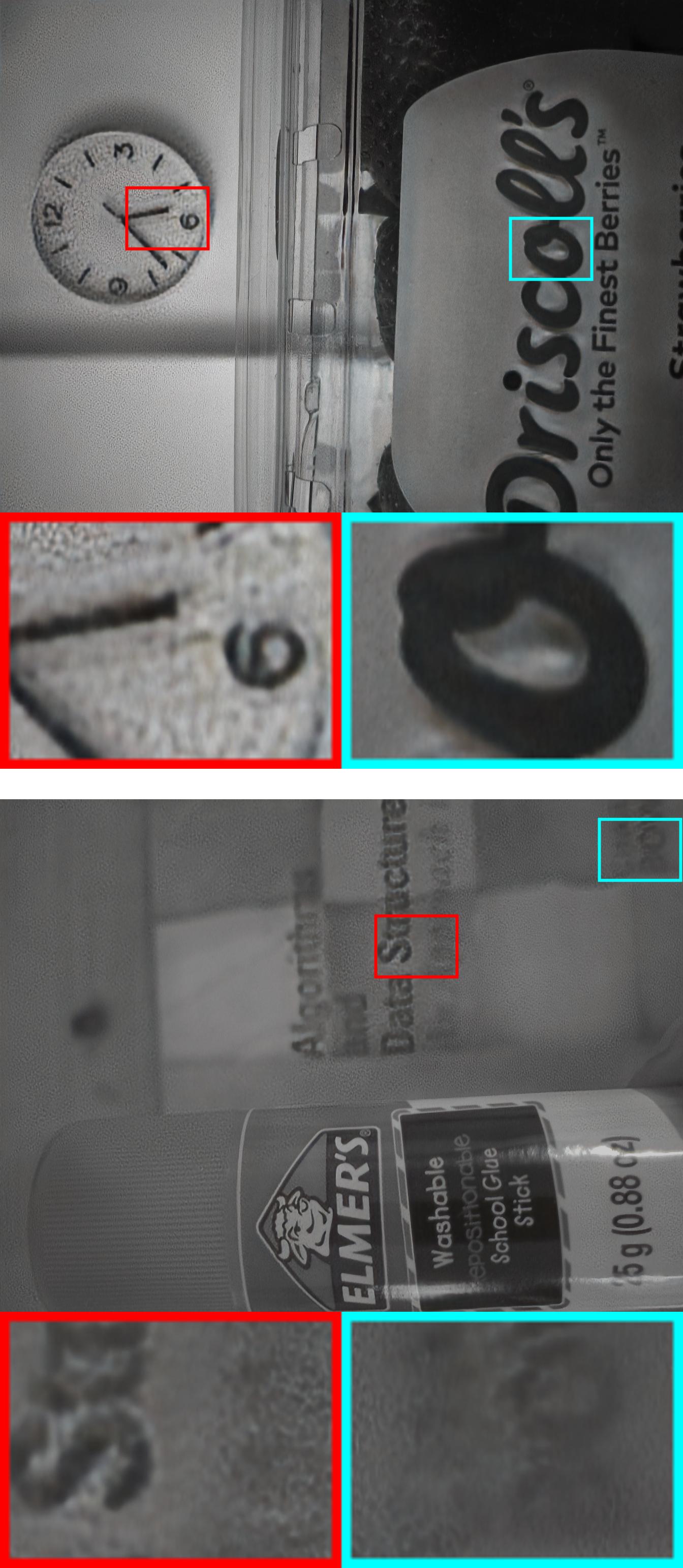} \label{subfig:ECCV_data_sharp_im_DPDNet}}
	\subfigure[Our results]{
		\includegraphics[width=0.11\textwidth]{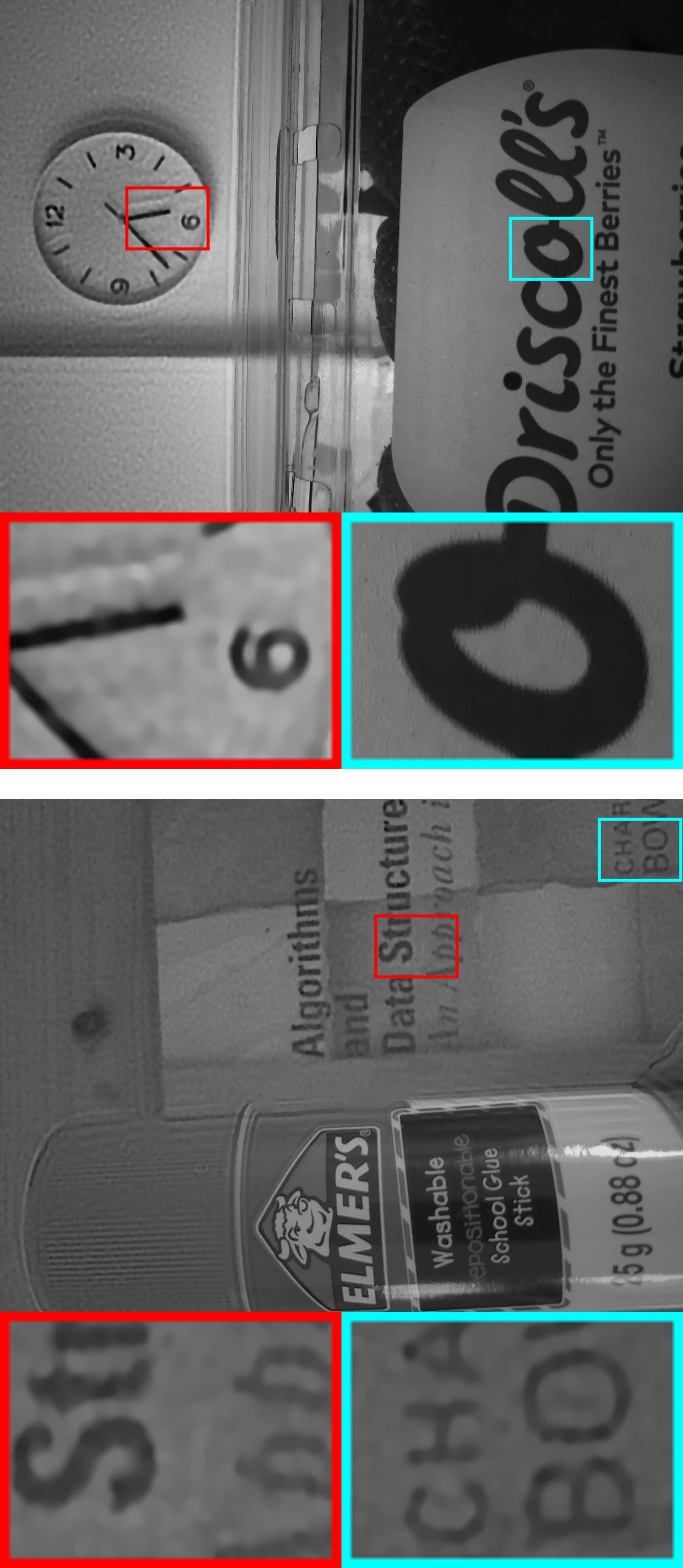}
		\hfill
		\includegraphics[width=0.11\textwidth]{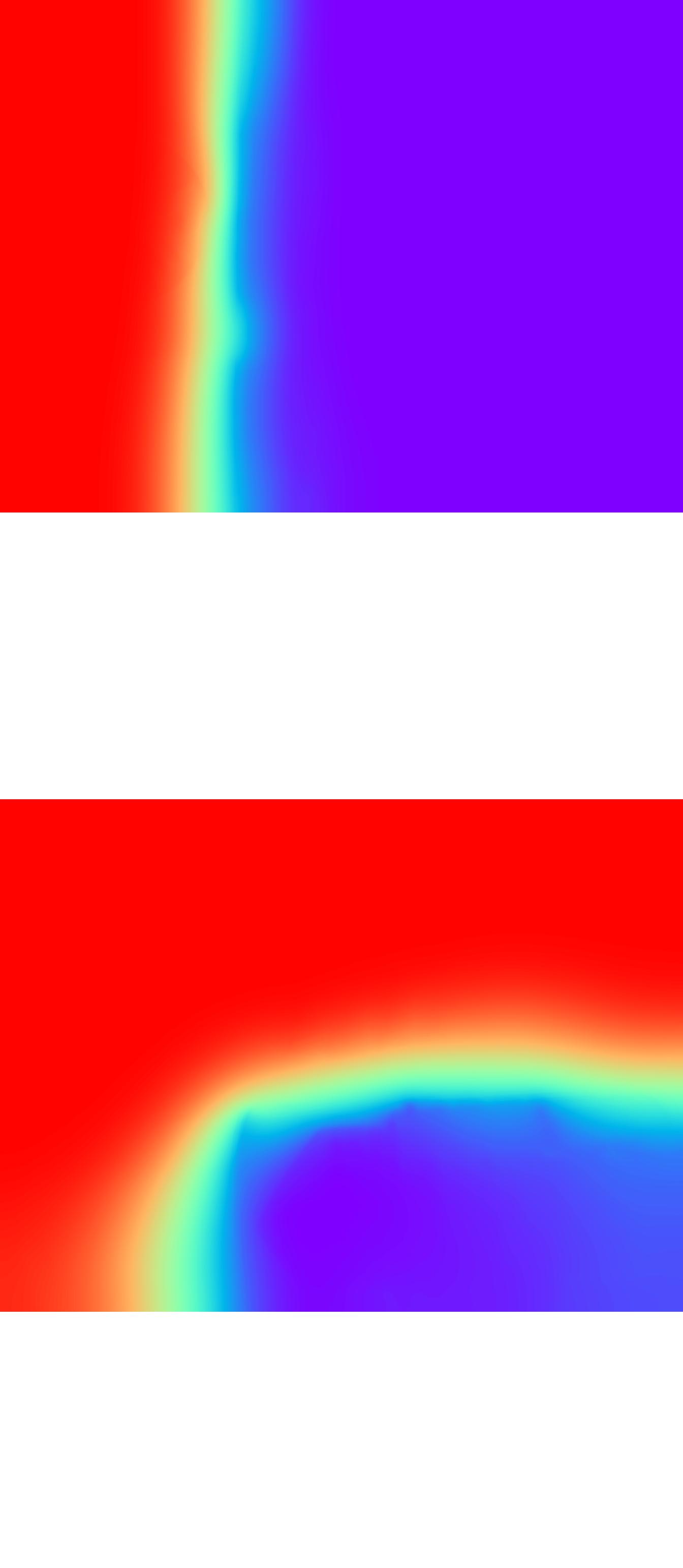}
		\label{subfig:ECCV_data_ours}}
	\vspace{-0.2in}
	\caption{
		Results on data from~\cite{abuolaim2020defocus}. Our method recovers all-in-focus images with fewer artifacts, while using the calibration data from our device.
	}
	\vspace{-0.3in}
	\label{fig:ECCV_data}
\end{figure}

\noindent{\bf Results on Data from Abuolaim and Brown~\cite{abuolaim2020defocus}.}
Even though Abuolaim and Brown~\cite{abuolaim2020defocus} train their model on data from a Canon camera, they also capture Pixel 4 data for qualitative tests. We run our method on their Pixel 4 data, using the calibration from our device, and show that our recovered all-in-focus image has fewer artifacts (Fig.~\ref{fig:ECCV_data}). This demonstrates that our method generalizes well across devices of the same model, even without re-calibration.

\vspace{-0.1in}
\section{Discussion and Conclusion}

\begin{table}[t]
	\centering
	\resizebox{\linewidth}{!}{
		\begin{tabular}{l|ccc|ccc}
			\multirow{2}{*}{Method} & \multicolumn{3}{c|}{All-in-focus Image} & \multicolumn{3}{c}{Defocus Map}\\
			& $\PSNR\uparrow$ & $\SSIM\uparrow$ & $\MAE\downarrow$ & $\AIWE(1)\downarrow$ & $\AIWE(2)\downarrow$ & $1 - \abs{\rho_s}\downarrow$ \\
			\hline
			Wiener Deconv.~\cite{Zhou2009ICCV} & 25.806 & 0.704 & 0.032 & 0.156 & 0.197 & 0.665\\
			DPDNet ~\cite{abuolaim2020defocus} & 25.591 & 0.777 & 0.034 & - & - & -\\
			DMENet~\cite{Lee_2019_CVPR} & - & - & - & 0.144 & 0.183 & 0.586\\
			Punnappurath \etal~\cite{punnappurath2020modeling} & - & - & - & 0.124 & 0.161 & 0.444\\
			Garg \etal~\cite{garg2019learning} & - & - & - & 0.079 & 0.102 & 0.208\\
			Wadhwa \etal~\cite{wadhwa2018synthetic} & - & - & - & 0.141 & 0.177 & 0.540\\
			Ours & {\cellcolor{lightred} \centering} 26.692 & \cellcolor{lightred} 0.804 &\cellcolor{lightred} 0.027 & \cellcolor{lightred} 0.047 & \cellcolor{lightred} 0.076 & \cellcolor{lightred} 0.178\\
			Ours w/ guided filtering & 26.692 & 0.804 & 0.027 & 0.059 & 0.083 & 0.193\\
		\end{tabular}
	}
	\vspace{-0.1in}
	\caption{Quantitative evaluations of defocus deblurring and defocus map estimation methods on our DP dataset. ``-'' indicates not applicable. We use the affine-invariant metrics from~\cite{garg2019learning} for defocus map evaluation. Our method achieves the best performance (highlighted in red) in both tasks. }
	\label{table:comparison}
	\vspace{-0.15in}
\end{table}

We presented a method that optimizes an MPI scene representation to jointly recover a defocus map and all-in-focus image from a single dual-pixel capture. We showed that image noise introduces a bias in the optimization that, under suitable assumptions, can be quantified and corrected for. We also introduced additional priors to regularize the optimization, and showed their effectiveness via ablation studies. Our method improves upon past work on both defocus map estimation and blur removal, when evaluated on a new dataset we captured with a consumer smartphone camera.

\noindent{\bf Limitations and future directions.} We discuss some limitations of our method, which suggest directions for future research. First, our method does not require a large dataset with ground truth to train on, but still relies on a one-time blur kernel calibration procedure. It would be interesting to explore blind deconvolution techniques~\cite{fergus2006removing,levin2011understanding} that can simultaneously recover the all-in-focus image, defocus map, and unknown blur kernels, thus removing the need for kernel calibration. The development of parametric blur kernel models that can accurately reproduce the features we observed (i.e., spatial variation, lack of symmetry, lack of circularity) can facilitate this research direction. Second, the MPI representation discretizes the scene into a set of fronto-parallel depth layers. This can potentially result in discretization artifacts in scenes with continuous depth variation. In practice, we did not find this to be an issue, thanks to the use of the soft-blending operation to synthesize the all-in-focus image and defocus map.
Nevertheless, it could be useful to replace the MPI representation with a continuous one, e.g., neural radiance fields~\cite{mildenhall2020nerf}, to help better model continuously-varying depth. Third, reconstructing an accurate all-in-focus image becomes more difficult as defocus blur increases (e.g., very distant scenes at non-infinity focus) and more high-frequency content is missing from the input image. This is a fundamental limitation shared among all deconvolution techniques. Using powerful data-driven priors to hallucinate the missing high frequency content (e.g., deep-learning-based deconvolution techniques) can help alleviate this limitation. Fourth, the high computational complexity of our technique makes it impractical for real-time operation, especially on resource-constrained devices such as smartphones. Therefore, it is worth exploring optimized implementations.

\begin{table}
	\centering
	\resizebox{\linewidth}{!}{
		\begin{tabular}{l|ccc|ccc}
			\multirow{2}{*}{Method} & \multicolumn{3}{c|}{All-in-focus Image} & \multicolumn{3}{c}{Defocus Map}\\
			& $\PSNR\uparrow$ & $\SSIM\uparrow$ & $\MAE\downarrow$ & $\AIWE(1)\downarrow$ & $\AIWE(2)\downarrow$ & $1 - \abs{\rho_s}\downarrow$ \\
			\hline
			Full & \cellcolor{orange} 26.692 & \cellcolor{orange} 0.804 & \cellcolor{orange} 0.027 & \cellcolor{lightred} 0.047 & \cellcolor{lightred} 0.076 & \cellcolor{lightred} 0.178\\
			No $\loss_{\mathrm{intensity}}$ & 14.882 & 0.158 & 0.136 & \cellcolor{lightred} 0.047 & \cellcolor{orange} 0.078 & \cellcolor{orange} 0.185\\
			No $\loss_{\mathrm{alpha}}$ & 24.748 & 0.726 & 0.037 & 0.161 & 0.206 & 0.795\\
			No $\loss_{\mathrm{entropy}}$ & \cellcolor{lightred} 27.154 & \cellcolor{lightred} 0.819 & \cellcolor{lightred} 0.026 & 0.057 & 0.085 & 0.190\\
			No $\loss_{\mathrm{aux}}$ & 26.211 & 0.768 & 0.030 & 0.148 & 0.190 & 0.610\\
			No $\bias$ & 26.265 & 0.790 & 0.028 & 0.063 & 0.092 & 0.214\\
		\end{tabular}
	}
	\vspace{-0.1in}
	\caption{Quantitative comparisons of ablation studies. We compare the full pipeline with removals of the regularization terms $\loss_{\mathrm{alpha}}$,  $\loss_{\mathrm{intensity}}$ and $\loss_{\mathrm{entropy}}$, the auxiliary data loss $\loss_{\mathrm{aux}}$, and bias correction term $\bias$ respectively. For all ablation experiments, we set the weights on remaining terms to be the same as the ones in the full pipeline. Best and second best results are highlighted in red and orange.
	}
	\label{table:ablation}
	\vspace{-0.16in}
\end{table}

\noindent{\bf Acknowledgments.} We thank David Salesin and Samuel Hasinoff for helpful feedback. S.X. and I.G. were supported by NSF award 1730147 and a Sloan Research Fellowship.

\vspace{-0.1in}
{\small
\bibliographystyle{ieee_fullname}
\bibliography{ref}
}

\clearpage
\onecolumn
\appendix

\section{Introduction}

In this supplementary material, we cover the following topics:
\begin{enumerate}
	\item In Sec.~\ref{sec:dualpixel_supp}, we describe blur kernel calibration in more detail, and explore how blur kernels change with respect to scene depth and focus distance. 
	\item In Sec.~\ref{sec:method_details}, we provide more technical details about our method. More specifically, we explain how we render defocus maps from the multi-plane image (MPI) representation, provide the derivation of the bias correction term, and define the total variation function $\tv(\cdot)$ and the edge map $\edge$ used in the regularization terms.
	\item In Sec.~\ref{sec:experiments}, we provide additional implementation details, and show comparison results and ablation studies on more data in our collected Google Pixel 4 dataset. {\bf To facilitate comparisons, we also provide an interactive HTML viewer~\cite{bitterli2017reversible} at the project website~\cite{ProjectWebsite}.}
	
\end{enumerate}

\vspace{-0.1in}
\section{Blur Kernel Calibration \label{sec:dualpixel_supp}}

\begin{figure}[!h]
	\centering
	\subfigure[Captured image]{\includegraphics[width=0.23\textwidth]{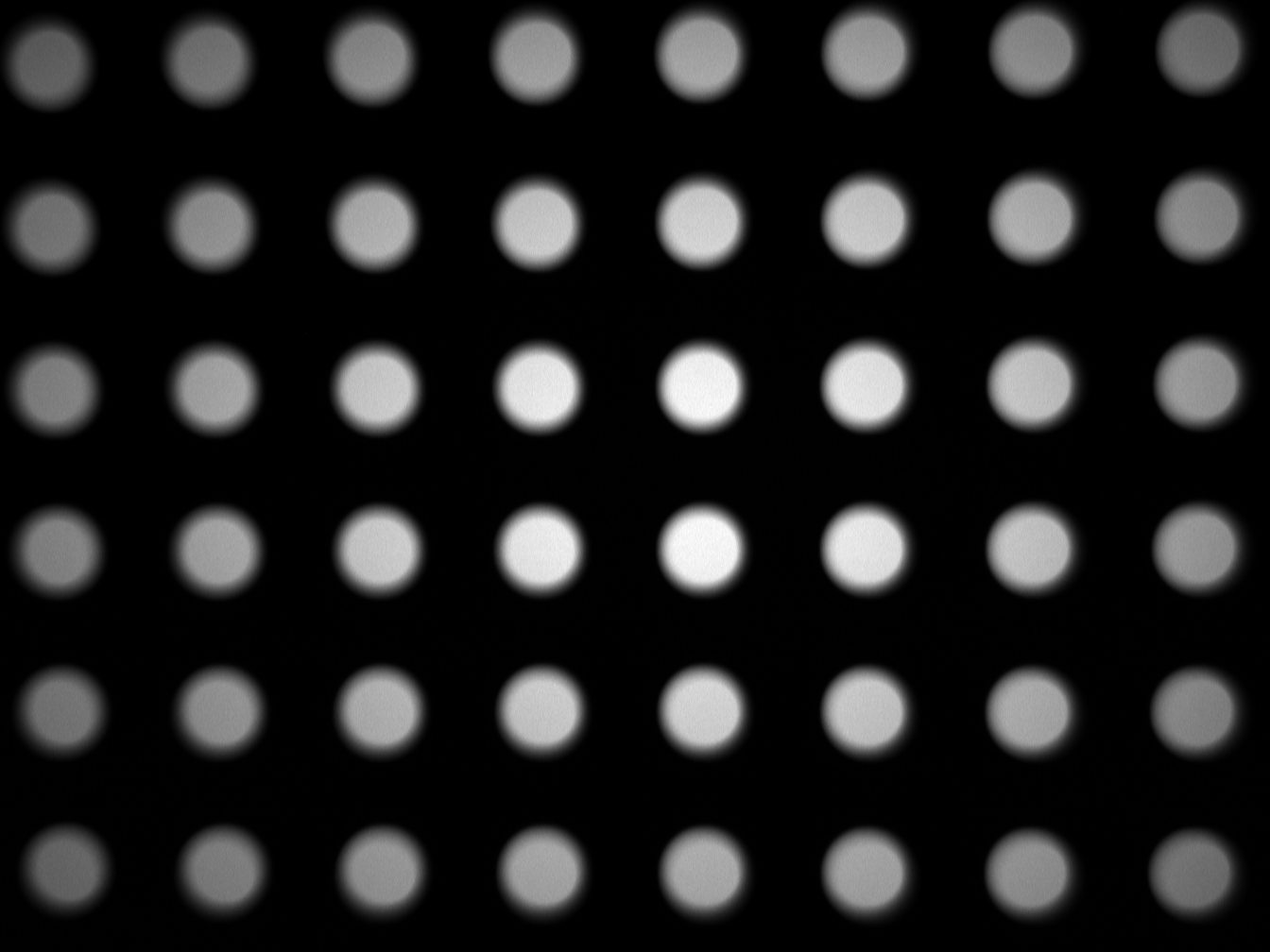}\label{subfig:calib_captured_supp}}
	\subfigure[Calibration pattern]{\includegraphics[width=0.23\textwidth]{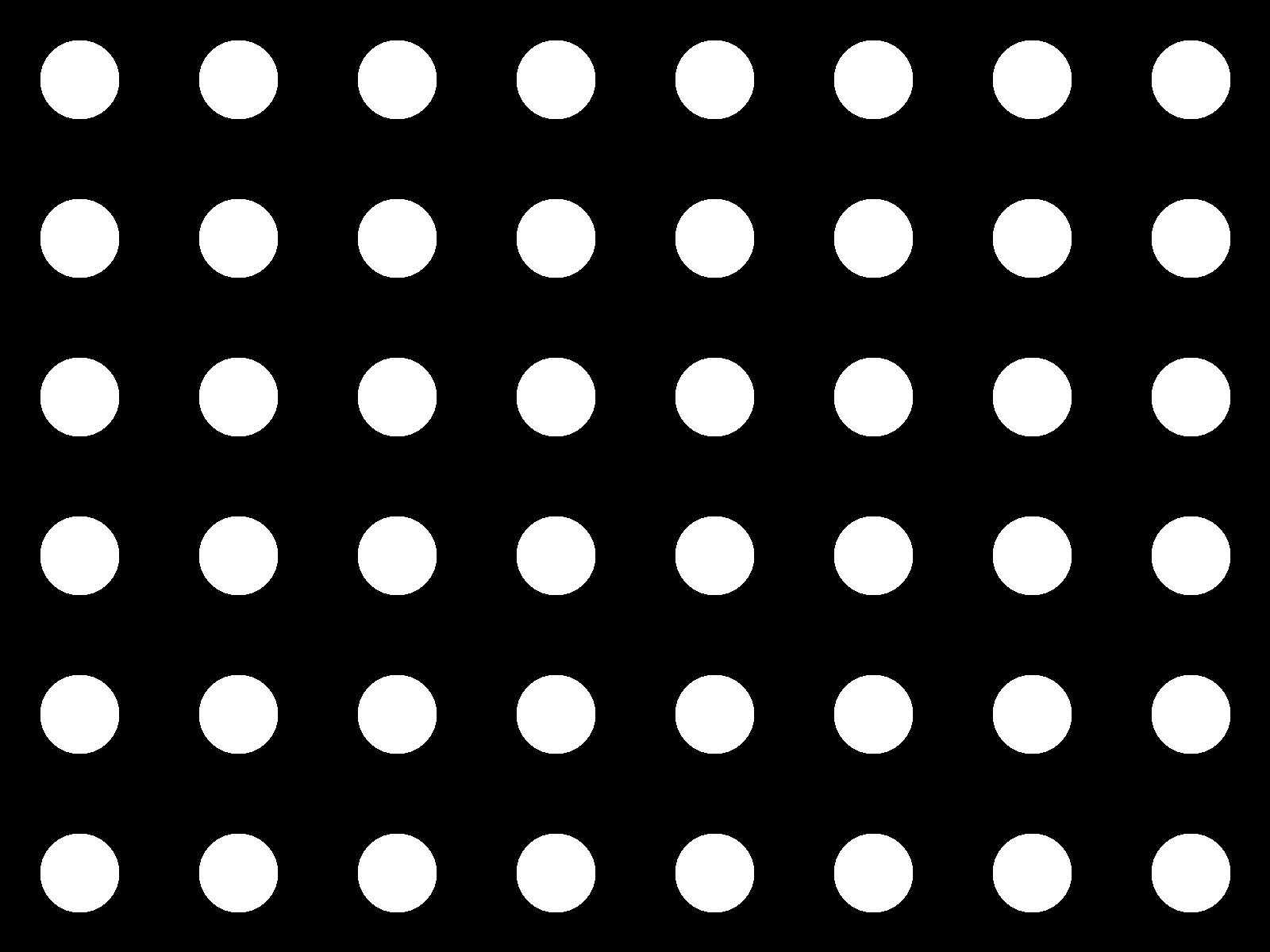}\label{subfig:calib_pattern_supp}}
	\subfigure[Left vignetting]{\includegraphics[width=0.23\textwidth]{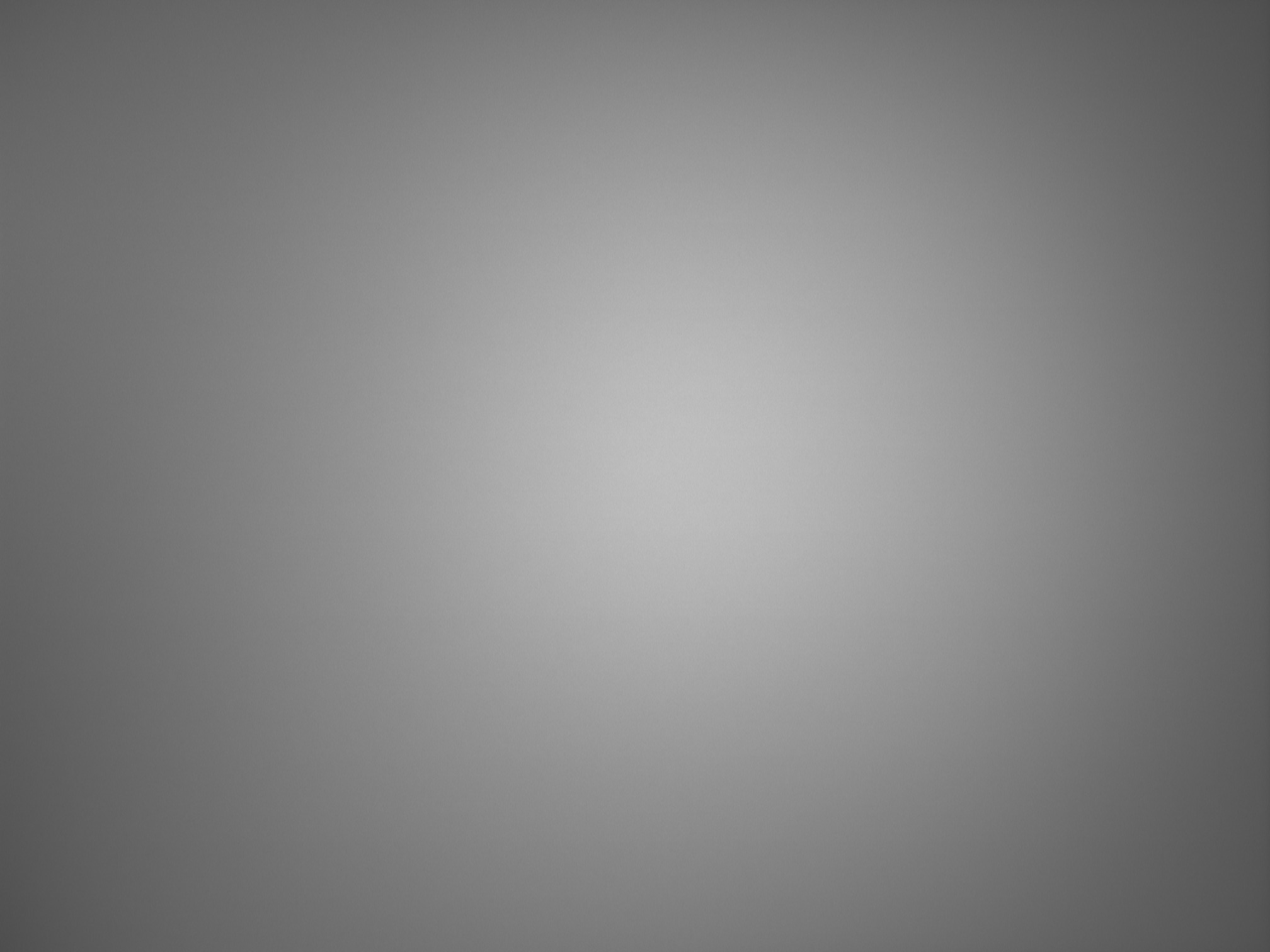}\label{subfig:white_left}}
	\subfigure[Right vignetting]{\includegraphics[width=0.23\textwidth]{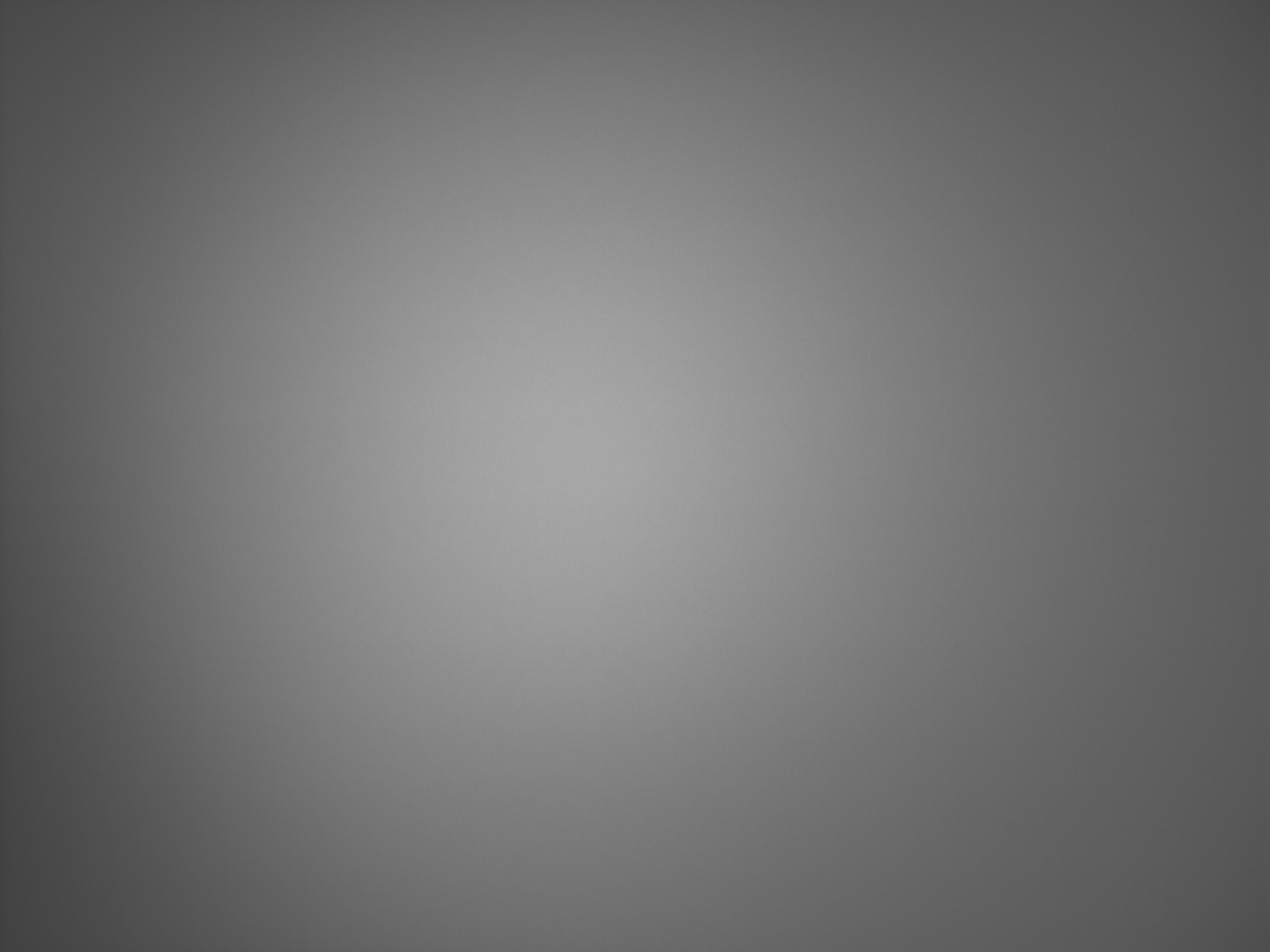}\label{subfig:white_right}}
	\caption{Captured image \subref{subfig:calib_captured_supp} of the calibration pattern \subref{subfig:calib_pattern_supp} that is used to calibrate the blur kernels. Left DP image \subref{subfig:white_left} and right DP image \subref{subfig:white_right} of a white sheet shot through a diffuser that is used to correct for vignetting.}
	\label{fig:dual_pixels_calib}
\end{figure}

We provide more information about our calibration procedure for the left and right blur kernels used as input to our method. We use a method similar to the one proposed by Mannan and Langer~\cite{mannan2016}, and calibrate blur kernels for left and right dual-pixel (DP) images independently (Fig.~\ref{fig:dual_pixels_calib}) for a specific focus distance.
Specifically, we image a regular grid of circular discs on a monitor screen at a distance of $\sim\unit[45]{cm}$ from the camera. We apply global thresholding and binarize the captured image, perform connected component analysis to identify the individual discs and their centers, and generate and align the binary sharp image $M$ with the known calibration pattern by solving for a homography between the calibration target disc centers and the detected centers. In order to apply radiometric correction, we also capture all-white and all-black images displayed on the same screen, and generate the grayscale latent sharp image as $I_l = M \odot I_w + (1 - M) \odot I_b$, where $\odot$ represents pixel-wise multiplication, and $I_w$ and $I_b$ are captured all-white and all-black images. Once we have the aligned latent image and the captured image, we can solve for spatially-varying blur kernels using the optimization proposed by Mannan and Langer~\cite{mannan2016}. Specifically, we solve for a $8\times6$ grid of kernels corresponding to $1344\times1008$ central field of view.

In addition to the blur kernels, we calibrate for different vignetting in left and right DP images. Specifically, for the same focus distance as above, we capture six images of a white sheet through a diffuser. We then average all left and right images individually to obtain the left and right vignetting patterns $W_l$ and $W_r$, respectively. 

\begin{figure}[h]
	\centering
	\subfigure[DP blur kernels with respect to scene depth]{\includegraphics[width=0.49\textwidth]{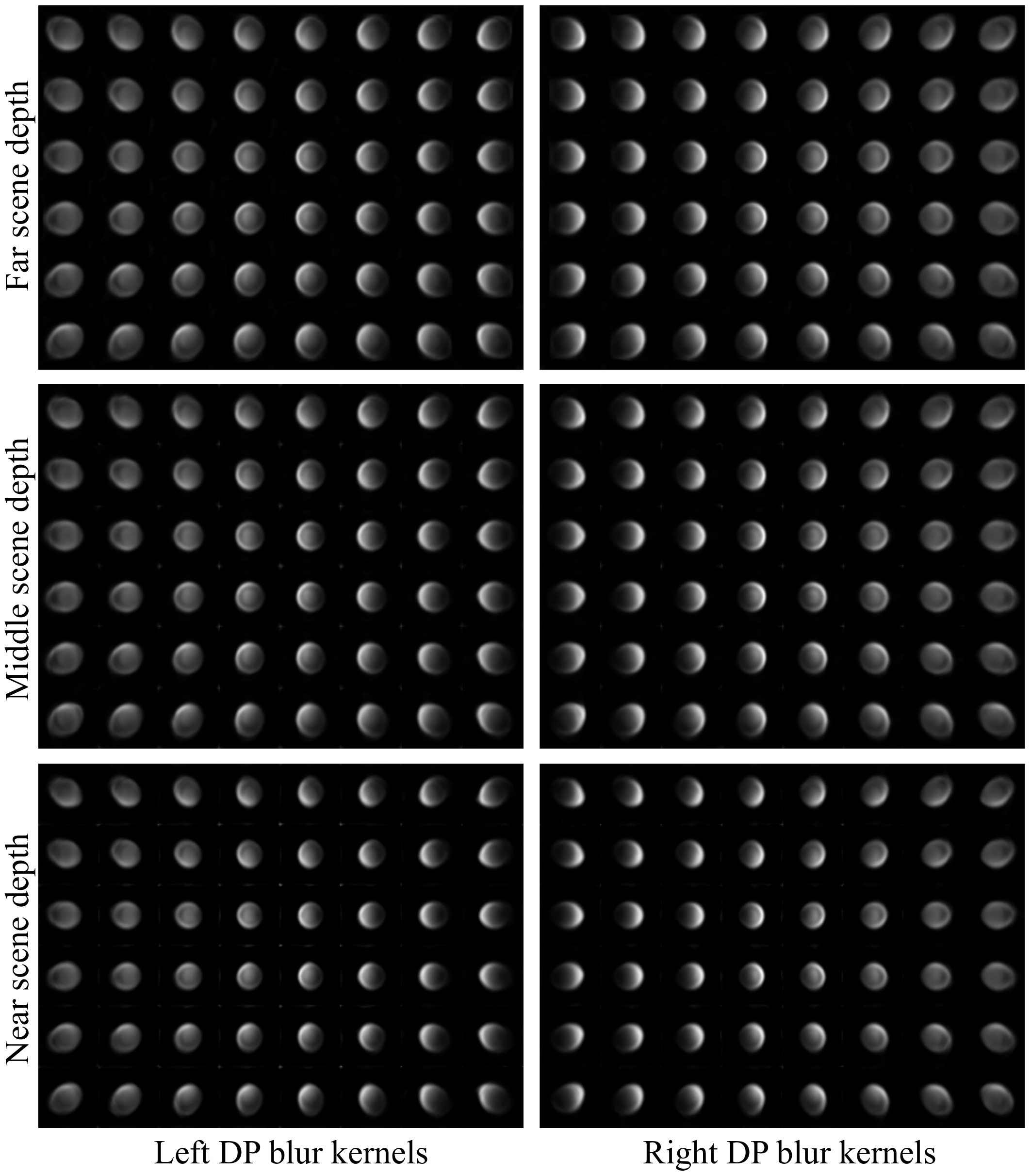}\label{subfig:kernels_depth}}
	\subfigure[DP blur kernels with respect to focus distance]{\includegraphics[width=0.49\textwidth]{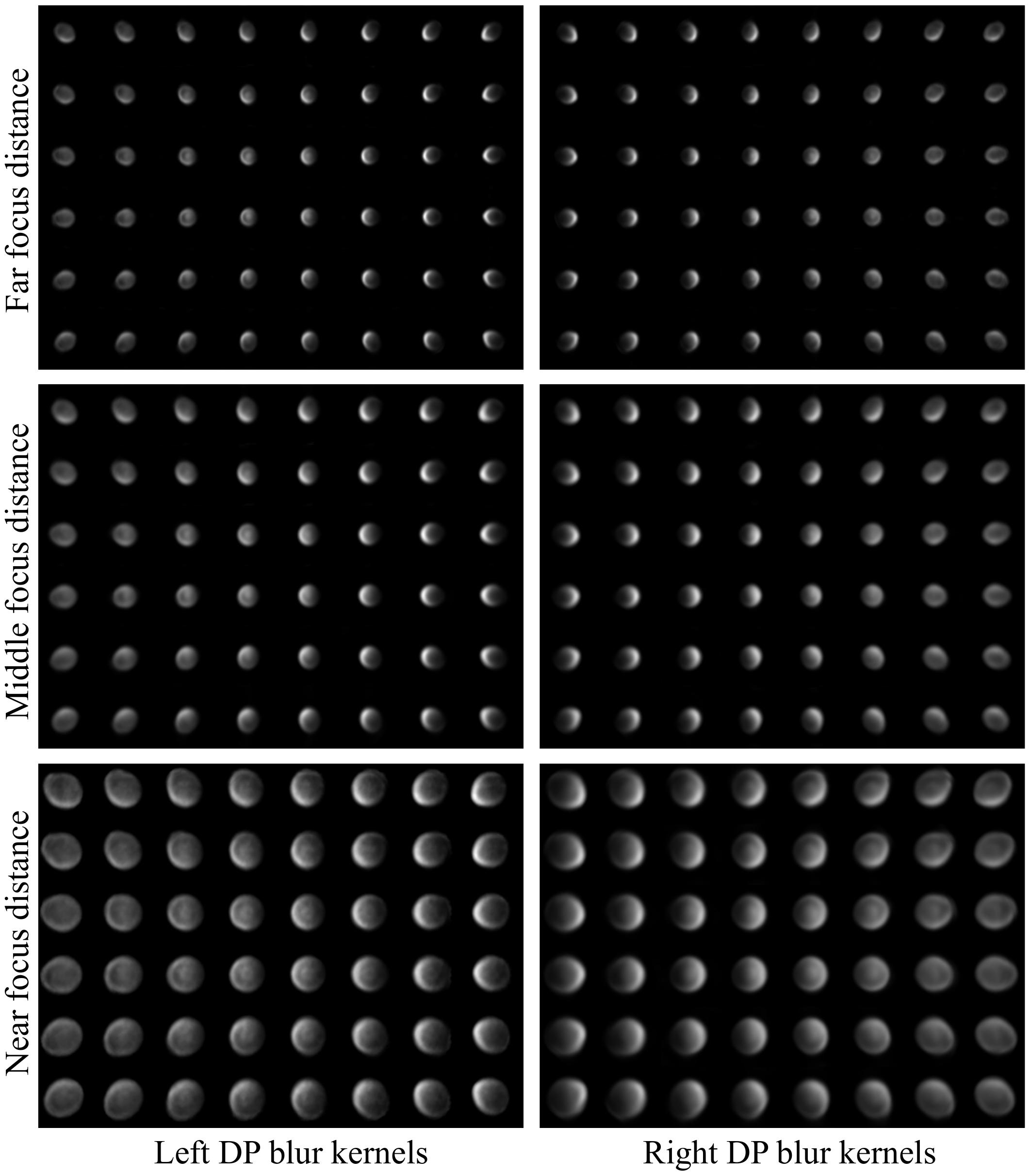}\label{subfig:kernels_focus_distance}}
	\caption{DP blur kernels with respect to scene depth and focus distance. We choose focus settings such that all scene contents are at or behind the focus plane, and calibrate for blur kernels either with the same focus settings but at different depths \subref{subfig:kernels_depth}, or at the same depth but with various focus distances \subref{subfig:kernels_focus_distance}.}
	\label{fig:kernels}
\end{figure}

Next, we explore how DP blur kernels change with respect to scene depth and focus distance (Fig.~\ref{fig:kernels}). As observed by Tang and Kutulakos~\cite{tang2012utilizing}, we find that kernels behave differently on the opposite sides of the focus plane. Therefore we choose focus settings such that all scene contents are at or behind the focus plane for all of our experiments, including this kernel analysis.
We observe that DP blur kernels are approximately resized versions of each other as the scene depth or focus distance changes, similar to the expected behavior for blur kernels in a regular image sensor.

\section{Additional Method Details \label{sec:method_details}}

In this section, we provide more technical details about our method. We explain how we render defocus maps from the MPI representation in Sec.~\ref{subsec:defocus_map}, provide the derivation of the bias correction term in Sec.~\ref{subsec:bias_correction_deri}, and finally define the total variation function $\tv(\cdot)$ and the edge map $\edge$ used in the regularization terms in Sec.~\ref{subsec:tv}.

\vspace{-0.1in}
\subsection{Defocus Map from MPI \label{subsec:defocus_map}}

We have shown in the main paper that an all-in-focus image can be composited from an MPI representation as:
\begin{equation}
	\hat{\I}_s = \sum_{i=1}^{N} \tc_i \cc_i
	= \sum_{i=1}^{N} \bracket{ \cc_i \ac_i \prod_{j=i+1}^{N} \paren{1-\ac_j}}\,.
	\label{eq:sharp_im_mpi_supp}
\end{equation}
We can synthesize a continuous-valued defocus map $\hat{\D}$ in a similar way as discussed by Tucker and Snavely~\cite{single_view_mpi}, by replacing all pixel intensities in Eq.~\eqref{eq:sharp_im_mpi_supp} with the defocus blur size $d_i$ of that layer:
\begin{equation}
	\hat{\D} = \sum_{i=1}^{N} \bracket{ d_i \ac_i \prod_{j=i+1}^{N} \paren{1-\ac_i}}\,.
\end{equation}

\vspace{-0.15in}
\subsection{Proof of Eq.~(4) of the Main Paper \label{subsec:bias_correction_deri}}

In this section, we provide a detailed derivation of the bias correction term. To be self-contained, we restate our assumed image formation model. Given an MPI representation, its corresponding DP images can be expressed as:
\begin{equation}
	\I_o^{\curly{l, r}} = \I_b^{\curly{l, r}} + \noise^{\curly{l, r}}\,,
	\label{eq:formulation_supp}
\end{equation}
where $\I_b^{\curly{l, r}}$ are the latent noise-free left and right defocused images, and $\noise^{\curly{l, r}}$ is additive white Gaussian noise with entries independent identically distributed with distribution $\Normal\paren{0, \sigma^2}$. Our goal is to optimize for an MPI with intensity-alpha layers $\paren{\hat{\cc}_i, \hat{\ac}_i}$, with defocus sizes $d_i, i \in \bracket{1, \dots, N}$, such that the $L_2$ loss $\norm{\hat{\I}_b^{\curly{l, r}} - \I_o^{\curly{l, r}}}_2^2$ is minimized.
We show that, in the presence of image noise, minimizing the above loss biases the estimated defocus map towards smaller blur values. Specifically, we quantify this bias and then correct for it in our optimization.

For simplicity, we assume for now that all scene contents lie on a single fronto-parallel plane with ground truth defocus size $d^{\star}$, and our scene representation is an MPI with a single opaque layer (i.e., $\hat{\ac}_i = \mathbf{1}$) with a defocus size hypothesis $d_i$. Under this assumption, the defocused image rendering equation (Eq.~(2) of the main paper)
\begin{equation}
	\hat{\I}_b^{\curlylr} = \sum_{i=1}^{N} \bracket{ {\paren{\kernel{d_i}{\curlylr} * \paren{\cc_i \ac_i}} \odot \prod_{j=i+1}^{N} \paren{\one-\kernel{d_j}{\curlylr} * \ac_j}}}\,
\end{equation}
reduces to
\begin{align}
	\hat{\I}_b^{\curly{l, r}} &= \kernel{d_i}{\curly{l, r}} * \hat{\cc}_i\,.
	\label{eq:defocused_im_mpi_2}
\end{align}
Similarly, Eq.~\eqref{eq:formulation_supp} becomes:
\begin{align}
	\I_o^{\curly{l, r}} = \kernel{d^{\star}}{\curly{l, r}} * {\cc}_i + \noise^{\curly{l, r}}.
	\label{eq:formulation_wiener}
\end{align}
We can express the above equations in the frequency domain as follows:
\begin{equation}
	\If_o^{\curly{l, r}} = \kernelf{d^{\star}}{\curly{l, r}} {\ccf}_i + \noisef^{\curly{l, r}} \,,
	\label{eq:formulation_wiener_fourier}
\end{equation}
where $\If_o^{\curly{l, r}}, \kernelf{d^{\star}}{\curly{l, r}}, {\ccf}_i$, and $\noisef^{\curly{l, r}}$ are the Fourier transforms of $\I_o^{\curly{l, r}}, \kernel{d^{\star}}{\curly{l, r}}, {\cc}_i$, and $\noise^{\curly{l, r}}$, respectively. Note that the entries of $\noisef^{\curly{l, r}}$ are also independent identically distributed with the same Gaussian distribution $\Normal \paren{0, \sigma^2}$ as the entries of $\noise^{\curly{l, r}}$.

We can obtain a maximum a posteriori (MAP) estimate of $\hat{\ccf}_i$ and $d_i$ by solving the following optimization problem~\cite{Zhou2009ICCV}:
\begin{align}
	\nonumber &\arg\max P\paren{\If_o^l, \If_o^r | \hat{\ccf}_i, d_i, \sigma} P\paren{\hat{\ccf}_i, d_i} \\
	=&\arg\max P\paren{\If_o^l, \If_o^r | \hat{\ccf}_i, d_i, \sigma} P\paren{\hat{\ccf}_i} \,.
	\label{eq:map}
\end{align}
According to Eq.~\eqref{eq:formulation_wiener_fourier}, we have
\begin{equation}
	P\paren{\If_o^l, \If_o^r | \hat{\ccf}_i, d_i, \sigma}
	\propto \exp\!\paren{\!-\frac{1}{2\sigma^2} \!\sum_{v = \curly{l,r}}\! \norm{\kernelf{d_i}{v} \hat{\ccf}_i- \If_o^v }^2 }\,.
\end{equation}
We also follow Zhou \etal~\cite{Zhou2009ICCV} in assuming a prior for the latent all-in-focus image such that:
\begin{equation}
	P\paren{\hat{\ccf}_i} \propto \exp \paren{-\frac{1}{2} \norm{\boldsymbol{\Phi} \hat{\ccf}_i }^2 } \,,
\end{equation}
where we define $\boldsymbol{\Phi}$ such that
\begin{equation}
	\abs{\boldsymbol{\Phi} \paren{f}} ^ 2 =  \frac{1} {\abs{\hat{\ccf_i} \paren{f}}^2}  \,,
	\label{eq:power}
\end{equation}
and $f$ is the frequency. As $\hat{\ccf}_i$ is the unknown variable, we approximate Eq.~\eqref{eq:power} by averaging the power spectrum over a set of natural images $\curly{\ccf_i}$:
\begin{equation}
	{\abs{\boldsymbol{\Phi} \paren{f}} ^ 2}  = \frac{1}{\int_{\ccf_i} \abs{\ccf_i \paren{f}}^2 \mu\paren{\ccf_i}} \,,
	\label{eq:power_expectation}
\end{equation}
where $\mu \paren{\ccf_i}$ represents the probability distribution of $\ccf_i$ in image domain.

Maximizing the log-likelihood of Eq.~\eqref{eq:map} is equivalent to minimizing the following loss:
\begin{equation}
	E\paren{d_i | \If_o^l, \If_o^r, \sigma}= 
	\min_{\hat{\ccf}_i} \paren{\sum_{v = \curly{l,r}} \norm{\kernelf{d_i}{v} \hat{\ccf}_i- \If_o^v }^2} + \norm{\sigma \boldsymbol{\Phi} \hat{\ccf}_i }^2 \,.
\end{equation}
$d_i$ can be estimated as the minimizer of the above energy function. Then given $d_i$, setting $\partial E / \partial \hat{\ccf}_i = 0$ yields the following solution of $\hat{\ccf}_i$, known as a \emph{generalized Wiener deconvolution with two observations}:
\begin{equation}
	\hat{\ccf}_i = \frac{\If_o^l \overline{\kernelf{d_i}{l}} + \If_o^r \overline{\kernelf{d_i}{r}} } {\abs{\kernelf{d_i}{l}}^2 + \abs{\kernelf{d_i}{r}}^2 + \sigma^2 \abs{ \boldsymbol{\Phi}}^2}\,,
	\label{eq:wiener_solution}
\end{equation}
where $\overline{\kernelf{d_i}{\curly{l,r}}}$ is the complex conjugate of ${\kernelf{d_i}{\curly{l,r}}}$, and $\abs{\kernelf{d_i}{\curly{l,r}}}^2 = \kernelf{d_i}{\curly{l,r}} \overline{\kernelf{d_i}{\curly{l,r}}}$.

We then evaluate the defocus size hypothesis $d_i$ by computing the minimization loss given the latent ground truth depth $d^{\star}$, and the noise level $\sigma$, that is,
\begin{align}
	E\paren{d_i| \kernelf{d^\star}{l}, \kernelf{d^\star}{r}, \sigma}&=\Expect_{\ccf_i, \If_o^l, \If_o^r} E\paren{d_i| \kernelf{d^\star}{l}, \kernelf{d^\star}{r}, \sigma, \ccf_i, \If_o^l, \If_o^r} \\
	&=\Expect_{\ccf_i, \If_o^l, \If_o^r}  \bracket{ \paren{\sum_{v = \curly{l,r}} \norm{\kernelf{d_i}{v} \hat{\ccf}_i- \If_o^v }^2} + \norm{\sigma \boldsymbol{\Phi} \hat{\ccf}_i }^2} \,,
\end{align}
where $\Expect \paren{\cdot}$ is the expectation. Substituting $\hat{\ccf}_i$ with Eq.~\eqref{eq:wiener_solution} gives us:
\vspace{-0.05in}
\begin{align}
	\nonumber &E\paren{d_i| \kernelf{d^\star}{l}, \kernelf{d^\star}{r}, \sigma} \\ 
	=&\Expect_{\ccf_i, \If_o^l, \If_o^r}  \bracket{\paren{ \sum_{v = \curly{l,r}} \norm{\kernelf{d_i}{v} \frac{\If_o^l \overline{\kernelf{d_i}{l}} + \If_o^r \overline{\kernelf{d_i}{r}} } {\abs{\kernelf{d_i}{l}}^2 + \abs{\kernelf{d_i}{r}}^2 + \sigma^2 \abs{ \boldsymbol{\Phi}}^2} - \If_o^v }^2 }+ \norm{\sigma \boldsymbol{\Phi} \frac{\If_o^l \overline{\kernelf{d_i}{l}} + \If_o^r \overline{\kernelf{d_i}{r}} } {\abs{\kernelf{d_i}{l}}^2 + \abs{\kernelf{d_i}{r}}^2 + \sigma^2 \abs{ \boldsymbol{\Phi}}^2} }^2}  \,.
\end{align}
Then substituting $\If_o^v$ with Eq.~\eqref{eq:formulation_wiener_fourier}, we get:
\vspace{-0.05in}
\begin{align}
	\nonumber &E\paren{d_i| \kernelf{d^\star}{l}, \kernelf{d^\star}{r}, \sigma} \\ 
	\nonumber =& \Expect_{\ccf_i, \noisef^l, \noisef^r} \Bigg[ \paren{ \sum_{v = \curly{l,r}} \norm{\kernelf{d_i}{v} \frac{ \paren{\kernelf{d^\star}{l} \ccf_i + \noisef^{l}} \overline{\kernelf{d_i}{l}} + \paren{\kernelf{d^\star}{r} \ccf_i + \noisef^{r}} \overline{\kernelf{d_i}{r}} } {\abs{\kernelf{d_i}{l}}^2 +  \abs{\kernelf{d_i}{r}}^2  + \sigma^2 \abs{ \boldsymbol{\Phi}}^2} - \paren{\kernelf{d^\star}{v} \ccf_i + \noisef^{v}} }^2 } + \\
	& \hphantom{\Expect_{\ccf_i, \noisef^l, \noisef^r} \Bigg[} \norm{\sigma \boldsymbol{\Phi} \frac{ \paren{\kernelf{d^\star}{l} \ccf_i + \noisef^{l}} \overline{\kernelf{d_i}{l}} + \paren{\kernelf{d^\star}{r} \ccf_i + \noisef^{r}} \overline{\kernelf{d_i}{r}} } {\abs{\kernelf{d_i}{l}}^2 +  \abs{\kernelf{d_i}{r}}^2  + \sigma^2 \abs{ \boldsymbol{\Phi}}^2} }^2 \Bigg] \,.
\end{align}
We now define $B = {\abs{\kernelf{d_i}{l}}^2  + \abs{\kernelf{d_i}{r}}^2  + \sigma^2 \abs{ \boldsymbol{\Phi}}^2}$. We can rearrange the above equation as:
\vspace{-0.05in}
\begin{align}
	\nonumber & E\paren{d_i| \kernelf{d^\star}{l}, \kernelf{d^\star}{r}, \sigma} \\
	\nonumber =& \Expect_{\ccf_i, \noisef^l, \noisef^r}  \Bigg[ \paren{ \sum_{v = \curly{l,r}} \norm{ \frac{\ccf_i \bracket{\kernelf{d_i}{v}  \paren{\kernelf{d^\star}{l} \overline{\kernelf{d_i}{l}} + \kernelf{d^\star}{r} \overline{\kernelf{d_i}{r}}} - \kernelf{d^\star}{v} B} } {B} + \frac{\kernelf{d_i}{v} \paren{\noisef^{l} \overline{\kernelf{d_i}{l}} + \noisef^{r} \overline{\kernelf{d_i}{r}}} }{B} - \noisef^{v} }^2 } + \\ 
	& \hphantom{\Expect_{\ccf_i, \noisef^l, \noisef^r} \Bigg[ } \norm{\sigma \boldsymbol{\Phi} \frac{ \ccf_i \paren{\kernelf{d^\star}{l} \overline{\kernelf{d_i}{l}} + \kernelf{d^\star}{r} \overline{\kernelf{d_i}{r}}} } {B} + \sigma \boldsymbol{\Phi} \frac{\noisef^{l} \overline{\kernelf{d_i}{l}} + \noisef^{r} \overline{\kernelf{d_i}{r}} } {B} }^2 \Bigg] \,.
\end{align}
Given that the entries of $\noisef^{\curly{l,r}}$ are independent identically distributed with distribution $\Normal\paren{0, \sigma^2}$, we have $\Expect \paren{\noisef^{v}} = \mathbf{0}, \Expect \paren{{\noisef^{v}}^2} = \boldsymbol{\sigma}^2$ and $\Expect \paren{\noisef^{l} \noisef^{r}} = \mathbf{0}$, and we can simplify the above equation as:
\begin{align}
	\nonumber &E\paren{d_i| \kernelf{d^\star}{l}, \kernelf{d^\star}{r}, \sigma} \\ 
	\nonumber =& \Expect_{\ccf_i, \noisef^l, \noisef^r} \Bigg[ \paren{ \sum_{v = \curly{l,r}} \norm{ \frac{\ccf_i \bracket{\kernelf{d_i}{v}  \paren{\kernelf{d^\star}{l} \overline{\kernelf{d_i}{l}} + \kernelf{d^\star}{r} \overline{\kernelf{d_i}{r}}} - \kernelf{d^\star}{v} B} } {B} }^2 + \norm{\frac{\kernelf{{d}_i}{v} \paren{\noisef^{l} \overline{\kernelf{d_i}{l}} + \noisef^{r} \overline{\kernelf{d_i}{r}}} }{B} - \noisef^{v} }^2 } + \\ 
	& \hphantom{\Expect_{\ccf_i, \noisef^l, \noisef^r} \Bigg[} \norm{\sigma \boldsymbol{\Phi} \frac{ \ccf_i \paren{\kernelf{d^\star}{l} \overline{\kernelf{d_i}{l}} + \kernelf{d^\star}{r} \overline{\kernelf{d_i}{r}}} } {B}}^2 + \norm{\sigma \boldsymbol{\Phi} \frac{\noisef^{l} \overline{\kernelf{d_i}{l}} + \noisef^{r} \overline{\kernelf{d_i}{r}} } {B} }^2 \Bigg] \\
	\nonumber =& \Expect_{\ccf_i} \Bigg\{ \bracket{ \sum_{v = \curly{l,r}}  \norm{ \frac{\ccf_i \bracket{\kernelf{d_i}{v}  \paren{\kernelf{d^\star}{l} \overline{\kernelf{d_i}{l}} +  \kernelf{d^\star}{r} \overline{\kernelf{d_i}{r}}} - \kernelf{d^\star}{v} B} } {B} }^2 + \sigma^2 \paren{ \norm{\frac{{\kernelf{d_i}{v}}^2 + \sigma^2 \abs{ \boldsymbol{\Phi}}^2 }{B} }^2 + \norm{\frac{{\kernelf{d_i}{l}}  {\kernelf{d_i}{r}}  }{B} }^2} } + \\ 
	& \hphantom{\Expect_{\ccf_i} \Bigg\{} \norm{\sigma \boldsymbol{\Phi} \frac{ \ccf_i \paren{\kernelf{d^\star}{l} \overline{\kernelf{d_i}{l}} + \kernelf{d^\star}{r} \overline{\kernelf{d_i}{r}}} } {B}}^2  + \sigma^2 \paren{ \norm{\sigma \boldsymbol{\Phi} \frac{ {\kernelf{d_i}{l}} } {B} }^2 +  \norm{\sigma \boldsymbol{\Phi} \frac{ {\kernelf{d_i}{r}} } {B} }^2} \Bigg\} \,.
\end{align}
Recall that, in Eq.~\eqref{eq:power_expectation}, we defined $\boldsymbol{\Phi} \paren{f}$ such that $\frac{1}{\abs{\boldsymbol{\Phi} \paren{f}} ^ 2}  = \int_{\ccf_i} \abs{\ccf_i \paren{f}}^2 \mu\paren{\ccf_i}.$
Then we can further simplify $E\paren{d_i| \kernelf{d^\star}{l}, \kernelf{d^\star}{r}, \sigma}$ as:
\begin{align}
	\nonumber &E\paren{d_i| \kernelf{d^\star}{l}, \kernelf{d^\star}{r}, \sigma} \\
	\nonumber =& \sum_{f} \bracket{ \frac{\frac{1}{\abs{\boldsymbol{\Phi}} ^ 2} \abs{\kernelf{d^\star}{l} \kernelf{d_i}{r} - \kernelf{d^\star}{r} \kernelf{d_i}{l}}^2}{B} } + \sum_{f} \bracket{ \frac{\frac{1}{\abs{\boldsymbol{\Phi}} ^ 2} \sigma^2 \abs{ \boldsymbol{\Phi}}^2 \paren{ \abs {\kernelf{d^\star}{l}}^2 + \abs {\kernelf{d^\star}{r}}^2 }}{B} } + \\
	& \sum_{f} \bracket{ \sigma^2 \paren{\norm{\frac{{\kernelf{d_i}{l}}^2 + \sigma^2 \abs{ \boldsymbol{\Phi}}^2 }{B} }^2 + \norm{\frac{{\kernelf{d_i}{r}}^2 + \sigma^2 \abs{ \boldsymbol{\Phi}}^2 }{B} }^2 +  2 \norm{\frac{{\kernelf{d_i}{l}}  {\kernelf{d_i}{r}}  }{B} }^2 +  \norm{\sigma \boldsymbol{\Phi} \frac{ {\kernelf{d_i}{l}} } {B} }^2 +  \norm{\sigma \boldsymbol{\Phi} \frac{ {\kernelf{d_i}{r}} } {B} }^2 } } \\
	=& \sum_{f} \bracket{ \frac{\frac{1}{\abs{\boldsymbol{\Phi}} ^ 2} \abs{\kernelf{d^\star}{l} \kernelf{d_i}{r} - \kernelf{d^\star}{r} \kernelf{d_i}{l}}^2}{B} } + \sigma^2\sum_{f} \bracket{ \frac{ \abs {\kernelf{d^\star}{l}}^2 + \abs {\kernelf{d^\star}{r}}^2 }{B} +  \frac{\sigma^2 \abs{ \boldsymbol{\Phi}}^2}{B} + 1 } \\
	=& \sum_{f} \bracket{ \frac{\frac{1}{\abs{\boldsymbol{\Phi}} ^ 2} \abs{\kernelf{d^\star}{l} \kernelf{d_i}{r} - \kernelf{d^\star}{r} \kernelf{d_i}{l}}^2}{ \abs{\kernelf{d_i}{l}}^2 + \abs{\kernelf{d_i}{r}}^2  + \sigma^2 \abs{ \boldsymbol{\Phi}}^2} } + \sigma^2\sum_{f} \bracket{ \frac{ \abs {\kernelf{d^\star}{l}}^2 + \abs {\kernelf{d^\star}{r}}^2 + \sigma^2 \abs{\boldsymbol{\Phi}}^2 }{\abs{\kernelf{d_i}{l}}^2 + \abs{\kernelf{d_i}{r}}^2  + \sigma^2 \abs{ \boldsymbol{\Phi}}^2}  + 1 } \,.
	\label{eq:depth_loss}
\end{align}

If we define $C_1\big(\kernelf{d_i}{\curly{l,r}},\sigma,\boldsymbol{\Phi}\big) = \frac{\frac{1}{\abs{\boldsymbol{\Phi}} ^ 2} }{ \abs{\kernelf{d_i}{l}}^2 + \abs{\kernelf{d_i}{r}}^2  + \sigma^2 \abs{ \boldsymbol{\Phi}}^2}$, and $C_2(\sigma) = \sigma^2\sum_{f} 1$, then Eq.~\eqref{eq:depth_loss} boils down to Eq.~(4) of the main paper.

\subsection{Edge-aware Total Variation Function \label{subsec:tv}}

We first define a pixel-wise total variation function of a single-layer image $\I$ that in used in both the intensity smoothness prior $\loss_{\mathrm{intensity}}$ and the alpha and transmittance smoothness prior $\loss_{\mathrm{alpha}}$:
\begin{equation}
	\tv \paren{\I} = \sqrt{\I^2 * g - \paren{\I * g}^2}\,,
\end{equation}
where $g$ is a two-dimensional Gaussian blur kernel:
\begin{equation}
	g = \begin{bmatrix}
		\sfrac{1}{16} & \sfrac{1}{8} & \sfrac{1}{16} \\ 
		\sfrac{1}{8} & \sfrac{1}{4} & \sfrac{1}{8} \\ 
		\sfrac{1}{16} & \sfrac{1}{8} & \sfrac{1}{16}
	\end{bmatrix}\,.
\end{equation}
Each ``pixel'' in $\tv\paren{\I}(x,y)$ is equivalent to, for the $3 \times 3$ window surrounding pixel $(x, y)$ in $\I$, computing the sample standard deviation (weighted by a Gaussian kernel) of the pixel intensities in that window.
This follows easily from two facts: 1) as $g$ sums to $1$ by construction, $\I * g$ produces an image whose pixel intensities can be viewed as expectations of their surrounding $3 \times 3$ input patch; and 2) the standard deviation $\sqrt{\Expect \bracket{(X-\Expect \bracket{X})^{2}}}$ can be written equivalently as $\sqrt{\Expect \bracket{X^{2}}-\Expect \bracket{X}^{2}}$.

As done in prior work~\cite{single_view_mpi}, we would like to encourage edge-aware smoothness in addition to minimizing total variation, so a bilateral edge mask is computed using this total variation:
\begin{equation}
	\edge \paren{\I} = \mathbf{1} - \exp \paren{- \frac{\I^2 * g - \paren{\I * g}^2}{2 \beta^2} } .
\end{equation}
In this equation, $\beta$ is set to $\sfrac{1}{32}$ (assuming pixel intensities are in $\bracket{0, 1}$).
A joint total variation function that takes into account both the original and the edge-aware total variation is then defined as:
\begin{equation}
	\tv_{\edge} \paren{\I, \boldsymbol{E}} = \huber\paren{\tv\paren{\I}} + \paren{1 - \boldsymbol{E}} \odot \huber\paren{\tv\paren{\I}} .
\end{equation}

\vspace{-0.15in}
\section{Additional Implementation Details and Experimental Results \label{sec:experiments}}
\begin{figure*}[t]
	\centering
	\subfigure[\scriptsize Input image]{\includegraphics[width=0.157 \textwidth]{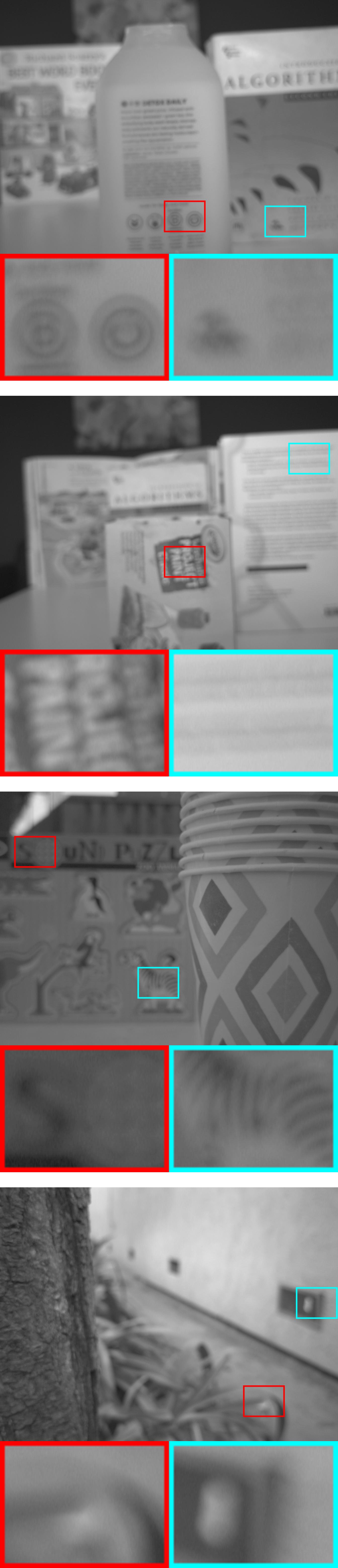} \label{subfig:inputs_dd_comparison}}
	\subfigure[\scriptsize GT all-in-focus image]{\includegraphics[width=0.157\textwidth]{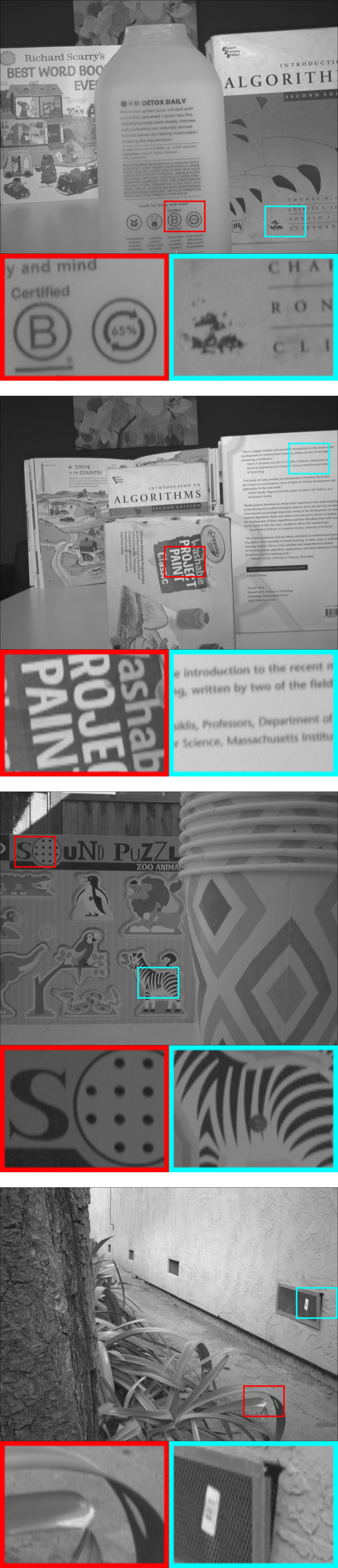} \label{subfig:sharp_im_gt_comparison}}
	\subfigure[\scriptsize Ours]{\includegraphics[width=0.157\textwidth]{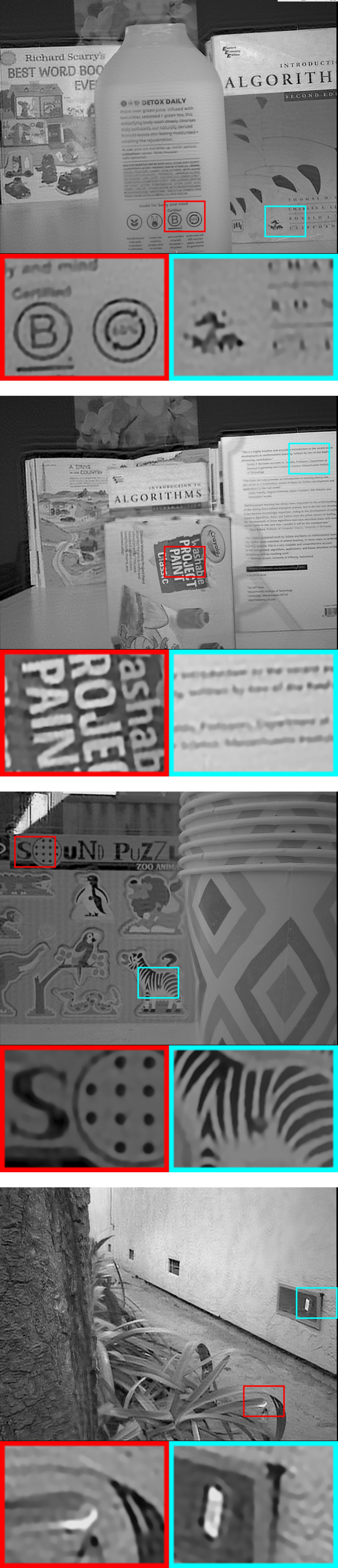} \label{subfig:sharp_im_ours_comparison}}
	\subfigure[\scriptsize Wiener deconv.~\cite{ Zhou2009ICCV}]{\includegraphics[width=0.157\textwidth]{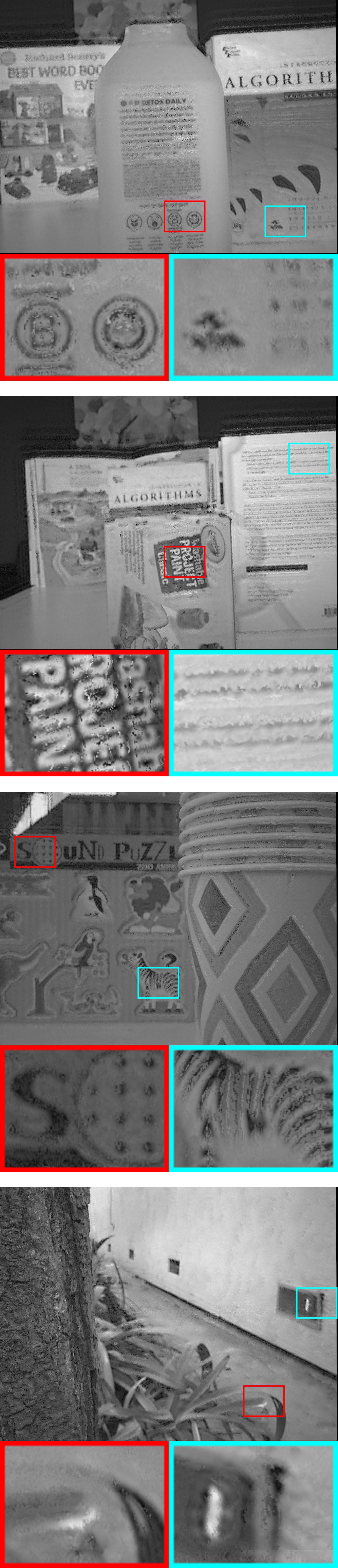} \label{subfig:sharp_im_wiener_comparison}}
	\subfigure[\scriptsize DPDNet \cite{abuolaim2020defocus}]{\includegraphics[width=0.157\textwidth]{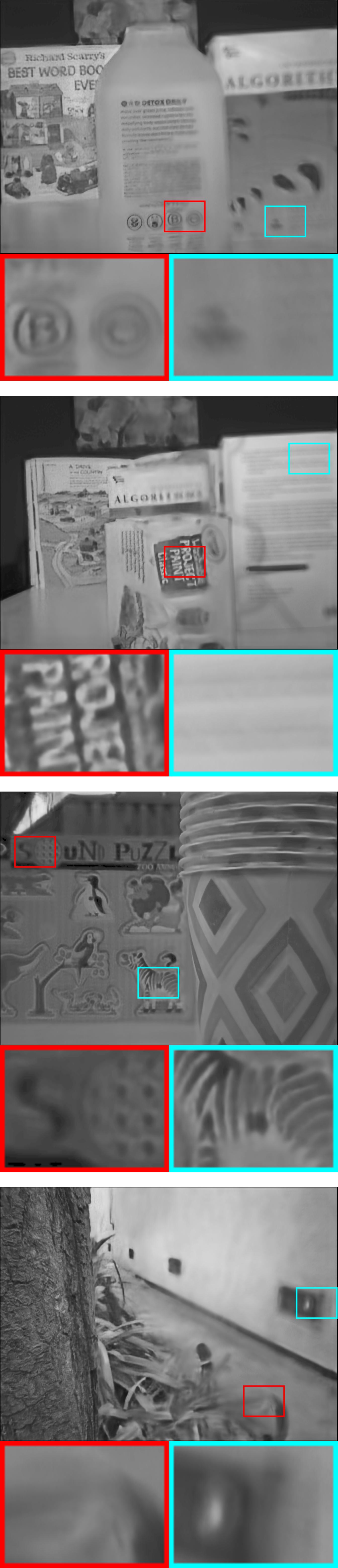} \label{subfig:sharp_im_ECCV_comparison}}
	\subfigure[\scriptsize DPDNet w/ calib
	\cite{abuolaim2020defocus}]{\includegraphics[width=0.157\textwidth]{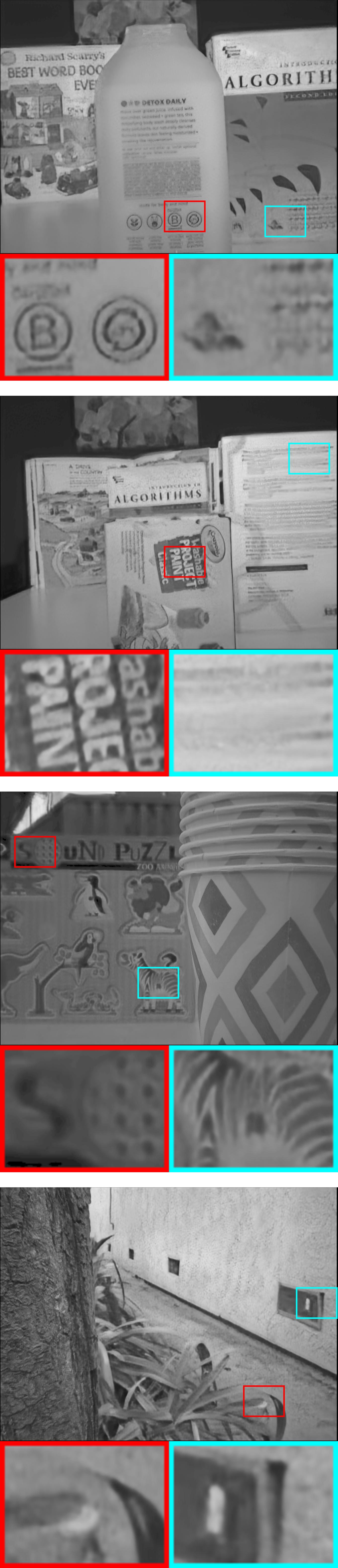} \label{subfig:sharp_im_ECCV_with_white_sheet_calibration_comparison}}
	\vspace{-0.2in}
	\caption{More qualitative comparisons of various defocus deblurring methods.}
	\label{fig:comparison_deblurring_supp}
	\vspace{-0.1in}
\end{figure*}

\begin{figure*}
	\centering
	\subfigure[\scriptsize Input image]{\includegraphics[width=0.105\textwidth]{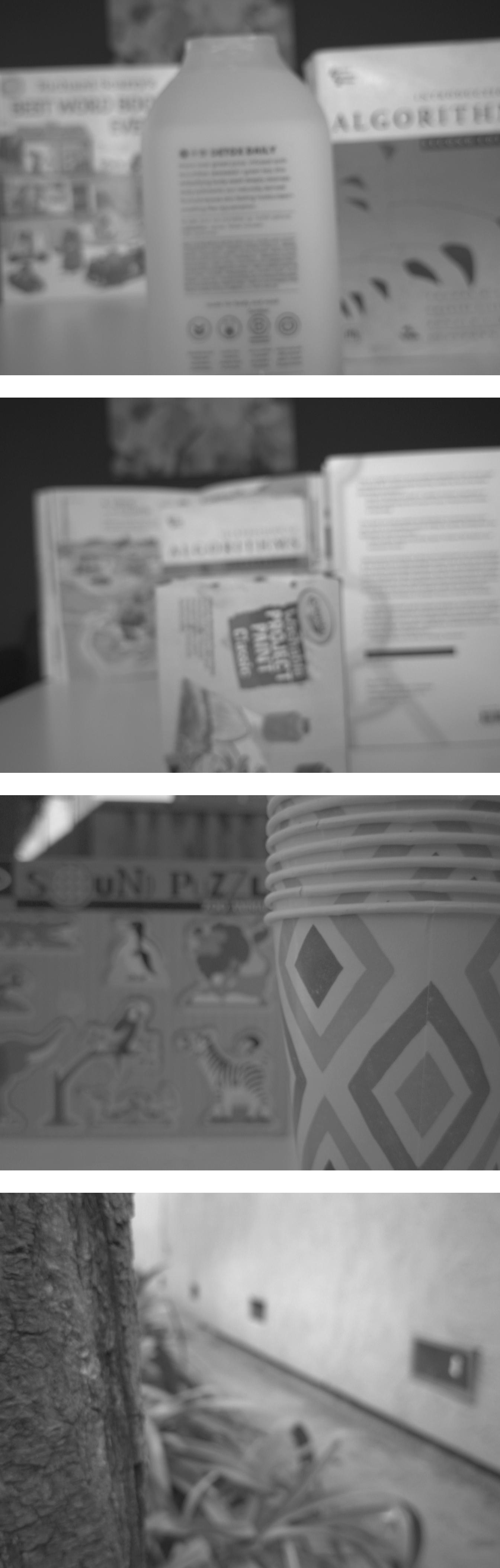}\label{subfig:inputs_defocus_comparison}}
	\subfigure[\scriptsize Ground Truth]{\includegraphics[width=0.105\textwidth]{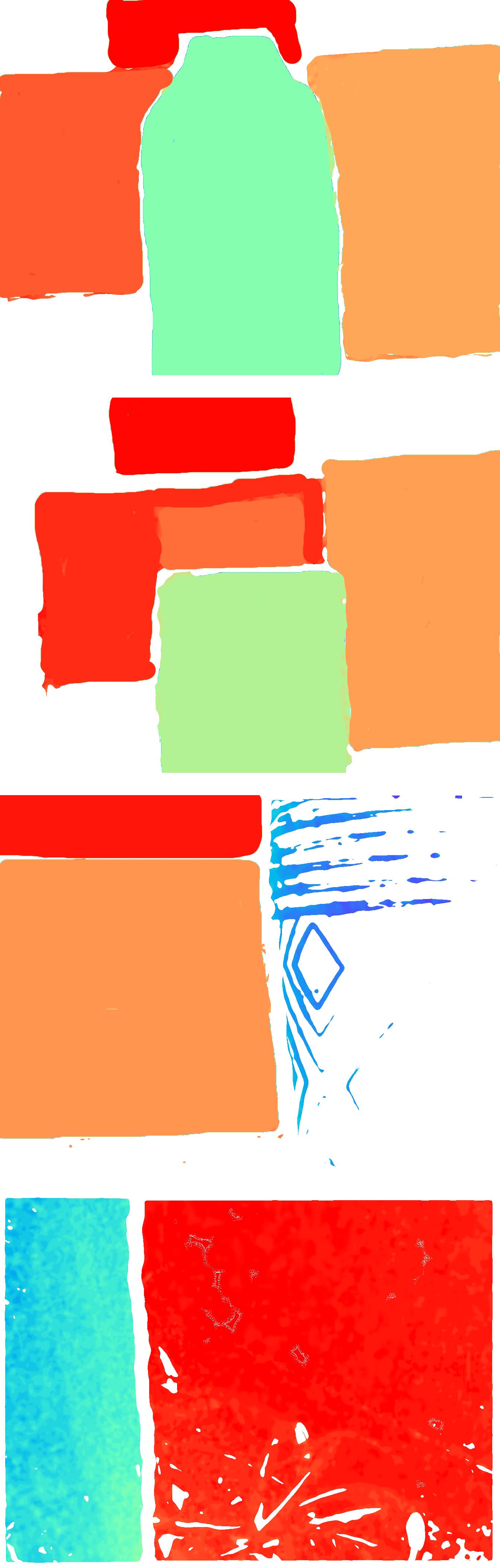}\label{subfig:defocus_gt_comparison}}
	\subfigure[\scriptsize Ours]{\includegraphics[width=0.105\textwidth]{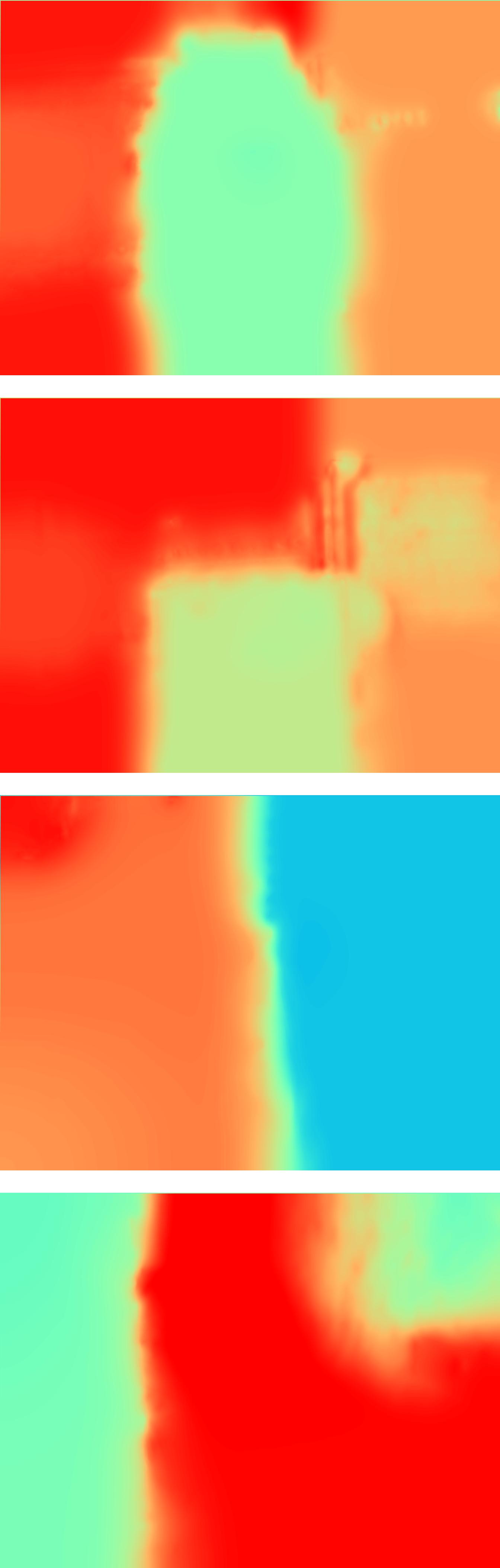}\label{subfig:defocus_ours_comparison}}
	\subfigure[\scriptsize Ours w/ GF]{\includegraphics[width=0.105\textwidth]{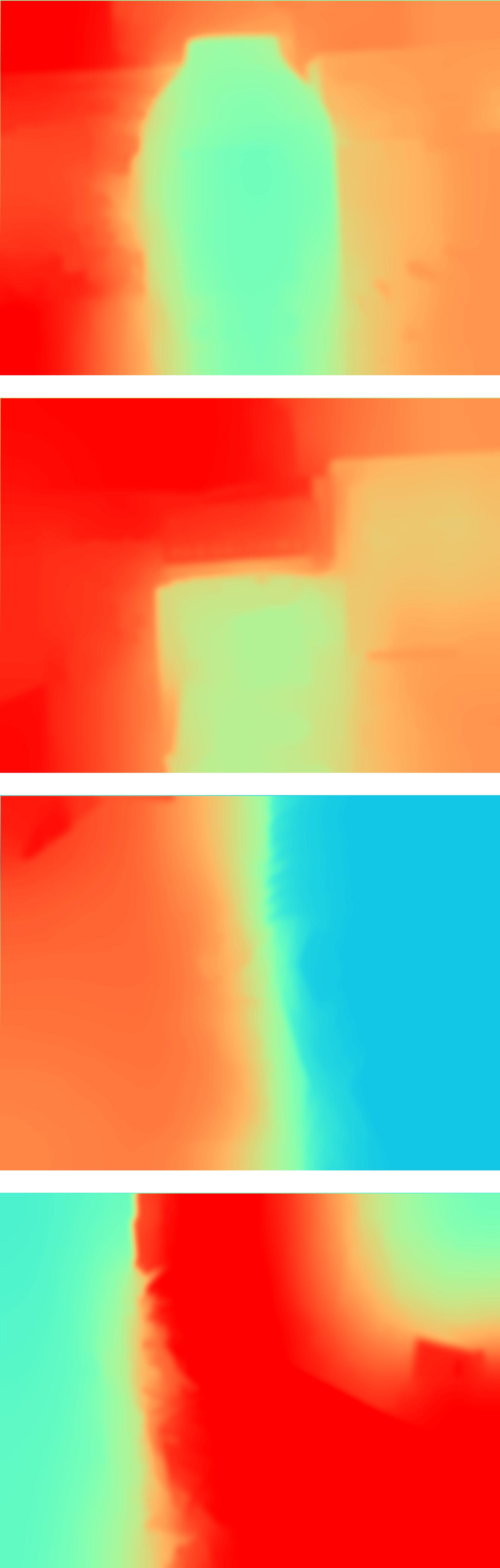}\label{subfig:defocus_ours_gf_comparison}}
	\subfigure[\scriptsize Wiener~\cite{ Zhou2009ICCV}]{\includegraphics[width=0.105\textwidth]{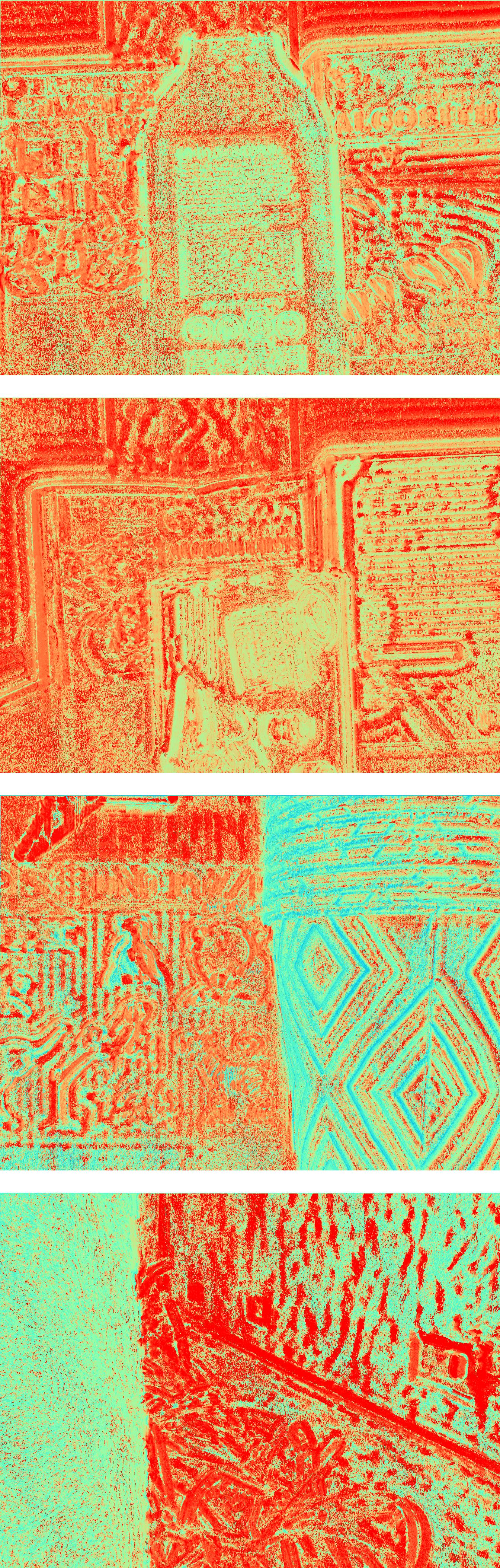}\label{subfig:defocus_wiener_comparison}}
	\subfigure[\scriptsize DMENet~\cite{Lee_2019_CVPR}]{\includegraphics[width=0.105\textwidth]{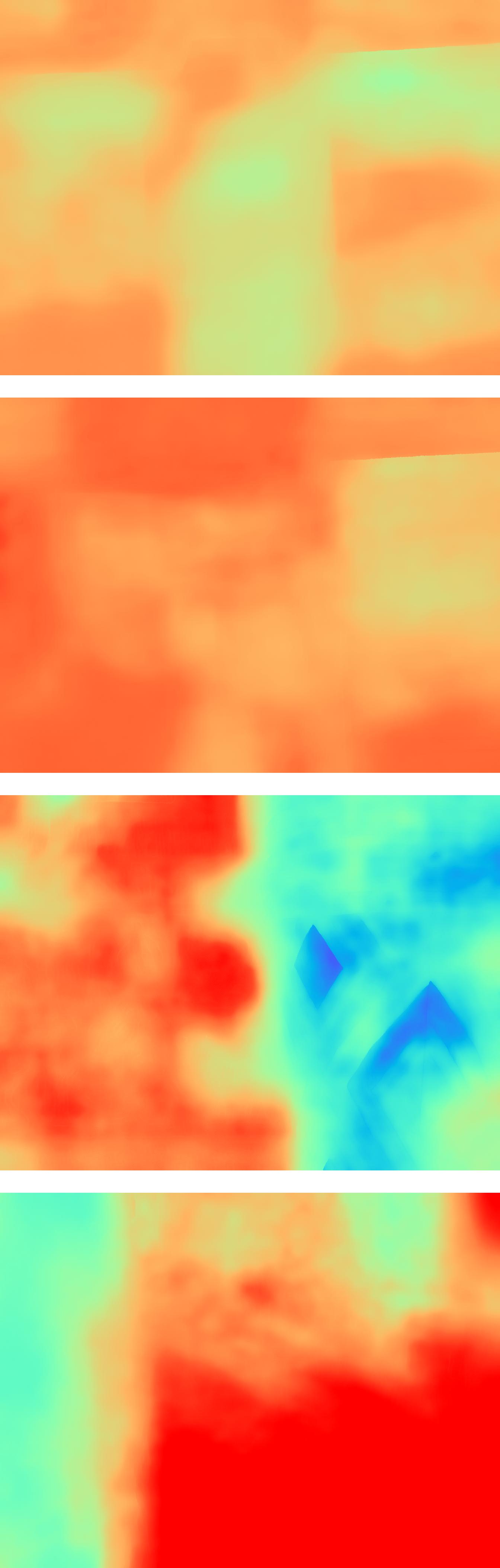}\label{subfig:depth_DMENet_comparison}}
	\subfigure[\scriptsize \cite{punnappurath2020modeling}]{\includegraphics[width=0.105\textwidth]{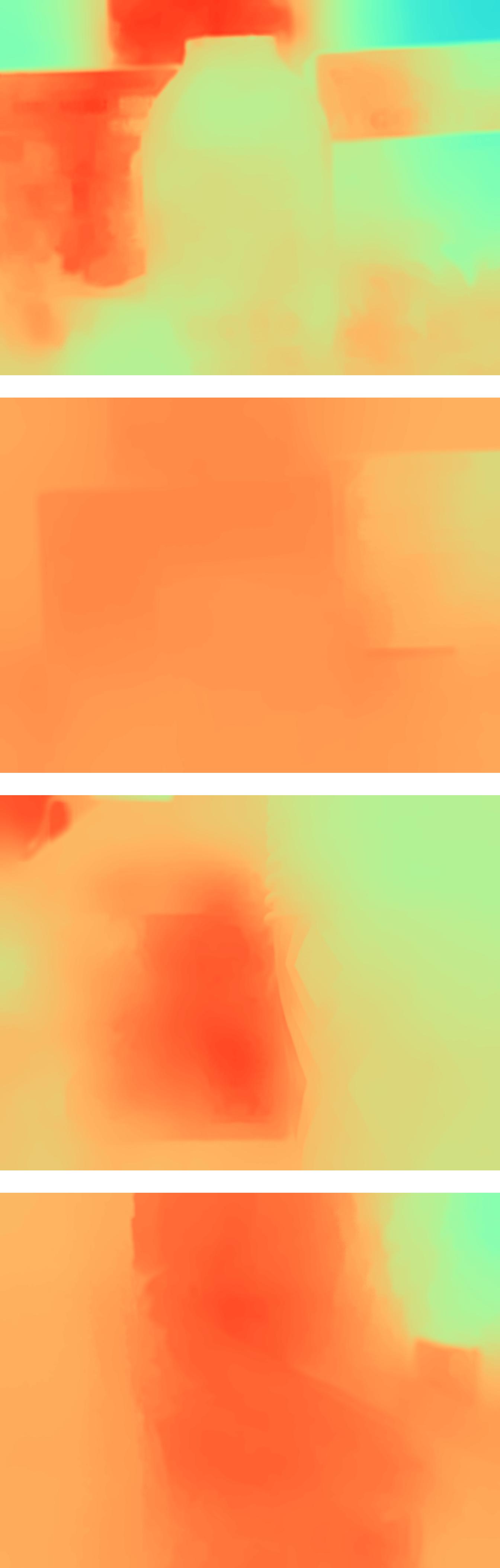}\label{subfig:defocus_ICCP_comparison}}
	\subfigure[\scriptsize Garg ~\cite{garg2019learning}]{\includegraphics[width=0.105\textwidth]{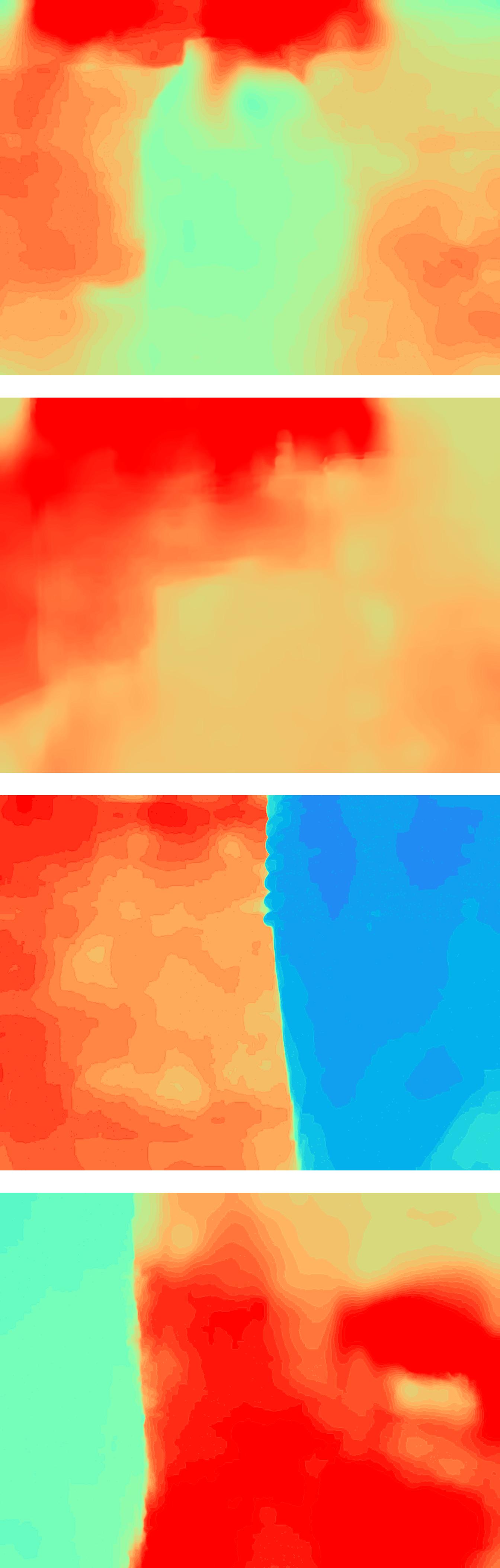}\label{subfig:defocus_ICCV_comparison}}
	\subfigure[\scriptsize Wadhwa ~\cite{wadhwa2018synthetic}]{\includegraphics[width=0.105\textwidth]{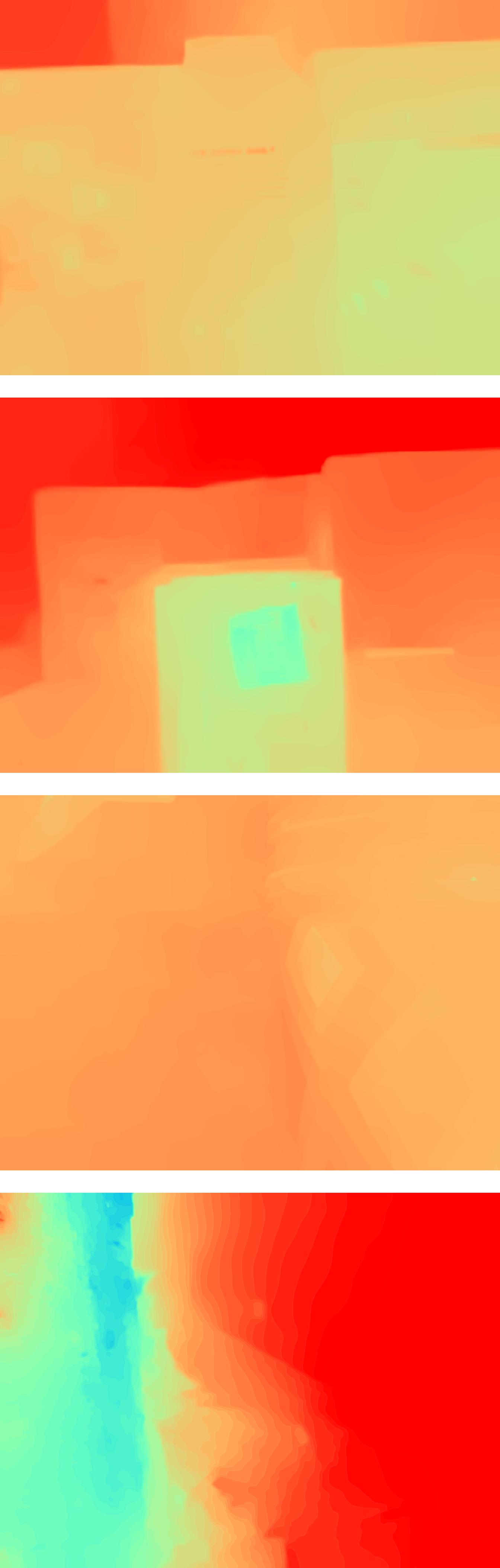}\label{subfig:defocus_Siggraph_comparison}}
	\vspace{-0.2in}
	\caption{More qualitative comparisons of defocus map estimation methods.}
	\label{fig:comparison_defocus_map_supp}
	\vspace{-0.1in}
\end{figure*}

\begin{figure*}[t]
	\centering
	\subfigure[Input image]{\includegraphics[width=0.12\textwidth]{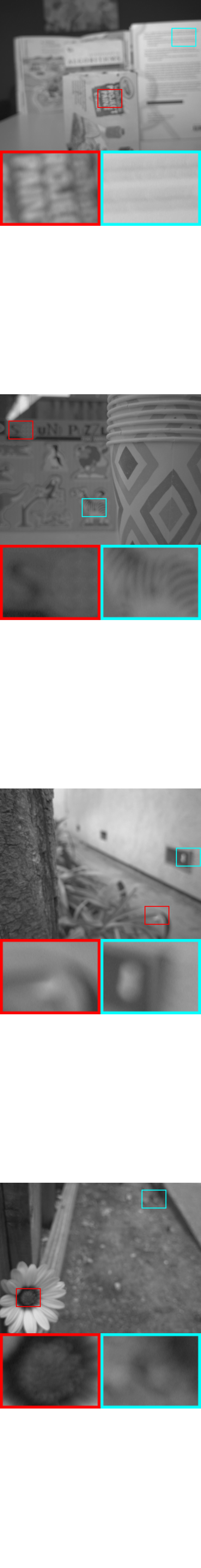}\label{subfig:inputs_combined_ablation}}
	\subfigure[Ground truth]{\includegraphics[width=0.12\textwidth]{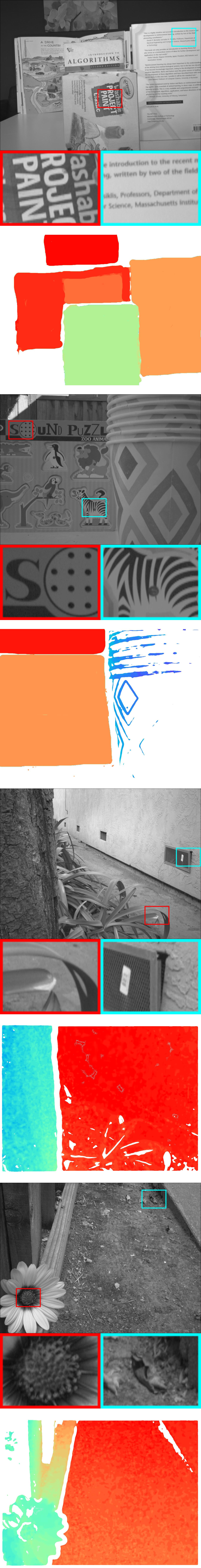}\label{subfig:combined_gt_ablation}}
	\subfigure[Ours full]{\includegraphics[width=0.12\textwidth]{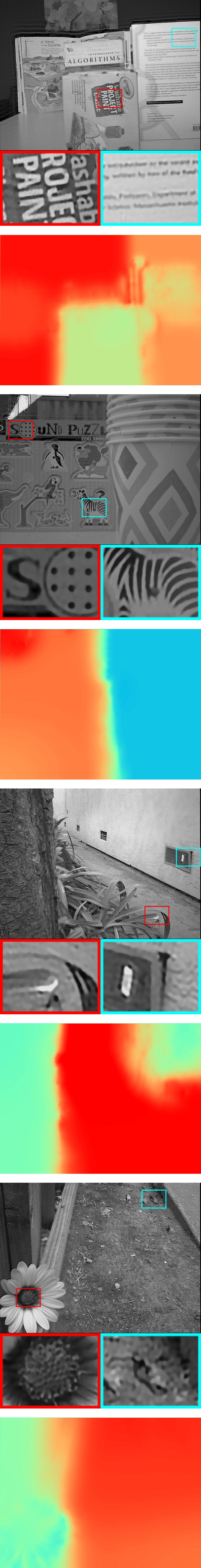}\label{subfig:combined_ours_ablation}}
	\subfigure[No $\loss_{\mathrm{intensity}}$]{\includegraphics[width=0.12\textwidth]{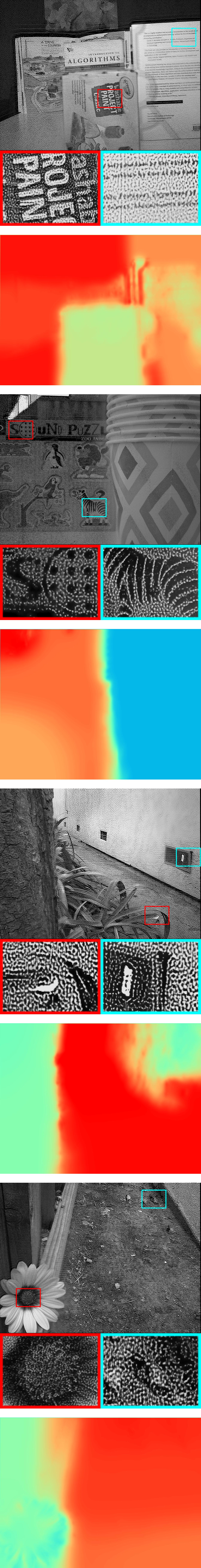}\label{subfig:combined_ours_no_im_tv_ablation}}
	\subfigure[No $\loss_{\mathrm{alpha}}$]{\includegraphics[width=0.12\textwidth]{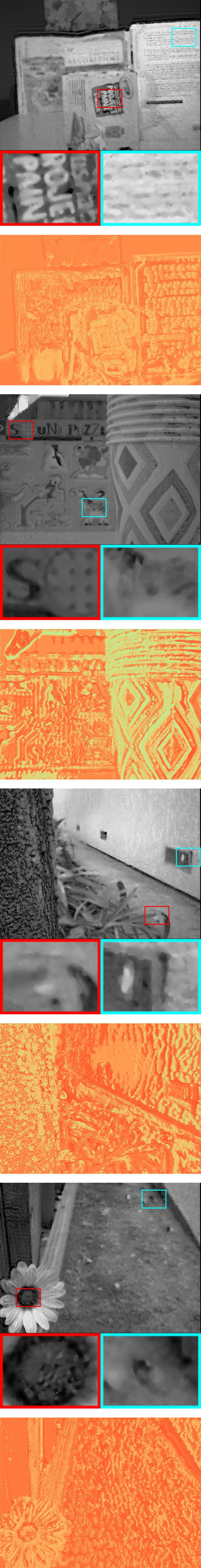}\label{subfig:combined_ours_no_alpha_tv_ablation}}
	\subfigure[No $\loss_{\mathrm{entropy}}$]{\includegraphics[width=0.12\textwidth]{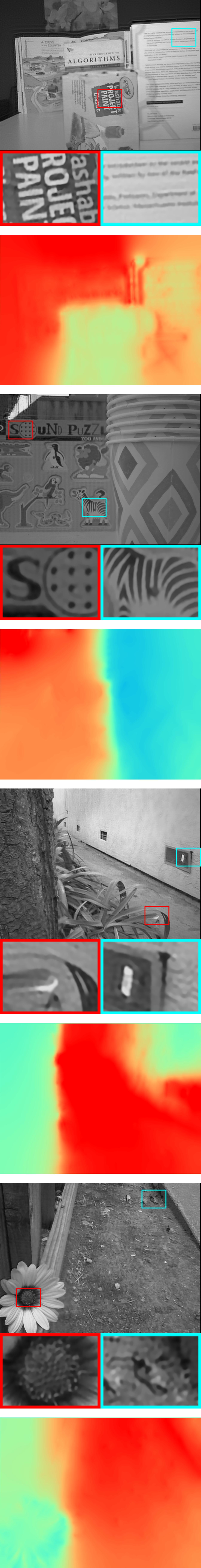}\label{subfig:combined_ours_no_entropy_ablation}}
	\subfigure[No $\loss_{\mathrm{aux}}$]{\includegraphics[width=0.12\textwidth]{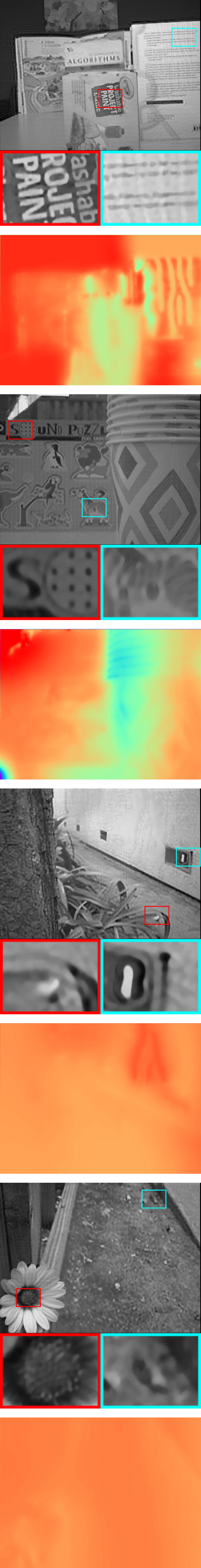}\label{subfig:combined_ours_no_auxiliary_ablation}}
	\subfigure[No $\bias$]{\includegraphics[width=0.12\textwidth]{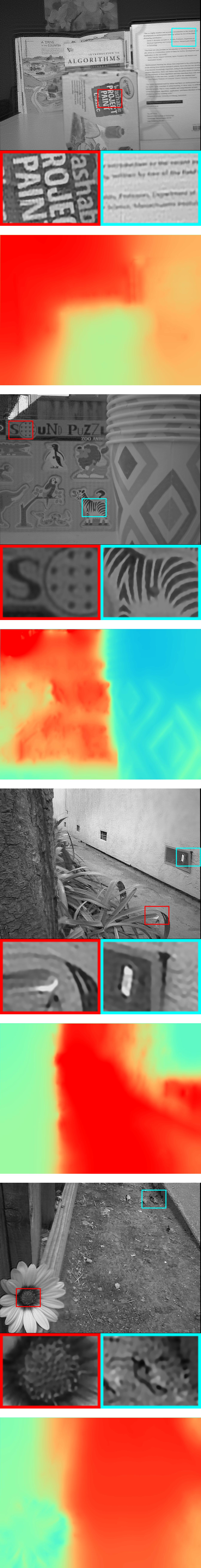}\label{subfig:combined_ours_no_bias_correction_ablation}}
	\vspace{-0.2in}
	\caption{More qualitative results on ablation study.}
	\label{fig:ablation_combined_supp}
	\vspace{-0.1in}
\end{figure*}

We first discuss more implementation details about our method, then show qualitative results on more data in our collected Google Pixel 4 dataset in Fig.~\ref{fig:comparison_deblurring_supp}-\ref{fig:ablation_combined_supp}. We also provide an interactive HTML viewer~\cite{bitterli2017reversible} at the project website~\cite{ProjectWebsite} to facilitate the comparisons.

\noindent{\bf Data normalization.} Before running the optimization, we first compute an intensity scaling factor $s = \sfrac{0.5} {\mathrm{mean}\paren{\I_{o}^{\curly{l,r}}}}$, and normalize the inputs $\bar{\I}_{o}^{\curly{l,r}} = s \I_{o}^{\curly{l,r}}$ to account for global intensity variations. After optimization, we undo the normalization by dividing the all-in-focus image by $s$. 

\noindent{\bf Scaling factors of each loss term.} Recall that our optimization loss is
\begin{equation}
	\loss = \lambda_1 \loss_{\mathrm{data}} + \lambda_2  \loss_{\mathrm{aux}} + \lambda_3 \loss_{\mathrm{intensity}} + \lambda_4 \loss_{\mathrm{alpha}} + \lambda_5 \loss_{\mathrm{entropy}}\,.
\end{equation}
$\loss_{\mathrm{data}}$ and $\loss_{\mathrm{aux}}$ have the same weight: $\lambda_1 = \lambda_2 = 2.5\cdot 10^4$. For most scenes, 
$\lambda_3 = 30$, $\lambda_4$ = $7.5\cdot 10^4$, and $\lambda_5$ = $12$. We set higher weights on the regularization terms $\loss_{\mathrm{intensity}}$, $\loss_{\mathrm{alpha}}$, and $\loss_{\mathrm{entropy}}$ for scenes with less texture, e.g., data from Abuolaim and Brown~\cite{abuolaim2020defocus}.

\noindent{\bf Kernel size of each MPI layer.} We manually determine the kernel sizes of the front and back layers, and evenly distribute MPI layers in disparity space. As mentioned in Sec.~\ref{sec:dualpixel_supp}, we choose focus settings such that all scene contents are at or behind the focus plane. Therefore the kernel size of the front MPI layer is usually set to a small positive number, e.g., in the range of $1 \times 1$ to $3 \times 3$, to mimic a 2D delta function, while the kernel size corresponding to the back MPI layer is set to a large enough value, e.g., in the range of $57 \times 57$ to $61 \times 61$, to represent blur kernels at infinity.

\noindent{\bf Quantitative Metrics.}
We use the commercial software, Helicon Focus~\cite{helicon_focus} to compute the ground truth all-in-focus images and the defocus maps using focus stacking. There may be a small \emph{shift} between the ground truth all-in-focus image and the all-in-focus image from the deblurring algorithms we evaluate. This is because one can apply an arbitrary transform to the blur kernels and an inverse transform to the recovered all-in-focus image to yield the same blurred image. We determine this shift for each algorithm by using OpenCV to align the ground-truth all-in-focus image with the all-in-focus image from the algorithm via an affine transform for a single specific scene, and then using that transform to align all images before computing the metrics for all-in-focus images.  We also crop a small border of 8 pixels before computing the metrics as it may contain invalid pixels after alignment.

\begin{figure*}[!h]
	\centering
	\includegraphics[width=0.98\textwidth]{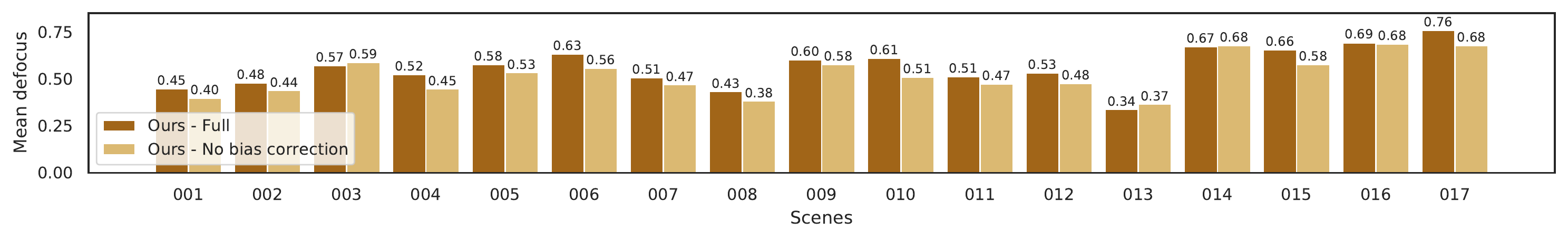}
	\caption{Mean of the predicted defocus map for our full pipeline vs an ablation where bias correction term is not applied. Defocus is measured as the relative scaling applied to the calibrated kernels. Without bias correction, the mean defocus is lower in 14 of the 17 scenes, i.e., the prediction is biased towards smaller defocus size.}
	\label{fig:bias_correction_supp}
\end{figure*} 

\noindent{\bf Additional results for bias correction.} Fig.~\ref{fig:bias_correction_supp} shows additional results for our ablation study. Specifically, it shows that without bias correction term $\bias$, the estimated defocus size is smaller on average as predicted by our analysis.

\end{document}